\documentclass{article} 
\usepackage{iclr2027_conference,times}

\iclrfinalcopy

\usepackage[utf8]{inputenc} 
\usepackage[T1]{fontenc}    
\usepackage{hyperref}       
\usepackage{url}            
\usepackage{booktabs}       
\usepackage{amsfonts}       
\usepackage{nicefrac}       
\usepackage{xcolor}         

\usepackage{microtype}
\usepackage{graphicx}
\usepackage{booktabs} 
\usepackage{xspace}
\usepackage[inline]{enumitem}
\usepackage{subcaption}
\usepackage{multirow}
\usepackage{makecell}
\usepackage{pairedcolors}
\usepackage{wrapfig}
\usepackage{hyperref}

\usepackage{amsmath}
\usepackage{amssymb}
\usepackage{mathtools}
\usepackage{amsthm}

\usepackage[capitalize,noabbrev,nameinlink]{cleveref}

\title{\acronym: Fast Safe Reinforcement Learning Benchmarking}

\author{%
  Tristan Tomilin\thanks{Corresponding author: \texttt{t.tomilin@tue.nl}}
  \quad Mourad Boustani
  \quad Mickey Beurskens
  \quad Khoi H. B. Nguyen
  \quad Thiago D.~Sim\~{a}o \\
  Eindhoven University of Technology
}

\usepackage[acronym]{glossaries}

\newacronym{mdp}{MDP}{Markov decision process}

\newacronym{cmdp}{CMDP}{constrained Markov decision process}

\usepackage{graphicx}
\usepackage{natbib}
\usepackage[capitalize,noabbrev]{cleveref}

\usepackage{xargs}
\usepackage[dvipsnames]{xcolor}
\usepackage[table]{xcolor}
\usepackage[textwidth=15mm,
	backgroundcolor=white,
    tickmarkheight=2pt,
	textsize=tiny]{todonotes}
\newcommand{\acronym}{CRAX\xspace}
\newcommand{\fullacronym}{\textbf{C}onstrained \textbf{R}L \textbf{A}ccelerated with JA{\textbf{X}}}

\definecolor{someblue}{HTML}{598BE7}
\colorlet{linkcolor}{someblue}
\hypersetup{colorlinks=true,urlcolor=linkcolor,citecolor=linkcolor,linkcolor=linkcolor}

\definecolor{safegreen}{RGB}{34, 139, 34}

\usepackage{comment}
\usepackage{silence}
\begin{document}

\maketitle
    
\begin{abstract}
Safety is a core concern for deploying reinforcement learning (RL) agents in real-world domains such as robotics and autonomous driving. While benchmarks have been central to progress in RL, existing 3D physics-based safety benchmarks remain computationally slow, limiting large-scale experimentation and rapid prototyping. To address this gap, we propose \textbf{\acronym}~(\fullacronym). Built on top of the MuJoCo XLA (MJX) physics engine, \acronym~leverages vectorized operations and hardware acceleration, yielding up to 200x faster training over comparable CPU-based safety benchmarks. The benchmark features eight tasks spanning three difficulty levels and multiple agent morphologies. Evaluating seven popular safe RL methods, we find that none dominates across tasks, and that learning safe policies from pixels remains largely unsolved.
\end{abstract}

\section{Introduction}
\label{sec:intro}

\newcommand{\saferl}{SafeRL\xspace}

Although the progress in reinforcement learning~\citep[RL;][]{DBLP:books/lib/SuttonB2018} has been sped up by the use of hardware acceleration, this progress has not yet translated to the research in safe RL \citep[\saferl;][]{DBLP:journals/jmlr/GarciaF15}.
Benchmarks are a catalyst driving innovation in research; for instance, ImageNet~\citep{DBLP:conf/cvpr/DengDSLL009} motivated the introduction of CNNs~\citep{DBLP:conf/nips/KrizhevskySH12}, and the Arcade Learning Environment~\citep{DBLP:journals/jair/BellemareNVB13} supported the development of deep Q-networks~\citep{DBLP:journals/nature/MnihKSRVBGRFOPB15}.
In a similar trend, accelerated hardware brought a new wave of benchmarks that facilitate research in multiple areas of RL, including goal-conditioned~\citep{DBLP:conf/iclr/BortkiewiczPMDA25},
    multi-agent~\citep{DBLP:conf/nips/RutherfordEG0LI24},
    offline~\citep{jackson2025a}, and
    open-ended~\citep{DBLP:conf/icml/MatthewsBESJCF24} RL.
Within the \saferl literature, Safety Gym~\citep{ray2019benchmarking} and, more recently, Safety-Gymnasium~\citep{DBLP:conf/nips/JiZZP0SGZD023} have established a set of common tasks that facilitate comparisons between \saferl algorithms. 
Nevertheless, \saferl research still relies mostly on CPU-based simulations and does not leverage GPU computation.

Effective research in RL, particularly in high-dimensional problems, requires fast data collection, as training RL typically requires a large number of environment interactions.
This demand can compromise the research development phase, where we retrain such agents numerous times from scratch, while performing:
hyperparameter tuning to ensure a fair comparison, tests in multiple environments to evaluate generalization, repeated runs for statistical significance, and ablation studies to assess individual components of an algorithm.
Together, these requirements form a bottleneck for the development of new algorithms.
Fast simulation is therefore essential to support research on RL.


By leveraging high-fidelity simulators with hardware acceleration, RL is increasingly closing the gap to real-world applications such as robotics.
Simulation platforms such as BRAX~\citep{DBLP:conf/nips/FreemanFRGMB21} and Isaac Lab~\citep{mittal2025isaaclab} run on accelerated hardware, allowing researchers to leverage substantial speedups from parallel computing architectures.
Such platforms enable large-scale policy training in simulation and, in some cases, transfer to physical robots~\citep{zakka2025mujoco}.
Nevertheless, this approach still relies closely on reward engineering to specify the behavior desired from the agent.
However, in many situations, expressing such behaviors is easier through constraints~\citep{DBLP:conf/icml/RoyGRBP22}, particularly in safety-critical scenarios~\citep{ray2019benchmarking}.
Therefore, 
we focus on fast RL benchmarks with explicit constraints.


We introduce \textbf{\acronym}\footnote{The code and environments are accessible on 
\href{https://github.com/TTomilin/CRAX}{GitHub}.}~(\fullacronym), 
a novel hardware-accelerated \saferl benchmark leveraging MuJoCo, a general-purpose 3D physics engine.
The design principles of \acronym~are inspired by BRAX \citep{DBLP:conf/nips/FreemanFRGMB21} and Safety Gymnasium \citep{DBLP:conf/nips/JiZZP0SGZD023}.
\acronym provides a set of simulated tasks, robots, and algorithm baselines for evaluating \saferl leveraging parallel computing, resulting in higher simulation speeds than CPU-based setups, enabling more rigorous testing and faster algorithm development for the \saferl community.
Each task defines reward and cost signals, inducing a trade-off between performance and safety: achieving high reward typically incurs more cost, so satisfying safety constraints necessitates sacrificing some reward.
%
Among numerous constraint formulations~\citep{wachi2024survey}, we focus on bounding the expected cumulative discounted cost, the most widely adopted approach in the \saferl literature.
Nonetheless, \acronym can easily be extended to handle other formulations.

\acronym offers a diverse set of environments and algorithms. It includes eight tasks, eight agent morphologies, three difficulty levels per task, and three observation types (state vector, egocentric view, and allocentric view).
Furthermore, it reimplements seven \saferl algorithms in JAX, covering primal-dual, state-augmented, and trust-region approaches, with support for off-policy methods.
These include baseline algorithms for constrained RL, such as PPO Lagrangian~\citep[PPO-Lag;][]{ray2019benchmarking}, as well as unconstrained PPO~\citep{DBLP:journals/corr/SchulmanWDRK17} as a reference.
These implementations facilitate comparisons between new algorithms and relevant prior work.

The speed-ups of the hardware-accelerated implementation of \acronym enable:
\begin{enumerate}[leftmargin=*, topsep=2pt, itemsep=1pt, parsep=0pt]
    \item \textbf{Reliable assessment.}
        \acronym makes extensive evaluations affordable, such as wide hyperparameter sweeps and more repetitions per experiment, which increases confidence in the results.
    \item \textbf{Fast prototyping.}
        \acronym trains quickly, enabling rapid experimentation and speeding up the research process.
\end{enumerate}
\acronym reaches two orders of magnitude higher throughput than Safety-Gymnasium (\cref{sec:computational_efficiency}).
This throughput enables the evaluation of seven \saferl baselines across all tasks and difficulty levels, which shows that no method strictly dominates the others (\cref{sec:emperical_eval}).
We complement final performance with sample efficiency and constraint violations during training (\cref{app:sample-efficiency}), quantify how many seeds reliable comparisons require (\cref{app:seed_variance}), and sweep the hyperparameters of each method (\cref{app:sweeps}).
Beyond the baselines, we study learning from pixel observations (\cref{sec:learning_from_pixels}), the performance-safety trade-off under varying cost thresholds (\cref{app:safety_bounds}), safe curriculum learning and safety transfer (\cref{app:curriculum_transfer}), and off-policy methods (\cref{app:off_policy}).

\vspace{-0.5em}
\section{Related Work}
\label{sec:related_work}

\textbf{Safe Reinforcement Learning.}
In \saferl, besides achieving high-performance, agents also need to adhere to established safety requirements during learning and deployment~\citep{DBLP:journals/jmlr/GarciaF15}.
While safety can be encouraged indirectly through reward shaping, this approach places the burden of balancing performance and safety on the system designer. 
To reduce this burden, we can model safety requirements explicitly as constraints~\citep{DBLP:conf/icml/RoyGRBP22, kamran2022modern},
allowing the agent itself to autonomously find a trade-off between reward maximization and constraint satisfaction.
A wide range of safety formulations have been studied, including chance constraints, almost-sure constraints, and per-step constraints \citep{wachi2024survey}.
In practical work, the most common experimental settings bound the expected sum of discounted costs over time \citep{DBLP:journals/ral/YuanHZBGPS22,DBLP:conf/nips/JiZZP0SGZD023,DBLP:journals/corr/abs-2506-02255}.
While \acronym{} is agnostic to the specific safety formulation, our empirical evaluation adopts this setting due to its popularity.

\begin{table}[t]
\caption{Key characteristics of popular Reinforcement Learning benchmarks. \acronym~uniquely combines a focus on safety with hardware acceleration and 3D physics-based tasks.}
\label{tab:related_work}\centering
\resizebox{1\textwidth}{!}{\definecolor{orange_d}{RGB}{255, 193, 7}%
\newcommand{\partialo}{\color{orange_d}$\blacktriangle$\xspace}%
\newlength{\gapheight}%
\setlength{\gapheight}{0.7em}%
\newcommand{\yes}{\color{green_d}$\mathbf{\surd}$\xspace}%
\newcommand{\no}{\color{red_d}\large$\times$\xspace}%
\begin{tabular}{@{}lccc l r@{}}
\toprule
\textbf{Benchmark} & \textbf{Safety} & \textbf{GPU} & \textbf{3D Phys.} & \textbf{Type} & \textbf{Reference} \\
\midrule
OpenAI Gym              & \no & \no & \no & Classic control & \citep{DBLP:journals/corr/BrockmanCPSSTZ16}  \\
Procgen Benchmark       & \no & \no & \no & Procedural generation & \citep{DBLP:journals/corr/abs-2103-15332} \\
Atari (ALE)          & \no & \no & \no & Arcade games & \citep{DBLP:conf/ijcai/BellemareNVB15} \\[\gapheight]

Gymnax               & \no & \yes & \no & Classic control / MinAtar  & \citep{gymnax2022github} \\
JaxMARL              & \no & \yes & \no & Multi-agent & \citep{DBLP:conf/nips/RutherfordEG0LI24} \\
Craftax              & \no & \yes & \no & Open-ended / gridworld & \citep{DBLP:conf/icml/MatthewsBESJCF24} \\
Jumanji              & \no & \yes & \no & Combinatorial optimization & \citep{DBLP:conf/iclr/BonnetLBSADCMTK24} \\
XLand-MiniGrid       & \no & \yes & \no & Meta-RL & \citep{DBLP:conf/nips/NikulinKZASK24} \\
VMAS       & \no & \yes & \no & Multi Agent 2D Physics & \citep{bettini2022vmas} \\[\gapheight]
DeepMind Control Suite & \no & \no & \yes & 3D physics & \citep{DBLP:journals/corr/abs-1801-00690} \\
Meta-World           & \no & \no & \yes & Robotic manipulation & \citep{DBLP:conf/corl/YuQHJHFL19} \\[\gapheight]
Brax                         & \no & \yes & \yes & 3D physics & \citep{DBLP:conf/nips/FreemanFRGMB21} \\
Isaac Lab                         & \no & \yes & \yes & 3D physics & \citep{mittal2025isaaclab} \\[\gapheight]
AI Safety Gridworlds   & \yes & \no & \no & 2D gridworlds & \citep{leike2017ai} \\
HASARD              & \yes & \no & \no & FPS game & \citep{DBLP:conf/iclr/TomilinFP25} \\[\gapheight]
SafeOR-Gym          & \yes & \yes & \no  & Operations research & \citep{DBLP:journals/corr/abs-2506-02255} \\[\gapheight]
Bullet-Safety-Gym        & \yes & \no & \yes & Safe navigation & \citep{Gronauer2022BulletSafetyGym} \\
Safe-Control-Gym         & \yes & \no & \yes & Safe control & \citep{DBLP:journals/ral/YuanHZBGPS22} \\
Safety Gym (OpenAI)     & \yes & \no & \yes & Safe navigation & \citep{ray2019benchmarking} \\
Safety-Gymnasium    & \yes & \partialo & \yes  & Safe navigation / locomotion & \citep{DBLP:conf/nips/JiZZP0SGZD023} \\[\gapheight]
\rowcolor{orange!25} 
\acronym~(Ours)         & \yes & \yes & \yes  & Safe navigation / locomotion & \\
\specialrule{\heavyrulewidth}{0pt}{\belowbottomsep}
\multicolumn{6}{l}{\footnotesize{{\partialo}~A subset of two robotic manipulation suites are GPU-accelerated through "Safe Isaac Gym", which is part of Safety Gymnasium.}}
\end{tabular}}
\end{table}

\textbf{\saferl Benchmarks.}
Safe RL was predominantly studied in low-dimensional 2D settings, such as gridworlds in AI Safety Gridworlds~\citep{leike2017ai} and MiniGrid~\citep{chevalier2023minigrid} tasks adapted for safety.
More recent benchmarks have shifted toward environments for embodied, pixel-based learning \citep{dosovitskiy2017carla,li2022metadrive,DBLP:conf/iclr/TomilinFP25} and physics-based continuous control~\citep{DBLP:journals/ral/YuanHZBGPS22,DBLP:conf/nips/JiZZP0SGZD023}. These benchmarks are typically built on existing simulators rather than developing physics engines from scratch. For example, HASARD~\citep{DBLP:conf/iclr/TomilinFP25} is built on ViZDoom, while Safety-Gymnasium~\citep{DBLP:conf/nips/JiZZP0SGZD023} extends MuJoCo tasks, adding safety constraints. \acronym{} follows this design principle by building on MJX~\citep{mujocoMuJoCoMJX}, the JAX-based accelerated backend of MuJoCo~\citep{todorov2012mujoco}.

\textbf{Accelerated Benchmarking.}
RL experiments are typically data-intensive, as meaningful evaluation requires repeated environment interactions for hyperparameter tuning, statistical significance analysis, and testing across different tasks.
While GPUs are routinely used to accelerate neural network training, online RL remains constrained when environment rollouts are executed on the CPU, making simulation throughput the main bottleneck.
This has motivated moving both simulation and learning onto parallel hardware to accelerate the full training loop.
Brax~\citep{DBLP:conf/nips/FreemanFRGMB21} provides hardware-accelerated continuous-control environments in JAX.
VMAS~\citep{bettini2022vmas} and JaxMARL~\citep{DBLP:conf/nips/RutherfordEG0LI24} focus on scalable multi-agent RL.
The former creates 2D physics environments implemented in PyTorch, and the latter creates JAX-native variants of many popular multi-agent environments. Craftax~\citep{DBLP:conf/icml/MatthewsBESJCF24} explores procedurally generated grid-based worlds optimized for large-scale parallel training. \cref{tab:related_work} summarizes a number of such widely used RL benchmarks. \acronym expands the available selection of GPU-accelerated, safety-focused, and 3D physics-based environments.


\section{Constrained Reinforcement Learning}
\label{sec:crl}

A constrained Markov decision process~\citep[CMDP;][]{altman1999constrained} is an MDP~\citep{Puterman1994} with constraints, characterized by a tuple $\mathcal{M} = \langle S, A, P, r, c, d, \gamma \rangle$, where $S$ is a continuous state space, $A$ a continuous action space, $P$ a transition function $P \colon S {\times} A \to \text{Distr}(S)$, $r$ a reward function $r \colon S {\times} A \to \mathbb{R}^+$, $c$ a cost function $c \colon S {\times} A \to \mathbb{R}^+$, $d \in \mathbb{R}^+$ a cost threshold, and $\gamma \in [0, 1)$ a discount factor.

An RL agent interacting with a CMDP follows a stochastic policy $\pi \colon S \to \text{Distr}(A)$.
The value function $V^\pi(s)$ represents the expected cumulative discounted reward when following policy $\pi$ starting from state $s$
over a (potentially infinite) horizon~$T$:
$
V^\pi(s) = \mathbb{E}_\pi \left[ \sum_{t=0}^{T} \gamma^t r(s_t, a_t) \mid s_0 = s \right],
$
where the expectation $\mathbb{E}_\pi$ is taken over the trajectory distribution induced by policy $\pi$, with actions  $a_t \sim \pi(\cdot | s_t)$ and successor states $s_{t+1} \sim P(\cdot | s_t, a_t)$.
Similarly, the cost function $C^\pi(s)$ captures the expected cumulative discounted cost under policy $\pi$ starting from state $s$:
$
C^\pi(s) = \mathbb{E}_\pi \left[ \sum_{t=0}^{T} \gamma^t c(s_t, a_t) \mid s_0 = s \right].
$

The objective in the CMDP framework is to find an optimal policy $\pi^* \in \Pi$ that maximizes the expected cumulative reward while ensuring the expected cumulative cost remains below the threshold $d$ for all states $s \in S$. This constrained optimization problem is formulated as:
%
$
\max_{\pi \in \Pi} V^\pi(s) \quad \text{subject to} \quad C^\pi(s) \leq d, \quad \forall s \in S.
$
The constraint $C^\pi(s) \leq d$ enforces safety by requiring that the policy maintains cost levels below the specified threshold regardless of the initial state. This formulation, known as the expected cumulative cost constraint~\citep{wachi2024survey}, distinguishes CMDPs from standard MDPs, where the agent would simply maximize the reward without regard for cost constraints.

\begin{figure*}[t]
    \centering
    \captionsetup{justification=centering}
    \begin{subfigure}[t]{0.24\textwidth}
        \centering
        \includegraphics[width=\linewidth]{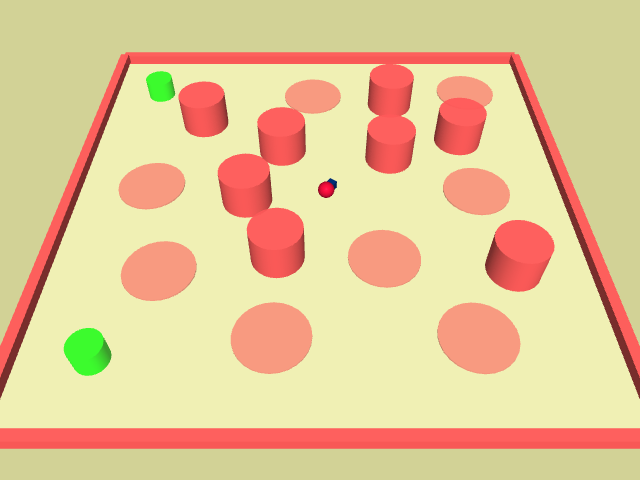}
        \caption*{\textbf{\texttt{Safe Goal}}\\{\footnotesize Reach the goal without colliding with the hazards}}
    \end{subfigure}
    \hfill
    \begin{subfigure}[t]{0.24\textwidth}
        \centering
        \includegraphics[width=\linewidth]{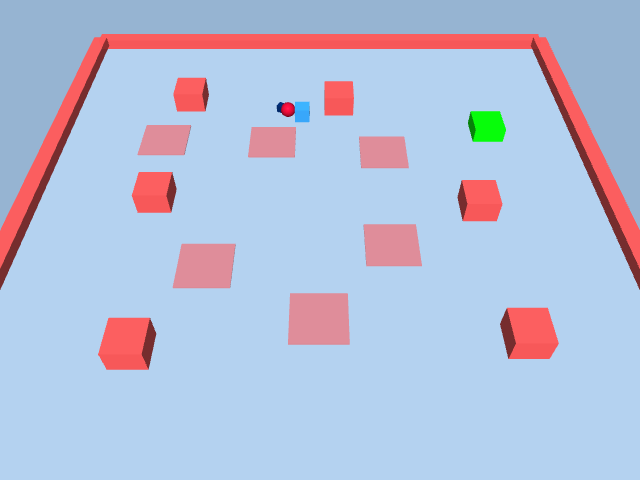}
        \caption*{\textbf{\texttt{Safe Push}}\\{\footnotesize Push the block to the goal while avoiding obstacles}}
    \end{subfigure}
    \hfill
    \begin{subfigure}[t]{0.24\textwidth}
        \centering
        \includegraphics[width=\linewidth]{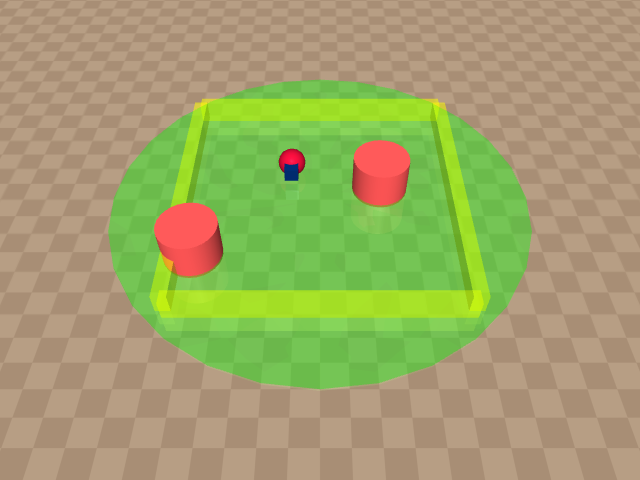}
        \caption*{\textbf{\texttt{Safe Circle}}\\{\footnotesize Move along a circular path while avoiding hazards}}
    \end{subfigure}
    \hfill
    \begin{subfigure}[t]{0.24\textwidth}
        \centering
        \includegraphics[width=\linewidth]{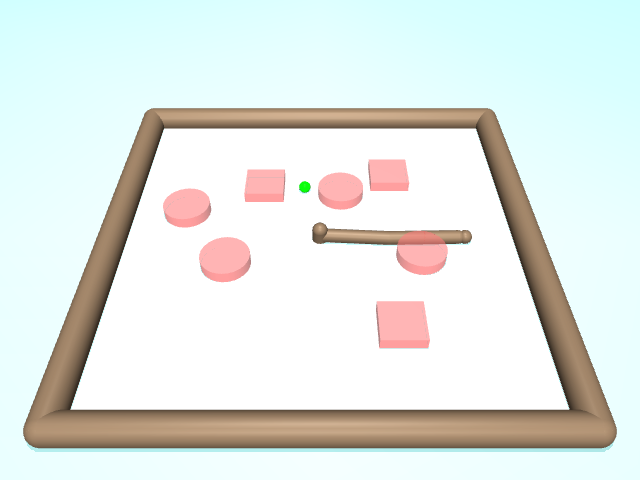}
        \caption*{\textbf{\texttt{Safe Reacher}}\\{\footnotesize Reach the target without touching the hazards}}
    \end{subfigure}

    \vspace{0.4em}

    \begin{subfigure}[t]{0.24\textwidth}
        \centering
        \includegraphics[width=\linewidth]{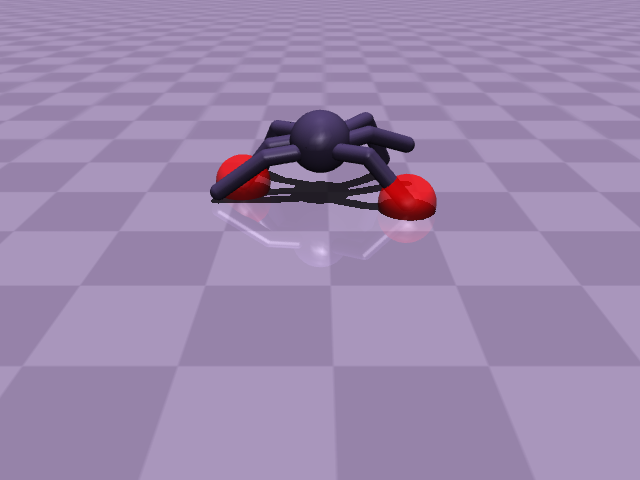}
        \caption*{\textbf{\texttt{Safe Lift}}\\{\footnotesize Move forward while lifting designated legs}}
    \end{subfigure}
    \hfill
    \begin{subfigure}[t]{0.24\textwidth}
        \centering
        \includegraphics[width=\linewidth]{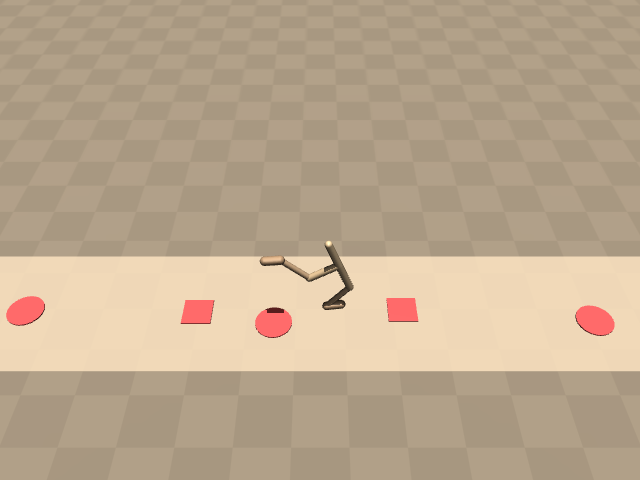}
        \caption*{\textbf{\texttt{Safe Pathway}}\\{\footnotesize Traverse the path without stepping on the hazards}}
    \end{subfigure}
    \hfill
    \begin{subfigure}[t]{0.24\textwidth}
        \centering
        \includegraphics[width=\linewidth]{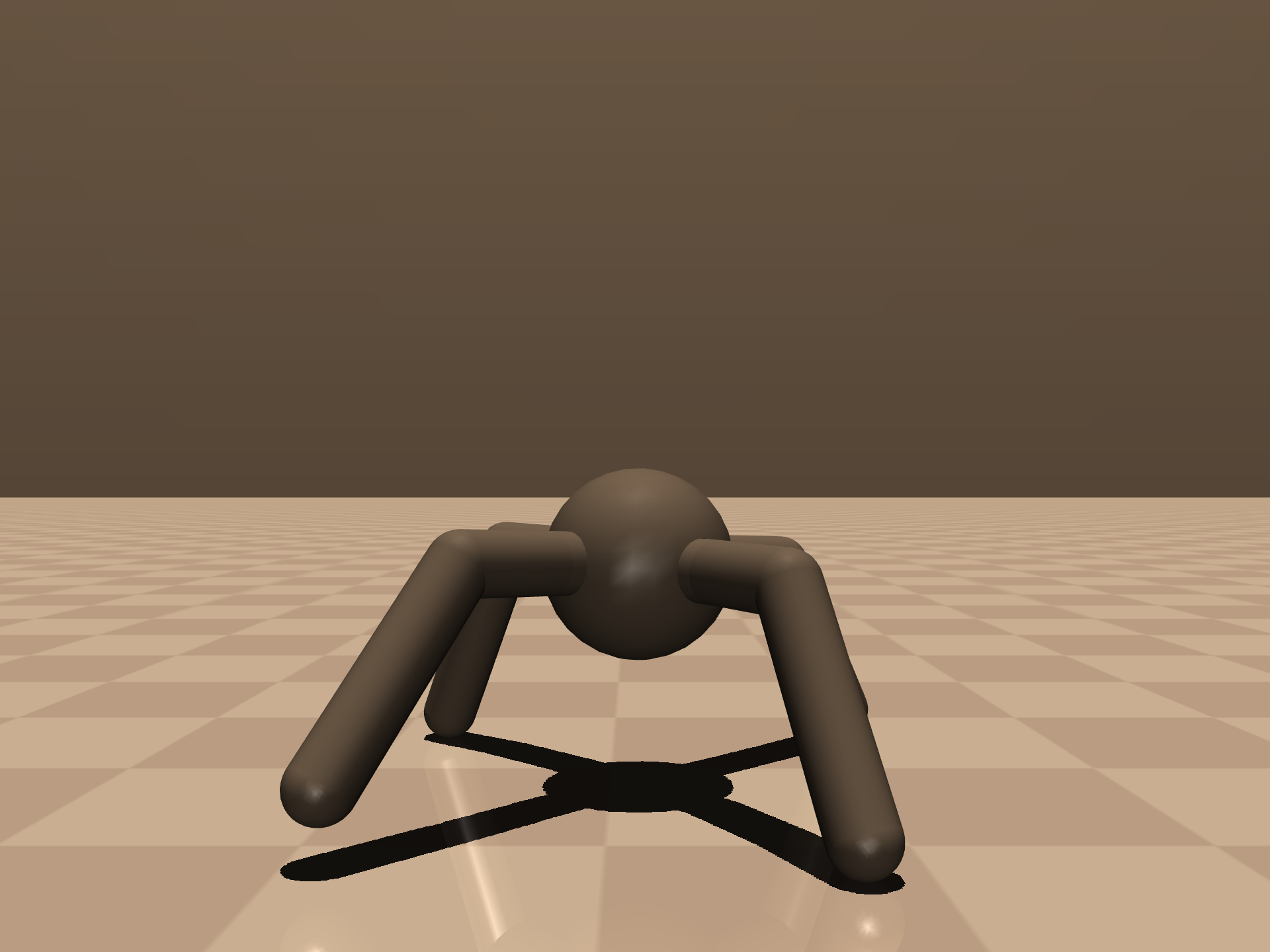}
        \caption*{\textbf{\texttt{Safe Velocity}}\\{\footnotesize Move fast while adhering to a velocity limit}}
    \end{subfigure}
    \hfill
    \begin{subfigure}[t]{0.24\textwidth}
        \centering
        \includegraphics[width=\linewidth]{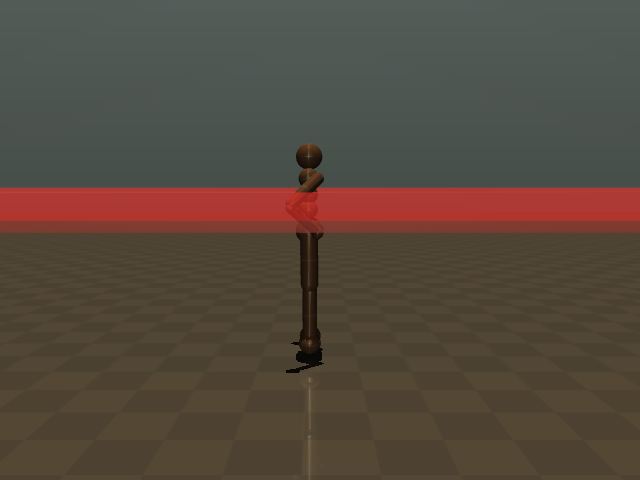}
        \caption*{\textbf{\texttt{Safe Height}}\\{\footnotesize Move forward under the height bound}}
    \end{subfigure}
    \caption{Overview of the CRAX benchmark environments.}
    \label{fig:env_overview}
\end{figure*}

\begin{table}[t]
\caption{Overview of the available suites in the benchmark (rows), and which agents are compatible with them (columns). Starred entries ($\star$) have been selected for evaluation in \cref{sec:emperical_eval}.}
\label{tab:suites}
\centering
\resizebox{\textwidth}{!}{\definecolor{green_d}{RGB}{34, 139, 34}%
\definecolor{red_d}{RGB}{220, 20, 60}%
\colorlet{3d_color}{purple_l!40}%
\colorlet{2d_color}{blue_l!40}%
\colorlet{static_color}{yellow_l!40}%
\newcommand{\yes}{\textcolor{green_d}{$\surd$}}%
\newcommand{\no}{\textcolor{black!25}{--}}%
\newcommand{\yeseval}{\textcolor{green_d}{$\surd$}\rlap{\textsuperscript{$\star$}}}%
\newcommand{\tc}{\cellcolor{3d_color}}%
\newcommand{\dc}{\cellcolor{2d_color}}%
\newcommand{\fc}{\cellcolor{static_color}}%
\newcolumntype{C}[1]{>{\centering\arraybackslash}m{#1}}%
\setlength{\aboverulesep}{0pt}%
\setlength{\belowrulesep}{0pt}%
\setlength{\extrarowheight}{1.5pt}%
\begin{tabular}{@{}l*{4}{C{1.45cm}}C{1.9cm}*{2}{C{1.45cm}}C{1.8cm}@{}}
\toprule
 & \multicolumn{4}{c}{\tc\textbf{3D}} & \multicolumn{3}{c}{\dc\textbf{2D}} & \multicolumn{1}{C{1.8cm}@{}}{\fc\textbf{Fixed base}} \\
\textbf{Task} & \textbf{Point} & \textbf{Ant} & \textbf{Humanoid} & \textbf{Spider} & \textbf{HalfCheetah} & \textbf{Walker2d} & \textbf{Hopper} & \textbf{Reacher} \\
\midrule
Goal     & \tc\yeseval & \tc\yes & \tc\yes     & \tc\yes     & \no & \no & \no & \no \\
Circle   & \tc\yeseval & \tc\yes & \tc\yes     & \tc\yes     & \no & \no & \no & \no \\
Push     & \tc\yeseval & \tc\yes & \tc\yes     & \tc\yes     & \no & \no & \no & \no \\
Velocity & \tc\yes     & \tc\yes & \tc\yeseval & \tc\yes     & \dc\yes & \dc\yes & \dc\yes & \no \\
Height   & \no & \no & \tc\yeseval & \no & \no & \no & \no & \no \\
Lift     & \no & \no & \no & \tc\yeseval & \no & \no & \no & \no \\
Pathway  & \no & \no & \no & \no & \dc\yes & \dc\yeseval & \dc\yes & \no \\
Reach    & \no & \no & \no & \no & \no & \no & \no & \fc\yeseval \\
\bottomrule
\end{tabular}}
\end{table}

\section{CRAX}
This section covers the design choices behind \acronym. The benchmark has been inspired by the GPU-accelerated RL environments in BRAX~\citep{DBLP:conf/nips/FreemanFRGMB21} and the safety environments of Safety Gymnasium~\citep{DBLP:conf/nips/JiZZP0SGZD023}. It is intended as a research and benchmarking platform for \saferl, providing ready-to-use environment suites, agent morphologies, and utility tooling to design \saferl experiments, alongside a set of pre-configured tasks of increasing difficulty. Every task exposes a cost signal in addition to a reward, and the entire training loop runs on the GPU through JAX. Two further principles shaped the design. First, the benchmark should be neither immediately solvable by current methods nor so hard that no progress is possible, which is why each suite comes in three difficulty levels. Second, it should include the tooling needed to assess the safety properties of the algorithms under evaluation, not only their returns.

\subsection{Environment Suites and Tasks}
\label{sec:suites_tasks}

Environment \textit{suites} define families of configurable tasks in simulated 3D environments. In each suite, an agent seeks to maximize reward while adhering to a predefined cost threshold.
\cref{fig:env_overview} summarizes the objectives associated with each suite.
Beyond suite-specific parameters, such as the number of obstacles in \textit{Point Goal}, every suite specifies its own reward and cost signals. 
An instantiation of a suite’s parameters constitutes a \textit{task}.
\cref{app:suites} provides a detailed description of all suites.
The Navigation and Velocity suites descend from Safety-Gymnasium~\citep{DBLP:conf/nips/JiZZP0SGZD023}, while Pathway, Height, Lift and Reach are new. \cref{app:sg_fidelity} brings an example of how closely an inherited suite tracks its original and compares training curves on the closest matching task.

\begin{figure}[!t]
  \centering
  \begin{subfigure}[b]{0.325\linewidth}
    \includegraphics[width=0.49\linewidth]{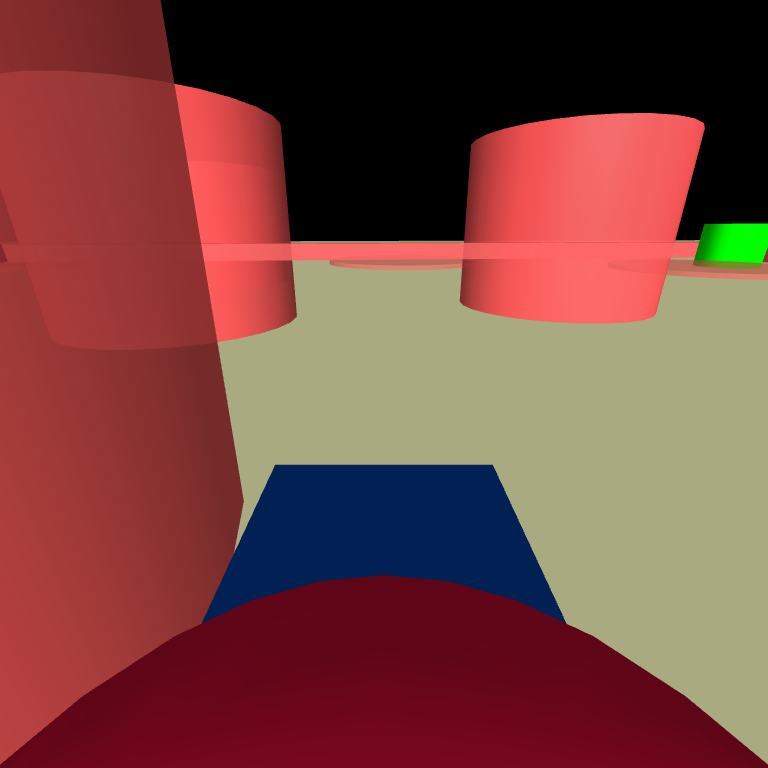}\hfill
    \includegraphics[width=0.49\linewidth]{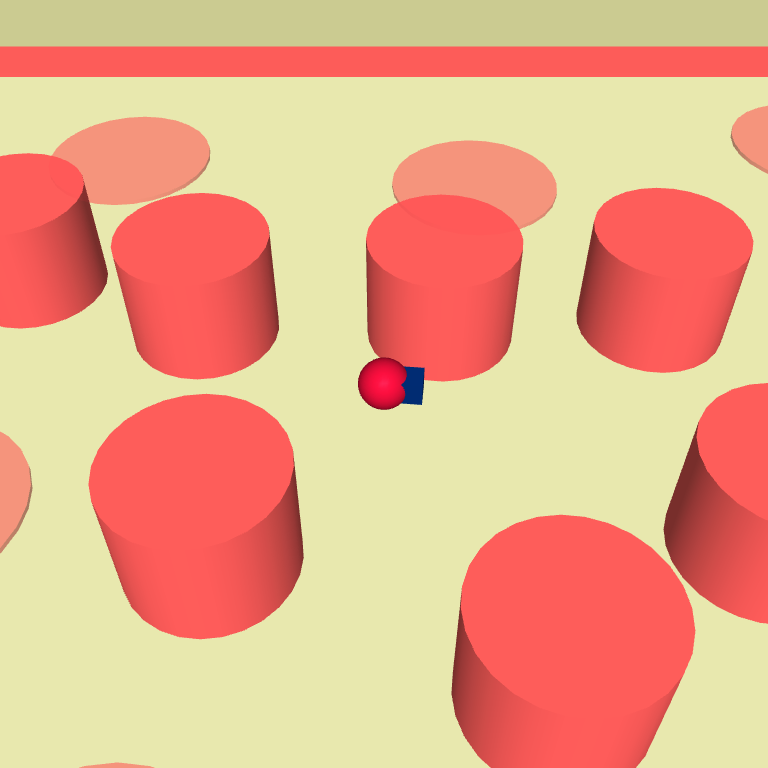}
    \caption*{\texttt{Point Goal}}
  \end{subfigure}\hfill
  \begin{subfigure}[b]{0.325\linewidth}
    \includegraphics[width=0.49\linewidth]{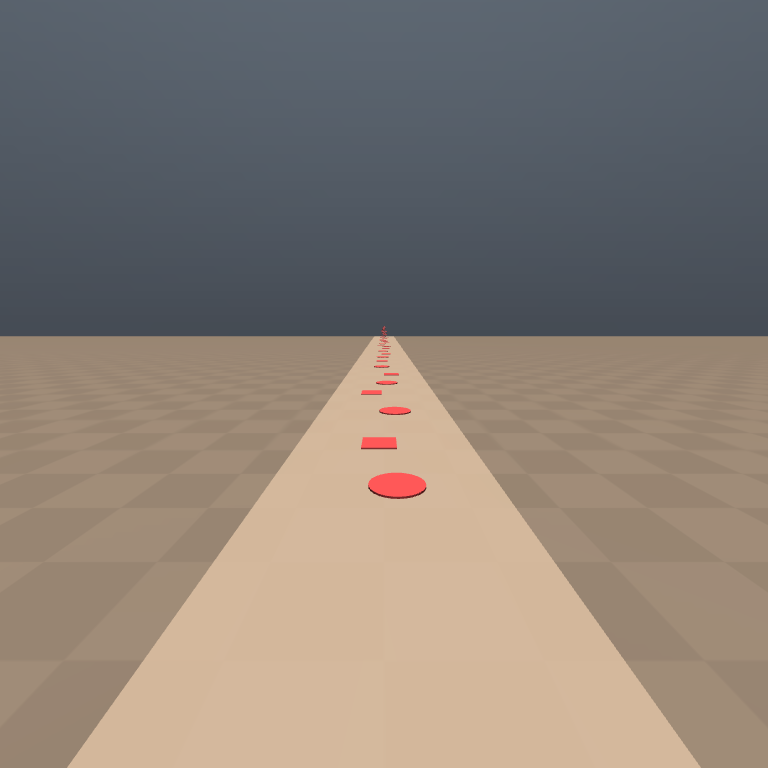}\hfill
    \includegraphics[width=0.49\linewidth]{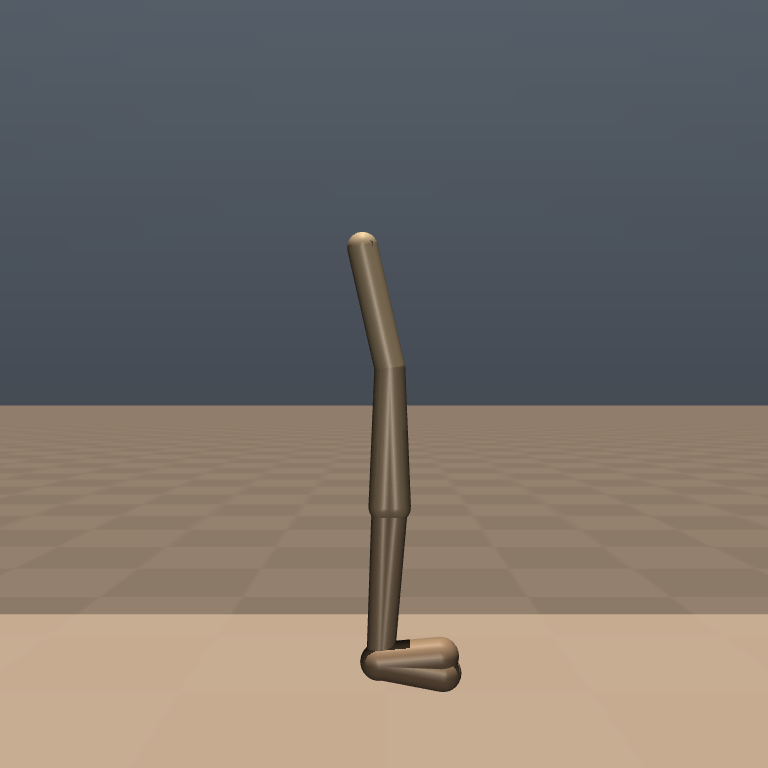}
    \caption*{\texttt{Walker Pathway}}
  \end{subfigure}\hfill
  \begin{subfigure}[b]{0.325\linewidth}
    \includegraphics[width=0.49\linewidth]{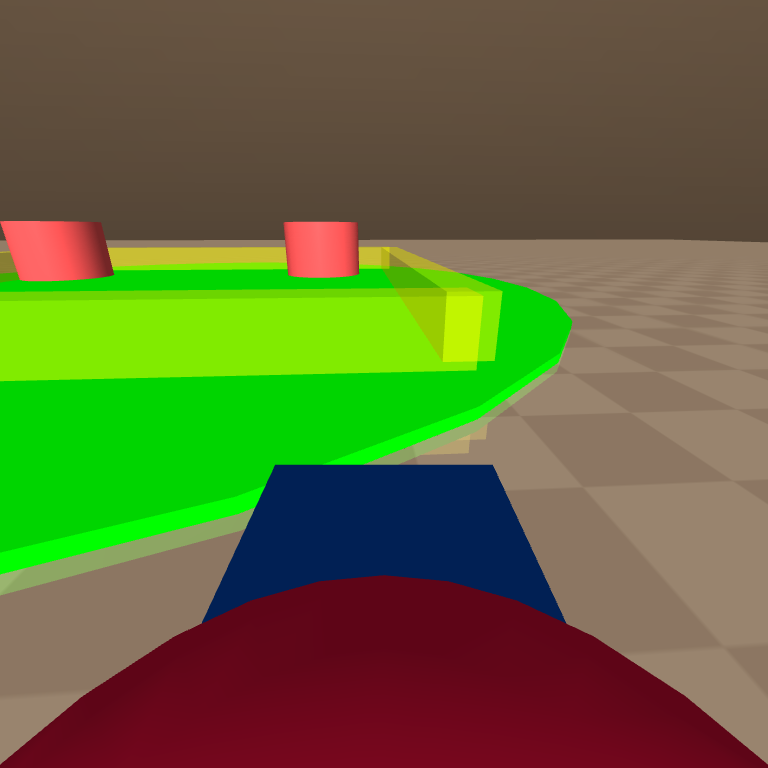}\hfill
    \includegraphics[width=0.49\linewidth]{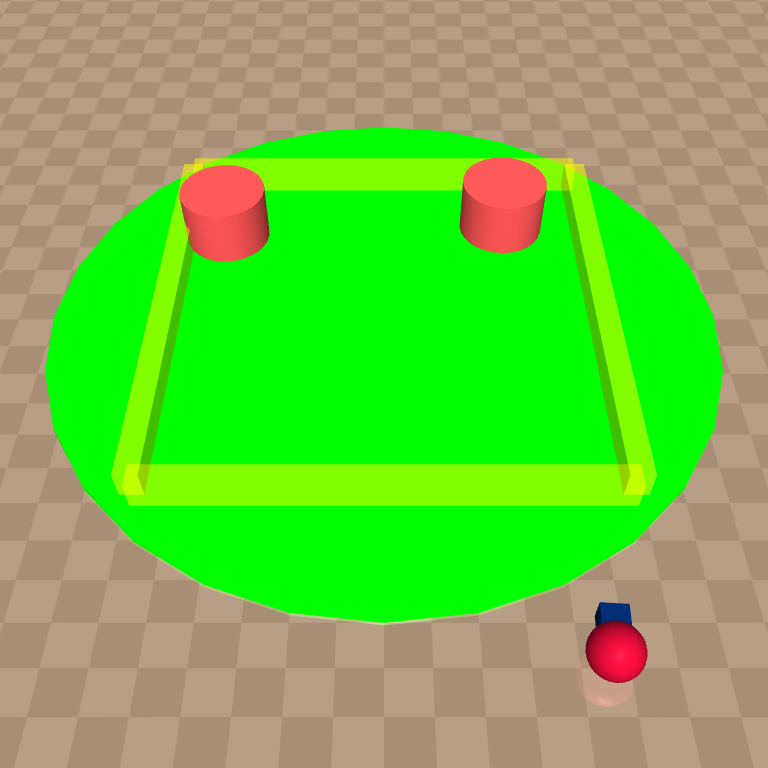}
    \caption*{\texttt{Point Circle}}
  \end{subfigure}
  \vspace{-4pt}
  \caption{\textbf{Egocentric} (left of each pair) and \textbf{allocentric} (right) views. Policies receive them at a $64\times64$ resolution. Here we show them at a higher resolution for legibility. The camera tracks the agent in \texttt{Point Goal} and \texttt{Walker Pathway}, but stays fixed in \texttt{Point Circle}.}
  \label{fig:modalities}
\end{figure}

\subsection{Agents}
\label{sec:agents}

Agents are the acting bodies in the environment.
Each agent has a unique morphology and lidar sensors to detect its surroundings.
The movement of some agents is constricted in one or more dimensions.    
Some suites are compatible with multiple agent types, while some of the suites allow for only a single agent type. 
\cref{tab:suites} provides an overview of the agent types and their compatibility with the suites.
\cref{app:agents} provides further details about the agents.

\subsection{Rewards, Costs and Constraint Types}
\label{sec:rewards_costs_constraints}

Each environment comes with a distinct default reward signal.
Cost signals are constructed using a variety of constraints.
\cref{app:suites} provides an overview of the reward and cost signals of each environment.
The benchmark employs five key constraint formulations. 
(1) \textbf{Hazard constraints}: binary cost on collision with solid obstacles or a cost growing with penetration into hazard regions.
(2) \textbf{Velocity constraints}: binary cost whenever the agent exceeds a speed limit.
(3) \textbf{Height constraints}: hinge cost that grows as the agent rises above a threshold.
(4) \textbf{Contact constraints}: binary cost whenever a restricted limb touches the ground.
(5) \textbf{Boundary constraints}: binary cost whenever the agent leaves the permitted region.
As established in \cref{sec:crl}, staying within the safety budget does not preclude further gains: higher returns can be achieved with more refined strategies while remaining safe. 
Moreover, the safety bound is adjustable, allowing one to impose stricter or more permissive requirements, and thereby modulate the task difficulty (\cref{app:safety_bounds}).

\subsection{Observation Modalities}
\label{sec:modalities}

By default, agents observe a vector of proprioceptive state together with lidar and compass readings (\cref{app:agents}), and all results use this modality unless stated otherwise.
\acronym~ also facilitates pixel-based observations.
The images are rendered on the GPU within the same compiled loop as the simulation, so pixel observations never leave the device, and any camera defined in a task can serve as an observation.
Every agent and task provides two views (\cref{fig:modalities}).
An egocentric camera is mounted on the agent, which is closer to how real robots perceive their surroundings.
An allocentric camera observes the agent from outside, and its placement depends on the task.
It follows the agent from above at an angle in \texttt{Point Goal}, stays fixed over the smaller arena of \texttt{Point Circle}, and follows the agent from the side in \texttt{Walker Pathway}.
Each actor and critic head declares which observations it consumes, so the modalities can be routed to different heads independently.

\begin{figure}[!t]
    \centering
    \hfill
    \includegraphics[width=.325\textwidth]{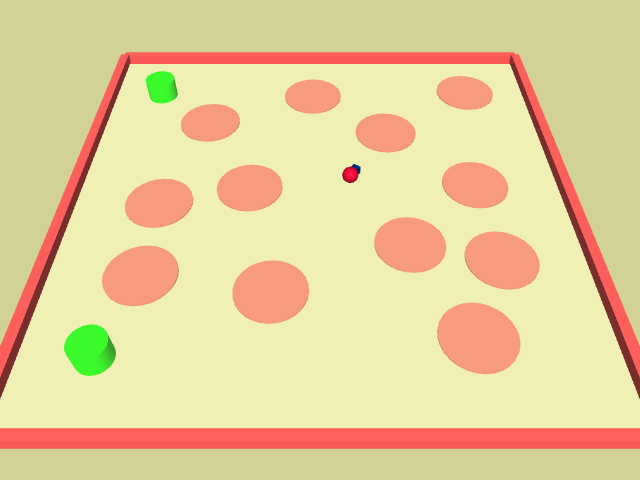}
    \hfill
        \includegraphics[width=.325\textwidth]{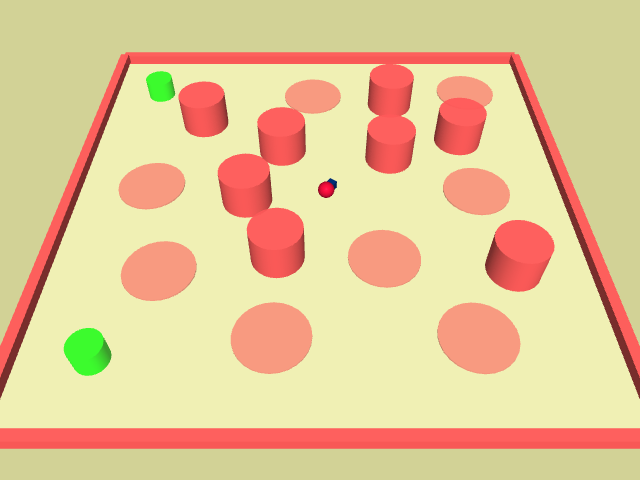}
    \hfill
        \includegraphics[width=.325\textwidth]{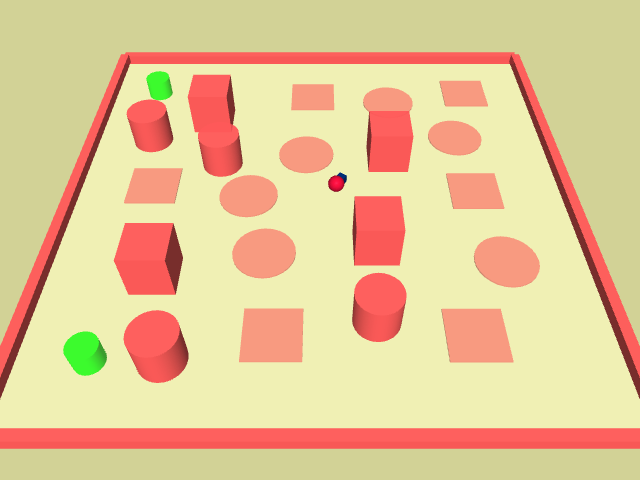}
    \hfill
    \caption{
    Higher \textbf{difficulty levels} of \texttt{Point Goal} (1 to 3, from left to right) decrease the size of the goals and increase the number and variety of hazards.
    }
    \label{fig:safe_point_goal_levels}
\end{figure}

\subsection{Difficulty Levels}

A useful benchmark ought to serve two complementary purposes. 
First, it should have a lenient evaluation setting that allows for comparing and analyzing existing methods. 
Second, it should pose a significant challenge to remain relevant for more advanced future methods.
To this end, we create each \acronym~suite in three difficulty levels. 
The lowest level tasks are designed such that most existing methods are capable of learning a reasonable policy and achieving meaningful performance, while the highest levels leave substantial room for improvement.
The difficulty increase between levels depends on the nature of the environment. For example, in \texttt{Point Goal}, higher levels introduce a greater number and variety of hazards that the agent must avoid (\cref{fig:safe_point_goal_levels}). 
In \texttt{Spider Lift}, each successive level requires the \textit{Spider} agent to keep one additional leg off the ground to avoid costs.
\cref{app:suites} provides the exact parameter settings defining each difficulty level, and \cref{fig:difficulty_grid} visually depicts the difficulty levels of the tasks.

\section{Empirical Evaluation}
\label{sec:emperical_eval}

To assess \acronym, we evaluate several popular \saferl baselines and one unconstrained RL baseline.
\begin{enumerate*}[label=(\arabic*)]
\item We include \textbf{PPO}~\citep{DBLP:journals/corr/SchulmanWDRK17} to serve as an unconstrained reference point, ignoring costs entirely.
\item \textbf{PPOCost}~\citep{DBLP:conf/iclr/TomilinFP25} extends PPO by treating costs as negative rewards.
\item \textbf{PPOLag}~\citep{ray2019benchmarking}, a primal-dual approach that updates both the policy and a learned Lagrange multiplier to balance return and safety.
\item \textbf{PPOPID}~\citep{DBLP:conf/icml/StookeAA20} refines \textbf{PPOLag}'s strategy by updating the Lagrange multiplier with a proportional-integral-derivative controller, allowing the reward-safety trade-off to adjust more responsively during training.
\item \textbf{PPOSauté}~\citep{DBLP:conf/icml/SootlaCJWMWA22} augments the observed state with a safety budget, treating the constraint as part of the dynamics.
\item \textbf{P3O}~\citep{zhang2022penalized} progressively increases a cost penalty coefficient when constraints are violated, encouraging the policy to adapt toward feasibility without explicit dual updates.
\item \textbf{FOCOPS}~\citep{zhang2020first} enforces safety by constraining policy updates through a trust-region formulation, optimizing reward while explicitly bounding expected cost.
\item Finally, \textbf{CRPO}~\citep{xu2021crpo} alternates between objectives without a multiplier, taking a reward-improving step when the estimated cost is within the threshold and a cost-reducing step otherwise.
\end{enumerate*}

\paragraph{Experimental Setup.}
We run all experiments for 500 million environment steps.
To choose the number of seeds, we measured how the confidence intervals narrow as seeds are added (\cref{app:seed_variance}). They narrow quickly up to five seeds and only slowly beyond, so we use ten seeds unless stated otherwise.
We report the undiscounted episodic reward and cost of the final policy, averaged over 128 evaluation episodes, as empirical estimates of $V^\pi$ and $C^\pi$ from \cref{sec:crl}, and call a policy safe when its episodic cost stays below the threshold $d = 25$.
All experiments are conducted on a dedicated compute node with a 72-core 3.2 GHz AMD EPYC 7F72 CPU and a single NVIDIA H100 GPU. 
\cref{app:hyperparameters} provides the exact hyperparameters with the respective sweeps.

\begin{figure}[!t]
    \centering
    \includegraphics[width=\linewidth]{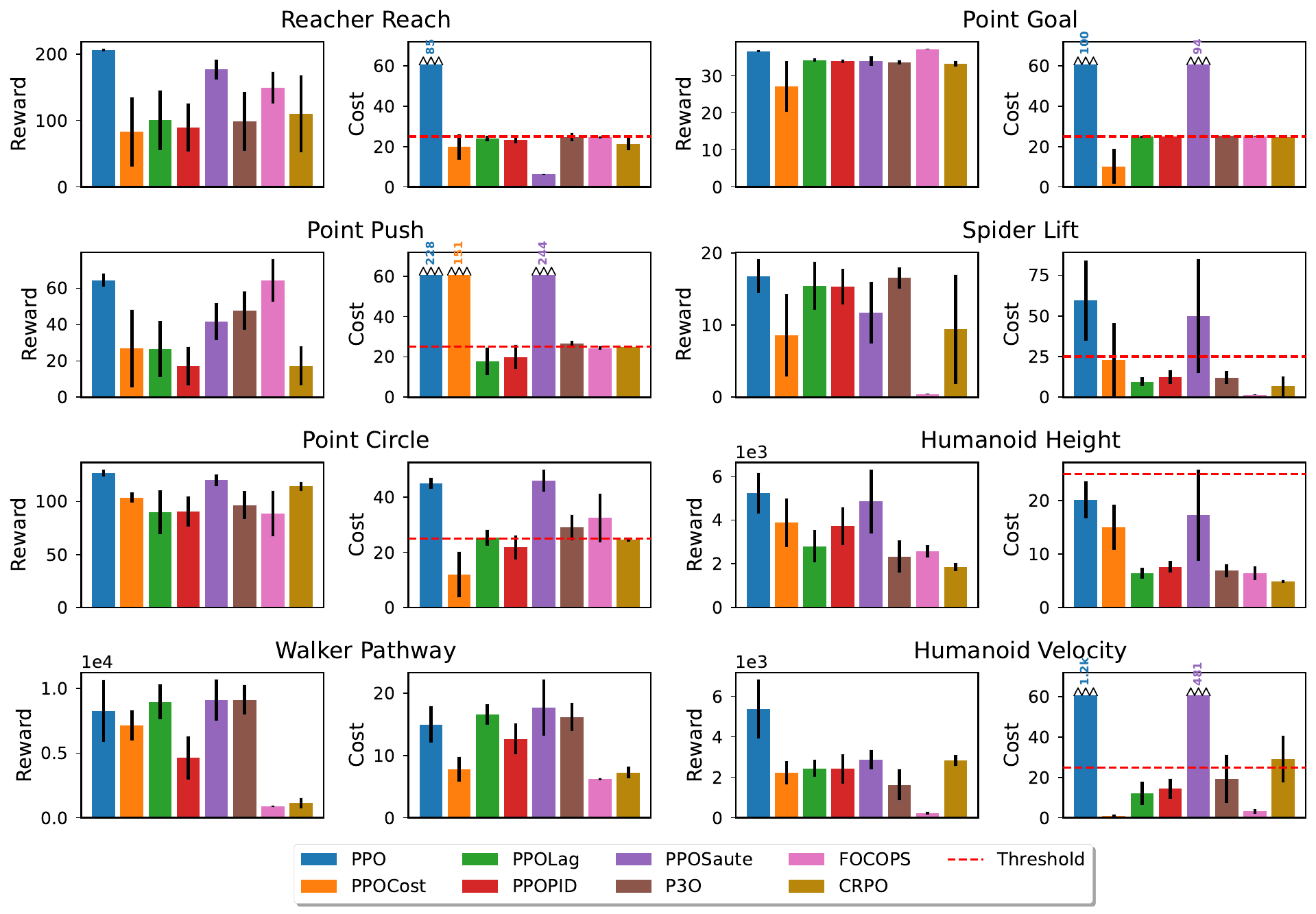}
    \caption{Rewards and costs of baseline methods on Level 1 tasks after 500M environment steps used for training. Error bars denote 95\% confidence intervals across 10 seeds.}
    \label{fig:baselines}
\end{figure}

\begin{table}[!t]
\caption{Algorithm summary results across 8 \acronym~environments. \textbf{Wins}: number of environments where the algorithm achieved the highest reward while being safe (cost $< 25$). \textbf{Safe\%}: percentage of environments where the algorithm was safe. \textbf{Total}: sum of wins and average safety percentage across levels. \textcolor{safegreen}{Green} indicates 100\% safe.}
\centering
\scalebox{.9}{\begin{tabular}{lcccccccc}
\toprule
 & \multicolumn{2}{c}{Level 1} & \multicolumn{2}{c}{Level 2} & \multicolumn{2}{c}{Level 3} & \multicolumn{2}{c}{Total} \\
\cmidrule(lr){2-3} \cmidrule(lr){4-5} \cmidrule(lr){6-7} \cmidrule(lr){8-9}
Algorithm & Wins $\uparrow$ & Safe\% $\uparrow$ & Wins $\uparrow$ & Safe\% $\uparrow$ & Wins $\uparrow$ & Safe\% $\uparrow$ & Wins $\uparrow$ & Safe\% $\uparrow$ \\
\cmidrule(lr){1-1} \cmidrule(lr){2-3} \cmidrule(lr){4-5} \cmidrule(lr){6-7} \cmidrule(lr){8-9}
PPO & 1 & 25 & 1 & 12 & 0 & 0 & 2 & 12 \\
PPOCost & 0 & 88 & 0 & 38 & 1 & 25 & 1 & 50 \\
PPOLag & \textbf{2} & 88 & 0 & \textcolor{safegreen}{100} & 0 & \textcolor{safegreen}{100} & 2 & 96 \\
PPOPID & 0 & \textcolor{safegreen}{100} & 0 & 50 & 1 & 88 & 1 & 79 \\
PPOSaute & 1 & 38 & 0 & 25 & 0 & 12 & 1 & 25 \\
P3O & \textbf{2} & 62 & \textbf{3} & 75 & \textbf{3} & 75 & \textbf{8} & 71 \\
FOCOPS & 1 & 75 & 2 & 88 & \textbf{3} & 88 & 6 & 83 \\
CRPO & 1 & 75 & 2 & 62 & 0 & 38 & 3 & 58 \\
\bottomrule
\end{tabular}}
\label{tab:alg_summary}
\end{table}

\subsection{Baseline Algorithm Analysis}

\cref{fig:baselines} shows the performance of baseline algorithms. PPO focuses entirely on maximizing rewards, providing a rough sense of the return achievable when safety is ignored. PPOCost simply subtracts costs from the reward, settling into a compromise, but offers no guarantee of adhering to the constraint. Some methods show distinct affinities for constraint types. PPOSauté satisfies the cost bound in \texttt{Reacher} and \texttt{Pathway} but otherwise behaves close to unconstrained PPO. FOCOPS performs best on navigation tasks (\texttt{Goal}, \texttt{Push}) and \texttt{Reacher}, but worst on forward-locomotion tasks (\texttt{Spider}, \texttt{Height}, \texttt{Pathway}). All baselines evaluated thus far are on-policy and most build on PPO, which keeps the comparison controlled across constraint-handling mechanisms and suits a fully compiled training loop. \acronym~nonetheless supports off-policy learning as well, and \cref{app:off_policy} compares SAC and SACLag with their on-policy counterparts, which they match in performance at about a third of the throughput.

\cref{tab:alg_summary} provides a summary of the evaluations. P3O and FOCOPS are the strongest baselines on \acronym. PPOLag achieves the highest safety percentage, being the only baseline to satisfy all cost bounds on Levels 2 and 3. However, it fails to reach high rewards. PPOPID is less conservative, trading stricter safety adherence for slightly higher performance. \cref{app:extended_results} provides more detailed baseline results and training curves, and \cref{app:sample-efficiency} reports how quickly each method reaches its performance and how much it violates safety along the way.

\vspace{-0.5em}
\subsection{Learning Safety from Pixels}
\label{sec:learning_from_pixels}

\begin{figure}[!t]
    \centering
    \includegraphics[width=\linewidth]{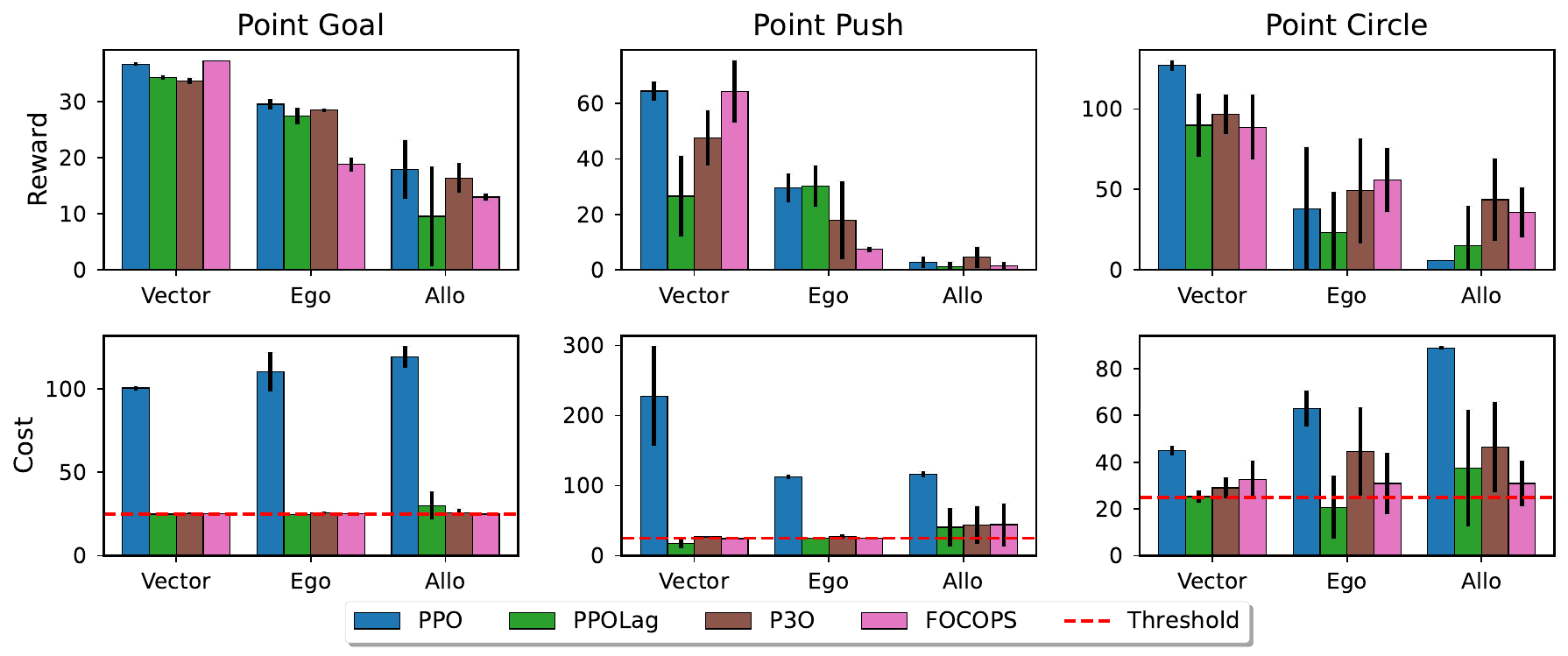}
    \caption{Final reward and cost on Level~1 when learning from \textbf{vector}, \textbf{egocentric (Ego)}, and \textbf{allocentric (Allo)} observations (See \cref{fig:modalities}). The dashed line marks the cost threshold. Error bars denote 95\% confidence intervals across 10 seeds.}
    \label{fig:obs_comparison}
\end{figure}

We train PPO, PPOLag, P3O, and FOCOPS from vector, egocentric, and allocentric observations.
Pixel-based agents receive $64\times64$ RGB frames, encode them with a convolutional network, and train with random-shift augmentation~\citep{kostrikov2021imageaugmentationneedregularizing}.
\cref{fig:obs_comparison} shows the results on Level~1 of \texttt{Point Goal}, \texttt{Point Push}, and \texttt{Point Circle}, as on the remaining five environments no method (including unconstrained PPO) learned a meaningful policy from pixels.
Every method earns less reward from egocentric frames than from vectors, and less still from allocentric frames, from which no method learns to push the block in \texttt{Point Push}.
Pixels also widen the confidence intervals and make the cost bound harder to satisfy, particularly from allocentric frames.
Learning from pixels thus worsens both sides of the trade-off, as the constrained methods give up reward without reliably staying within the bound.
Safe learning from pixels remains an open challenge in \acronym, particularly on the remaining environments.


\subsection{Computational Efficiency: A Case Study}
\label{sec:computational_efficiency}

\begin{figure}[!t]
    \centering
    \includegraphics[width=\linewidth]{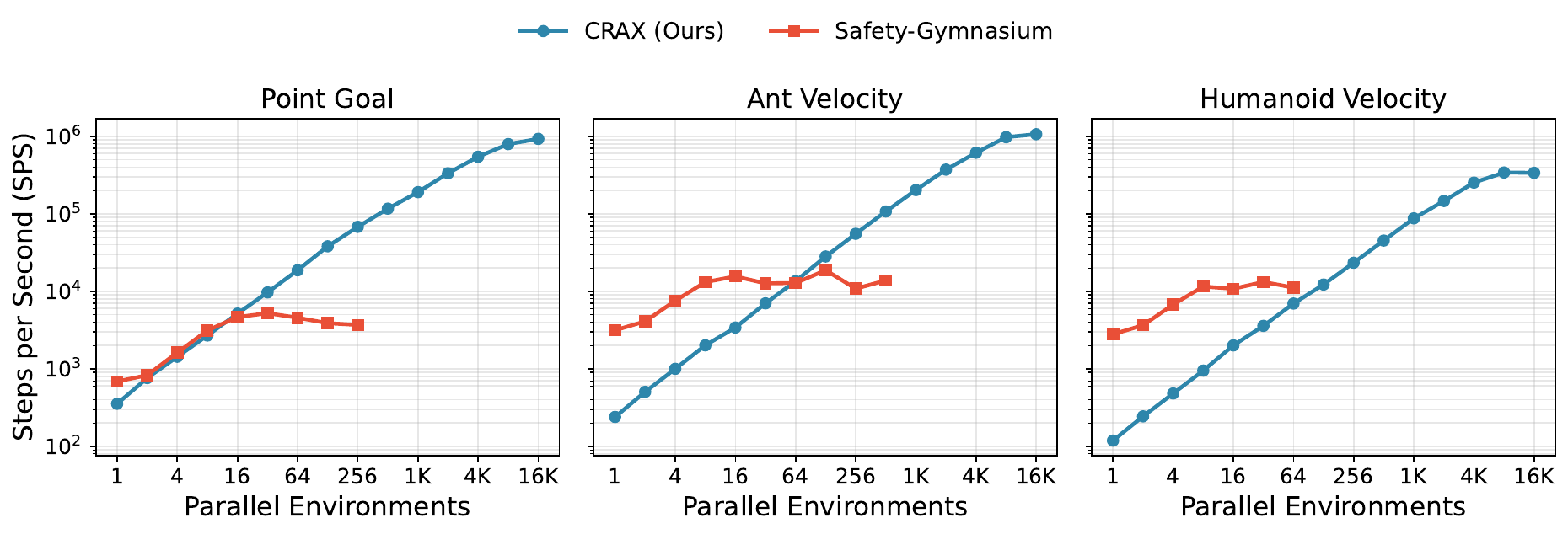}
    \caption{\textbf{Simulation throughput} of \acronym~and Safety-Gymnasium against the number of parallel environments on identical hardware. Safety-Gymnasium curves end where further scaling failed.}
    \label{fig:throughput}
\end{figure}

Simulation throughput is a core bottleneck in \saferl (\cref{sec:intro}). We compare \acronym~with Safety-Gymnasium (SG), the most widely used benchmark for continuous-control \saferl, on three tasks that exist in both, \texttt{Point Goal}, \texttt{Ant Velocity}, and \texttt{Humanoid Velocity} at Level~1, measuring steps per second (SPS) on identical hardware as the number of parallel environments grows.
\cref{fig:throughput} shows that SG is faster with few environments but saturates at 3K-20K SPS. Scaling beyond 256 environments exhausted the CPU memory on our hardware.
\acronym~keeps scaling to thousands of environments and peaks at roughly 340K SPS on \texttt{Humanoid Velocity} and around \textbf{1M SPS} on the other two tasks, which is 25 to 180 times the peak of SG.
To put this in concrete terms, our full evaluation suite (algorithms $\times$ environments $\times$ difficulty levels $\times$ seeds) amounts to hundreds of training runs and trillions of environment steps.
On a single H100 GPU, \acronym~completes this in 2 weeks. The equivalent run on Safety-Gymnasium would take close to a year.
This case study concerns throughput alone, and \cref{app:sg_fidelity} compares the two benchmarks on learning behavior.

\subsection{Reliable Assessment}
\label{sec:reliable_assessment}

\begin{wrapfigure}{r}{0.5\linewidth}
    \vspace{-40pt}
    \centering
    \includegraphics[width=\linewidth]{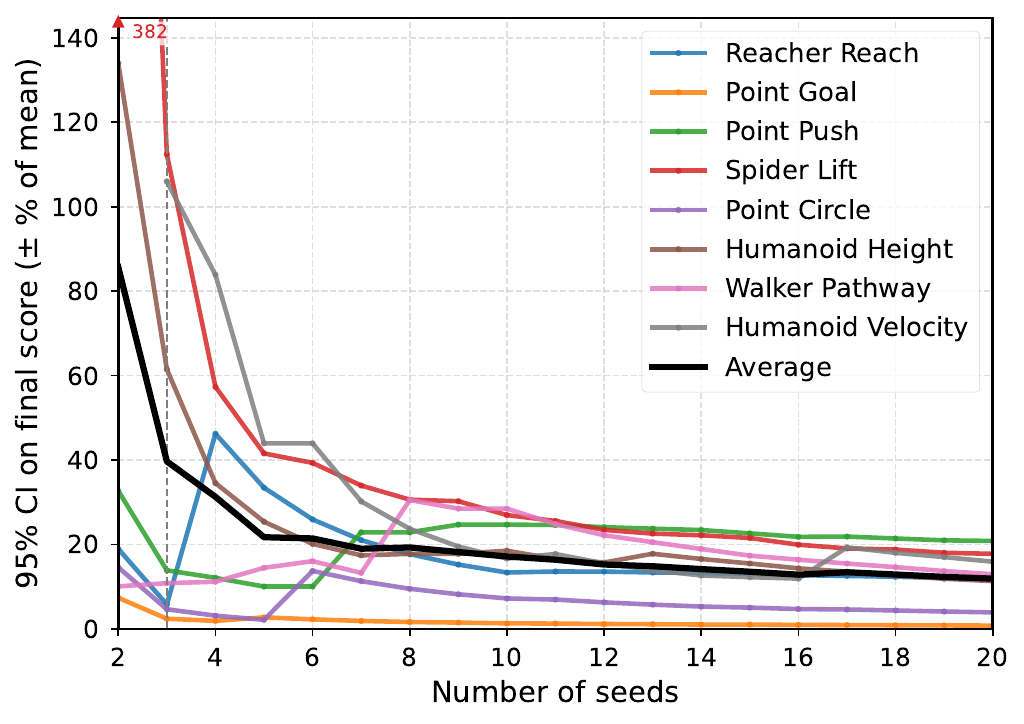}
    \vspace{-16pt}
    \caption{Half-width of the 95\% CI on the final score, as a percentage of the across-seed mean, against the number of seeds on Level~1. Spider Lift reaches 382\% at two seeds and is clipped.}
    \label{fig:seed_variance_main}
    \vspace{-10pt}
\end{wrapfigure}

Fast simulation matters most when it changes how methods are evaluated.
We trained every baseline with 20 seeds on all eight Level~1 environments and measured how the 95\% confidence interval of the final score narrows as seeds are added (\cref{fig:seed_variance_main}).
Relative to the mean, the interval shrinks from about 90\% at two seeds to 40\% at three and 22\% at five, after which gains are small.
Few seeds can also mislead, as on \texttt{Reacher Reach} and \texttt{Point Push} the interval at three seeds is narrower than at 20, because the first runs happened to agree.
Even at 20 seeds, the hardest environments retain intervals of about 30\%, since the agent occasionally fails to learn a useful policy (\cref{app:seed_variance}).

The same throughput makes hyperparameter sweeps affordable (\cref{app:sweeps}).
Cost stays close to the bound across most configurations of PPOLag, PPOPID, and FOCOPS, so their safety is robust to these choices, whereas reward depends strongly on the shared RL hyperparameters and much less on the safety-specific ones.
Lowering the PPO clip $\epsilon$ from 0.3 to 0.1 alone raises the reward of PPOLag, PPOPID, and P3O on \texttt{Point Push} from 25--34 to about 81, which overtakes FOCOPS.

\section{Conclusions}
We introduced \acronym, a hardware-accelerated \saferl benchmark built on MJX that reaches two orders of magnitude higher throughput than Safety-Gymnasium, making evaluations practical that are unaffordable on CPU-bound benchmarks.
Its eight tasks span eight agent morphologies and three difficulty levels, and agents can learn from state or from egocentric and allocentric camera images, with both on- and off-policy methods.
Using this throughput, we benchmarked seven \saferl algorithms, quantified how many seeds reliable comparisons require, ran broad hyperparameter sweeps, and measured how tightening the cost bound trades off against reward.
We found that curriculum learning and safety transfer help on some of the hardest tasks, whereas learning from pixels lowers both reward and constraint satisfaction, with most tasks unsolved from pixels.
We hope \acronym~makes such large-scale studies routine for the \saferl community.

\section{Limitations and Future Work}
\label{sec:limitations}
As of the time of writing, MJX does not yet support all features available in the CPU-based MuJoCo, such as certain rigid-body collision types. This restricts the range of scenarios that can currently be expressed within \acronym. In addition, we evaluate safety exclusively through expected cumulative cost constraints, leaving other safety formulations unexplored. Finally, our evaluation does not include model-based and multi-agent safe RL approaches, which remain as future work directions.

\clearpage



\bibliographystyle{iclr2027_conference}
\bibliography{references}
\newpage
\crefalias{section}{appendix}
\appendix

\section{Environment Descriptions}
\label{app:suites}

\subsection{Task Descriptions}
\label{sec:task-descriptions}

\cref{tab:suites} lists which agents each suite accepts, and \cref{tab:difficulty} gives the parameter varied across its difficulty levels.

\begin{figure}[!t]
    \centering
    \begin{subfigure}[b]{0.45\textwidth}
        \includegraphics[width=\textwidth]{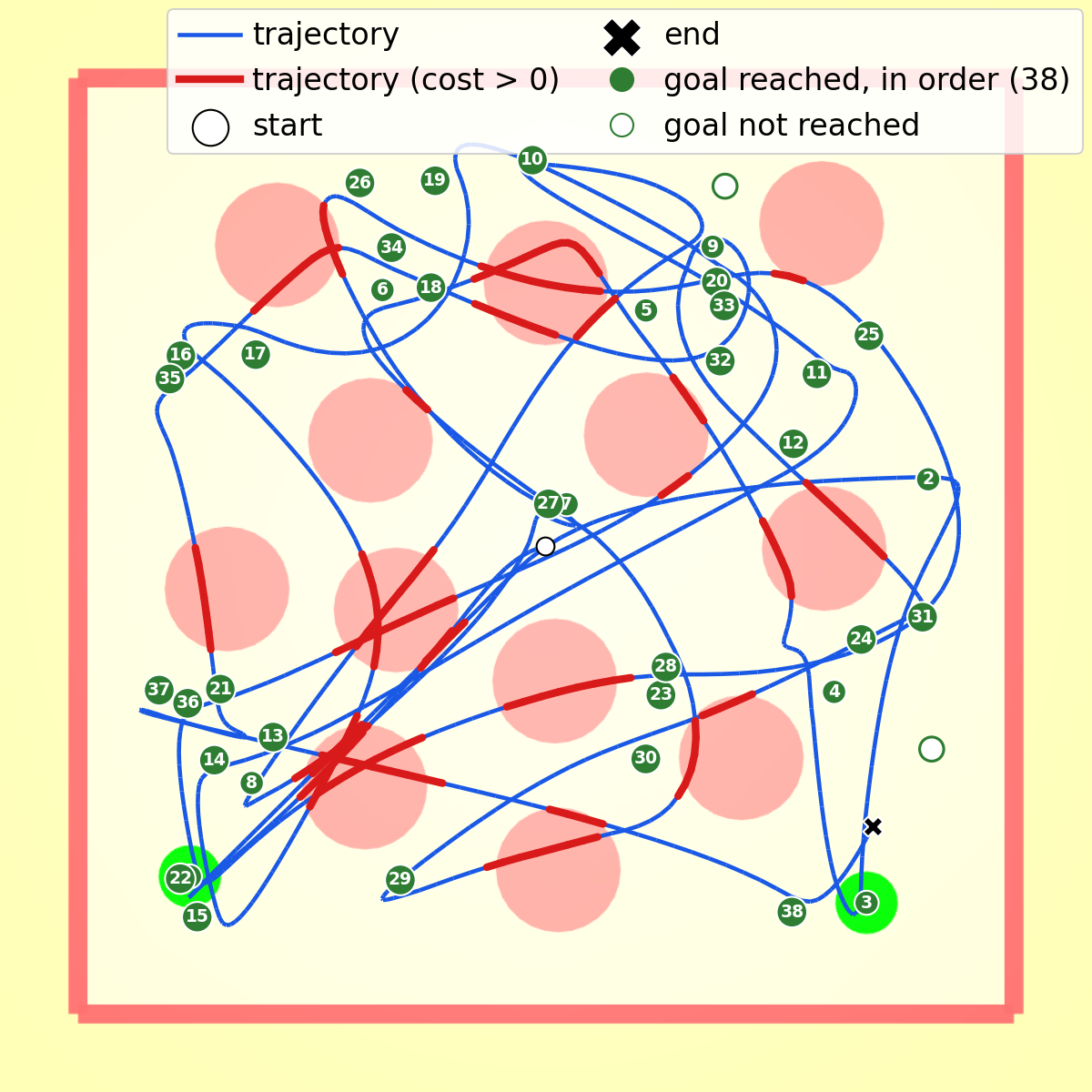}
        \caption{\texttt{Point Goal}, PPO}
    \end{subfigure}
    \hspace{0.02\textwidth}
    \begin{subfigure}[b]{0.45\textwidth}
        \includegraphics[width=\textwidth]{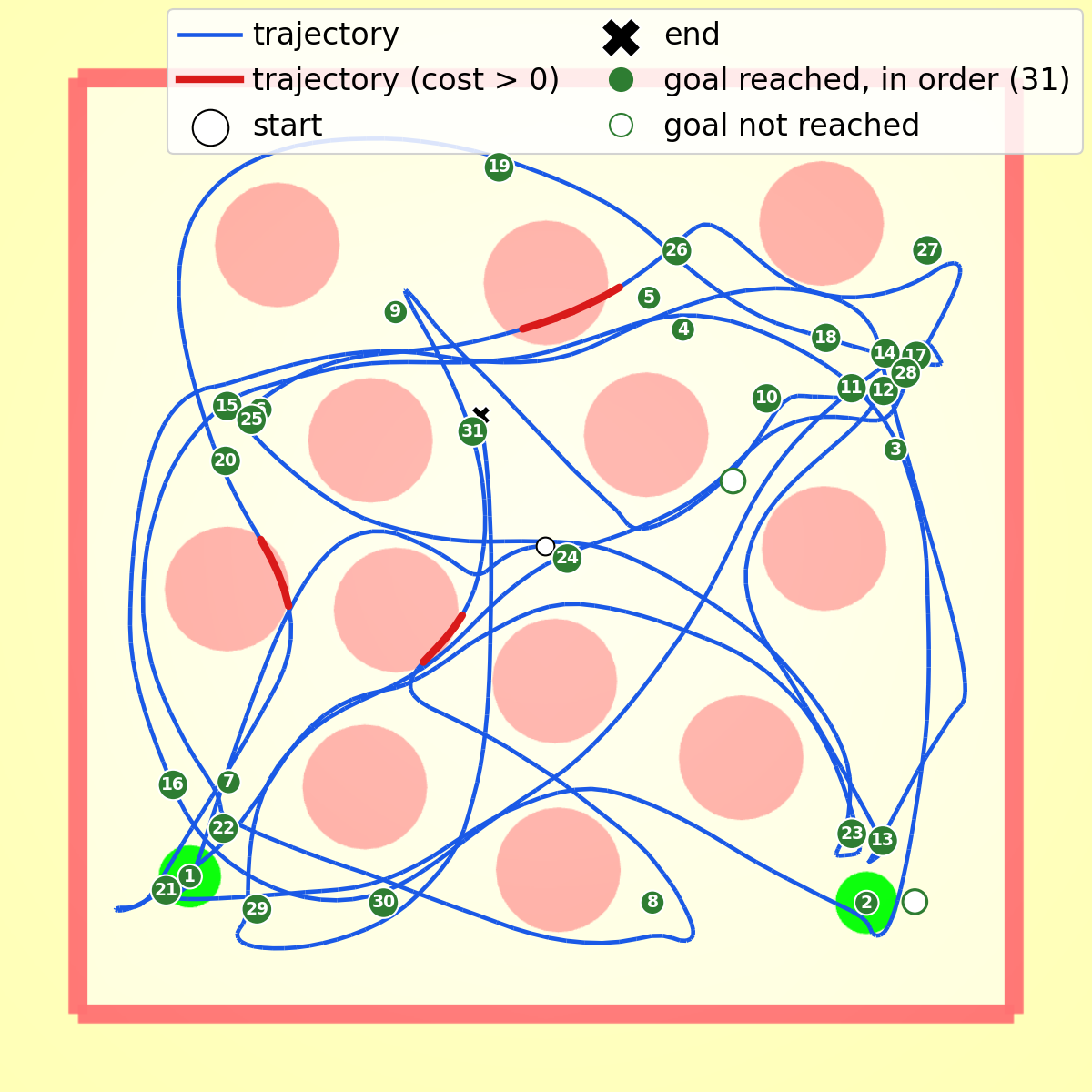}
        \caption{\texttt{Point Goal}, CRPO}
    \end{subfigure}
    \\[0.5em]
    \begin{subfigure}[b]{0.45\textwidth}
        \includegraphics[width=\textwidth]{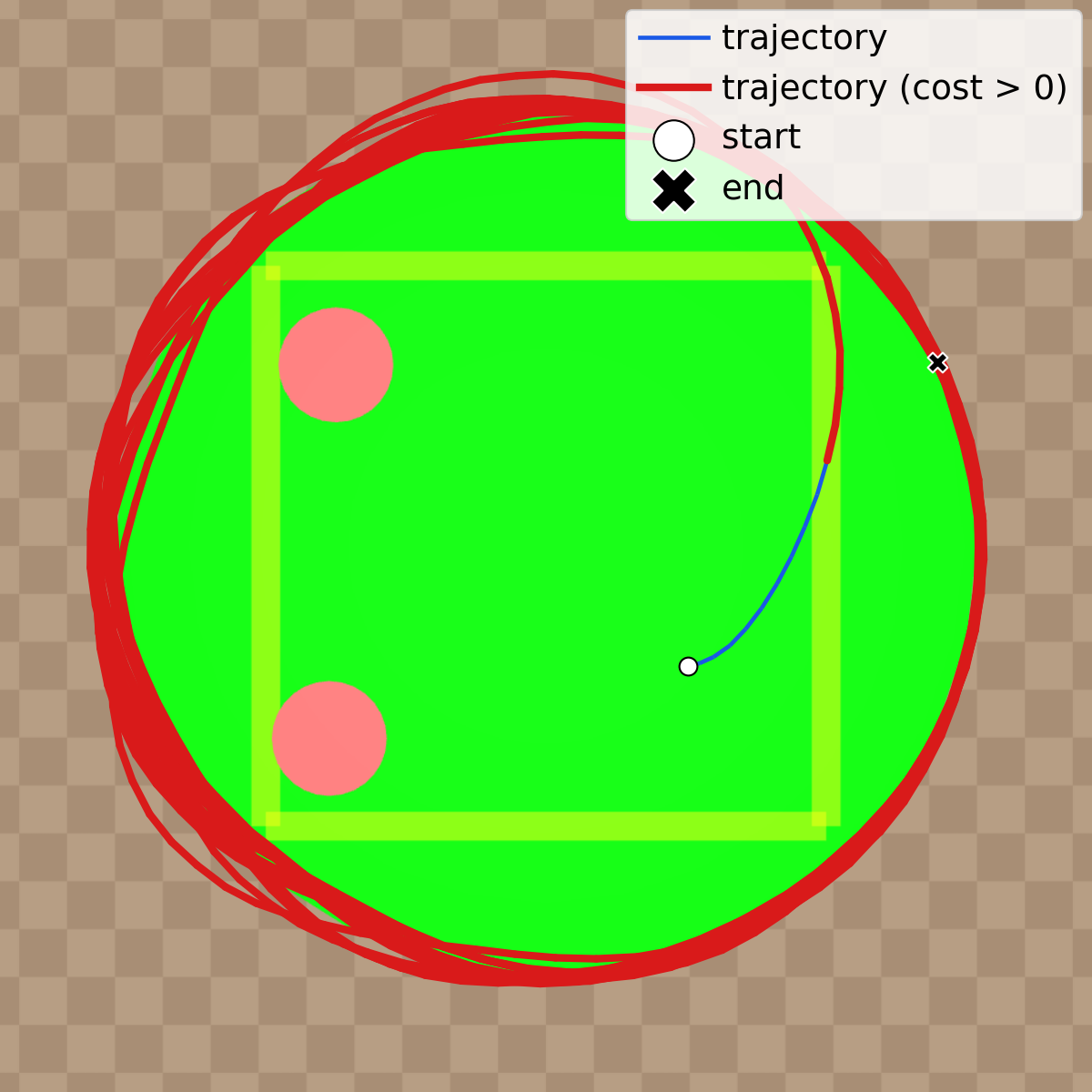}
        \caption{\texttt{Point Circle}, PPO}
    \end{subfigure}
    \hspace{0.02\textwidth}
    \begin{subfigure}[b]{0.45\textwidth}
        \includegraphics[width=\textwidth]{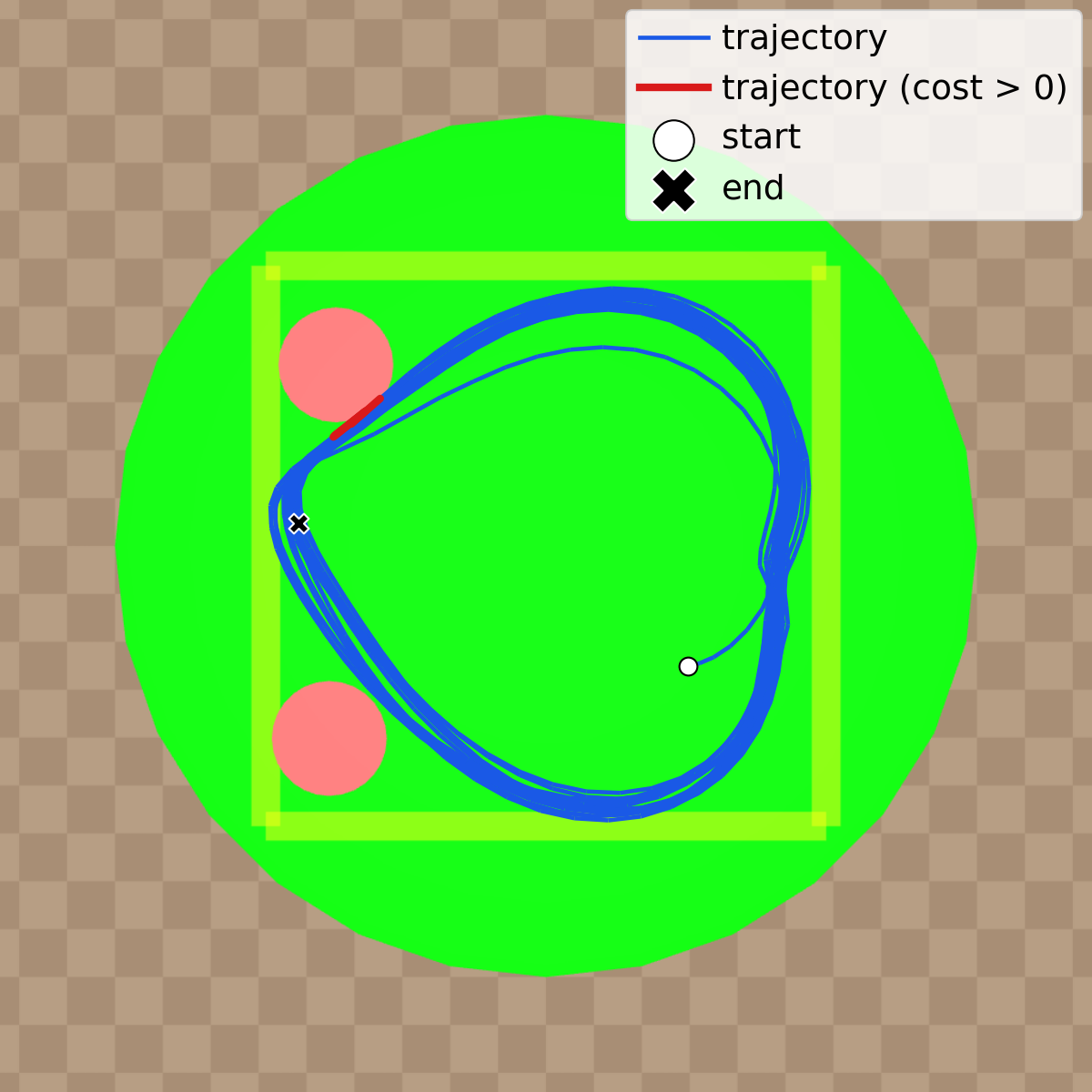}
        \caption{\texttt{Point Circle}, CRPO}
    \end{subfigure}
    \caption{Point agent trajectories over a single episode of \texttt{Point Goal} at Level~1 (top) and \texttt{Point Circle} at Level~3 (bottom). Red segments mark timesteps with nonzero cost. In \texttt{Point Goal}, PPO cuts straight through hazards on its way between goals, whereas CRPO routes around most of them and still reaches 31 goals compared to 33 for PPO. In \texttt{Point Circle}, PPO orbits at the target radius outside the arena boundary for the entire episode. CRPO instead follows a tighter loop that stays inside the boundary and only briefly grazes one hazard.}
    \label{fig:trajectories}
\end{figure}

\begin{figure}[!t]
    \centering
    \begin{subfigure}[b]{0.325\textwidth}
        \includegraphics[width=\textwidth]{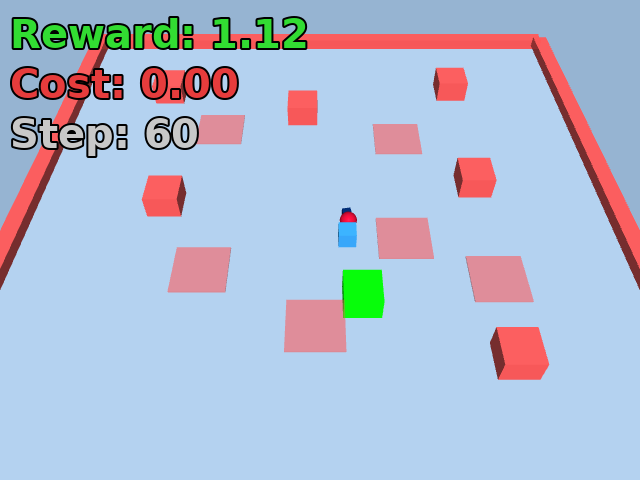}
    \end{subfigure}
    \hfill
    \begin{subfigure}[b]{0.325\textwidth}
        \includegraphics[width=\textwidth]{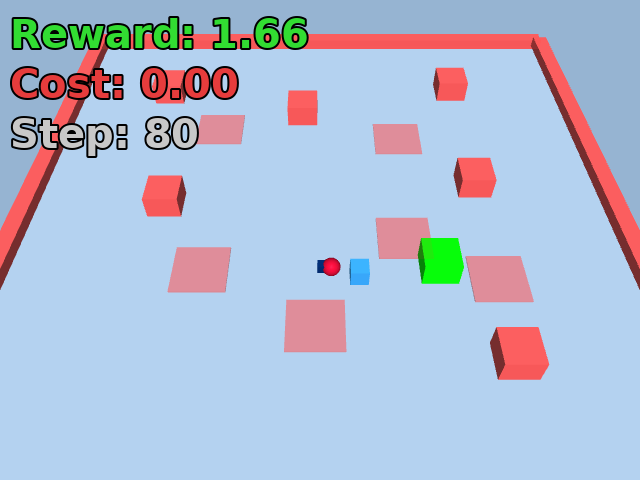}
    \end{subfigure}
    \hfill
    \begin{subfigure}[b]{0.325\textwidth}
        \includegraphics[width=\textwidth]{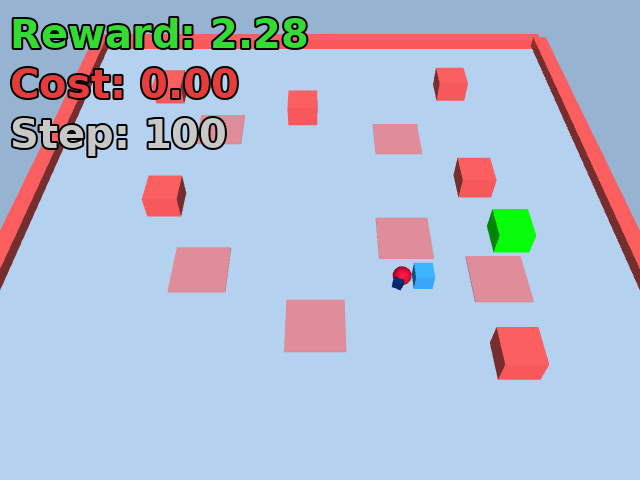}
    \end{subfigure}
    \caption{In levels~2 and~3 of \texttt{Push}, the goal that the agent must push the block into moves at a fixed velocity. To succeed, the agent ought to anticipate the goal's trajectory while avoiding hazard zones.}
    \label{fig:block_push}
\end{figure}

\subsubsection{Navigation Tasks}
In the first three tasks, a mobile agent moves through an arena containing hazards. They are compatible with Point, Ant, Humanoid, and Spider. \cref{fig:trajectories} contrasts the trajectories of an unconstrained (PPO) and a constrained (CRPO) policy in \texttt{Goal} and \texttt{Circle}.

\paragraph{Goal.}
Navigate to goal regions while avoiding hazards scattered throughout an arena. When the agent reaches a goal, it respawns at a new random location. Hazards may be collidable (blocking, with contact-based cost) or non-collidable (pass-through, with proximity-based cost). This is a continuous task with no terminal success state.


\paragraph{Circle.}
Maintain a circular orbit at a target radius around a fixed center point. The agent is rewarded for tangential velocity and penalized for deviating from the target radius. Higher levels shrink the arena boundary and add cylindrical hazards inside it.

\paragraph{Push.}
Push a movable block into goal regions while avoiding hazards. Unlike Goal, the \emph{block} rather than the agent, must reach the goal, so the agent has to coordinate approaching the block with pushing it in the right direction. Only agent-hazard interactions incur cost, and the block passes through hazards freely. At Levels 2 and 3, the goal moves at a fixed velocity, so the agent has to anticipate its trajectory rather than steer toward its current position (\cref{fig:block_push}).

\subsubsection{Locomotion Tasks}
In these tasks, an agent must move forward while respecting a constraint on its speed, posture, gait, or foot placement.

\paragraph{Velocity.}
Move forward while keeping the speed below a threshold, which is set relative to the speed an unconstrained agent of the same morphology reaches (\cref{tab:velocity_thresholds}).

\paragraph{Height.}
Move forward while staying below a maximum height threshold, simulating a low ceiling. Cost increases smoothly as height exceeds the threshold, encouraging the agent to crouch. The Humanoid implementation is based on the \texttt{HumanoidStandup} environment with an added forward locomotion reward.

\begin{figure}[!t]
    \centering
    \begin{subfigure}[b]{0.245\textwidth}
        \includegraphics[width=\textwidth]{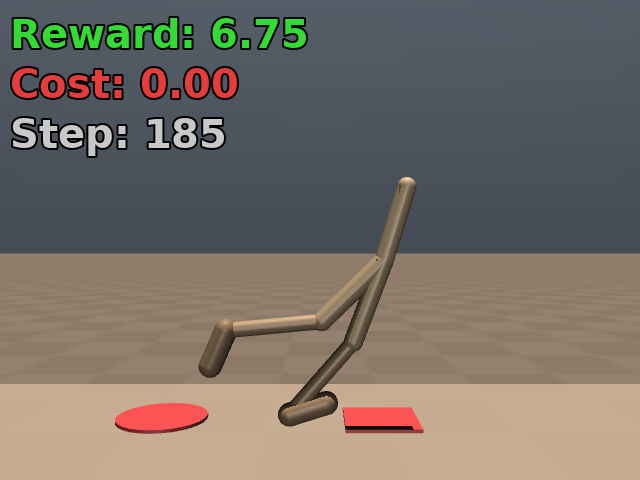}
    \end{subfigure}
    \hfill
    \begin{subfigure}[b]{0.245\textwidth}
        \includegraphics[width=\textwidth]{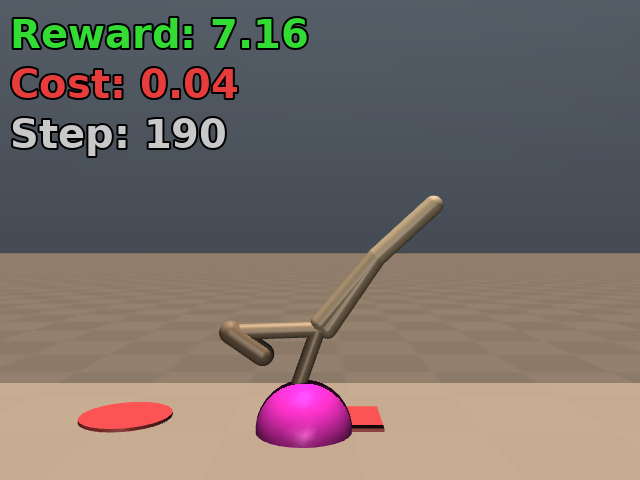}
    \end{subfigure}
    \hfill
    \begin{subfigure}[b]{0.245\textwidth}
        \includegraphics[width=\textwidth]{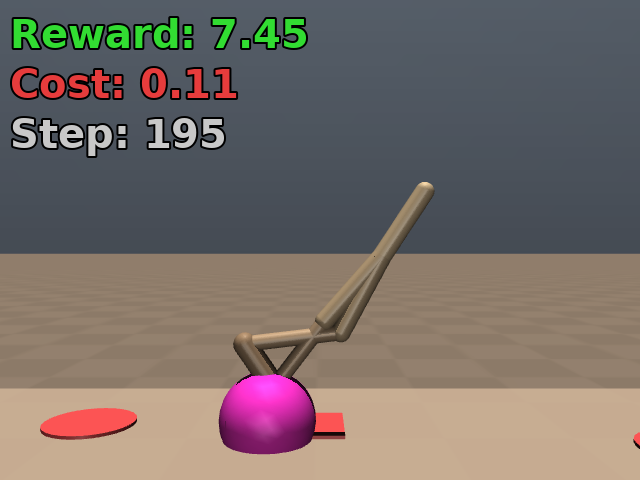}
    \end{subfigure}
    \hfill
    \begin{subfigure}[b]{0.245\textwidth}
        \includegraphics[width=\textwidth]{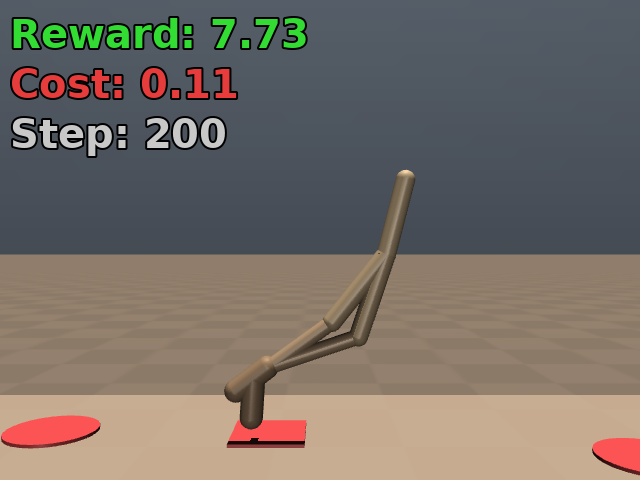}
    \end{subfigure}
    \caption{The \texttt{Pathway} agent incurs a per-step cost when its foot contacts a hazard, scaled by the penetration depth into the hazard region.}
    \label{fig:safe_pathway}
\end{figure}

\paragraph{Pathway.}
Traverse a path with non-collidable hazard zones placed along its route. Cost is incurred when feet step inside hazard regions and scales with penetration depth, so grazing the edge of a zone is cheaper than fully landing in it
(\cref{fig:safe_pathway}). Hazards are randomly placed with varying gaps and lateral offsets, so the agent has to time its steps carefully. Falling incurs a terminal cost.

\paragraph{Lift.}
Walk forward while keeping specified legs off the ground, forcing unusual gaits. Cost is incurred each timestep a restricted foot contacts the floor.

\subsubsection{Reaching Tasks}

\paragraph{Reach.}
Reach a randomly placed target while avoiding flat hazards scattered on the workspace. Currently, the only compatible agent is Reacher: a planar 2-link robotic arm. Cost is incurred when any part of the arm intersects a hazard, detected by sampling discrete points along both links. Reward emphasizes proximity to the target with a bonus for reaching it.

\subsection{Reward and Cost Functions}
\label{sec:technical-reference}

Throughout, $r_t$ and $c_t$ denote the reward and cost at timestep $t$, $d(\cdot,\cdot)$ Euclidean distance, and $\mathbf{1}[\cdot]$ the indicator function. Rewards are assembled from four recurring terms: forward progress $v_{\text{forward}} = (x_t - x_{t-1})/\Delta t$
along the agent's forward axis, a constant survival bonus $r_{\text{healthy}}$ paid while the agent remains in a healthy state, a sparse goal reward with an optional dense shaping term $d_{t-1} - d_t$, and a quadratic control penalty on the action vector $\mathbf{a}$. Each suite instantiates a subset of these with its own weights (\cref{tab:reward_functions}).

\begin{table}[h]
\caption{Per-timestep reward. Superscripts \text{bg} and \text{ab} in \texttt{Push} denote the block-goal and agent-block distances.}
\label{tab:reward_functions}
\centering
\small
\begin{tabular}{ll}
\toprule
\textbf{Task} & \textbf{Reward} \\
\midrule
Goal & $r_t = \alpha_{\text{dist}} (d_{t-1} - d_t) + \alpha_{\text{goal}} n_{\text{reached}}$ \\
Circle & $r_t = \alpha\, v_{\text{tan}} \,(1 + |r_{\text{actual}} - r_{\text{target}}|)^{-1}$ \\
Push & $r_t = \alpha_{\text{bg}} (d^{\text{bg}}_{t-1} - d^{\text{bg}}_t) + \alpha_{\text{ab}} (d^{\text{ab}}_{t-1} - d^{\text{ab}}_t) + \alpha_{\text{goal}} \mathbf{1}[\text{reached}]$ \\
Velocity & $r_t = \alpha\, r_{\text{base}}$ \\
Height & $r_t = v_{\text{forward}} + 1.0 - 0.01\|\mathbf{a}\|^2$ \\
Pathway & $r_t = \alpha (v_{\text{forward}} + r_{\text{healthy}})$ \\
Reach & $r_t = \alpha (1 - d_t/d_{\max})^\gamma + r_b \mathbf{1}[d_t < \epsilon]$ \\
Lift & $r_t = v_{\text{forward}} + r_{\text{healthy}} - 0.5\|\mathbf{a}\|^2$ \\
\bottomrule
\end{tabular}
\end{table}

The cost formulations follow the five constraint types introduced in \cref{sec:rewards_costs_constraints}. \cref{tab:cost_functions} gives their exact form, where $h_i$ indexes hazards, $d_i$ and $r_i$ are the distance to hazard $i$ and its radius, $\tau$ a velocity threshold, and $\mathcal{F}_{\text{restr}}$ the set of restricted feet.

\begin{table}[h]
\caption{Per-timestep cost. \texttt{Goal} and \texttt{Push} sum contact and proximity terms over all hazards, using the cylinder or cube form depending on hazard geometry. \texttt{Velocity} uses the binary form by default.}
\label{tab:cost_functions}
\centering
\small
\begin{tabular}{lll}
\toprule
\textbf{Task} & \textbf{Type} & \textbf{Cost} \\
\midrule
Goal, Push & Contact & $c_t = \sum_i c_{\text{col}} \mathbf{1}[\text{contact}(a, h_i)]$ \\
 & Proximity (cyl.) & $c_t = \sum_i c_{\text{prox}} \max(0, 1 - d_i/r_i)$ \\
 & Proximity (cube) & $c_t = \sum_i c_{\text{prox}} \mathbf{1}[|dx_i| \leq s \land |dy_i| \leq s]$ \\
\midrule
Circle & Proximity + boundary & $c_t = c_{\text{hazard}} + c_b \mathbf{1}[\text{outside boundary}]$ \\
\midrule
Velocity & Binary & $c_t = w \mathbf{1}[v_t > \tau]$ \\
 & Hinge & $c_t = w \max(0, v_t - \tau)$ \\
Height & Soft hinge & $c_t = w \max(0, h_t - h_{\max}) / \delta$ \\
\midrule
Pathway & Quadratic & $c_t = \beta \sum_i \max_f [\max(0, 1 - d_f^{(i)}/r_i)]^2 + c_{\text{term}} \mathbf{1}[\text{fell}]$ \\
Reach & Binary & $c_t = \beta \sum_i \mathbf{1}[\text{arm} \cap h_i]$ \\
Lift & Binary & $c_t = \beta \sum_{f \in \mathcal{F}_{\text{restr}}} \mathbf{1}[\text{contact}(f, \text{floor})]$ \\
\bottomrule
\end{tabular}
\end{table}


\begin{table}[h]
\caption{Parameter varied across difficulty levels in each suite. Level 3 of \texttt{Goal}
mixes cylindrical and cubic hazards, while Levels 1 and 2 use cylinders only. \texttt{Circle}
Level 1 bounds the $x$ axis only; Levels 2 and 3 bound both axes. 
}
\label{tab:difficulty}
\centering
\small
\begin{tabular}{llccc}
\toprule
\textbf{Suite} & \textbf{Parameter} & \textbf{Level 1} & \textbf{Level 2} & \textbf{Level 3} \\
\midrule
Goal       & Hazards (proximity + collidable) & 12 + 0 & 8 + 8 & 12 + 8 \\
Circle     & Boundary extent / hazards & $\pm1.125$ / 0 & $\pm1.05$ / 1 & $\pm0.975$ / 2 \\
Push       & Goal velocity & 0.0 & 0.3 & 0.6 \\
Reach      & Hazards in workspace & 4 & 7 & 10 \\
Pathway    & Max gap between hazards (m) & 6.0 & 4.0 & 2.0 \\
Height     & Max CoM height (m) & 1.30 & 1.15 & 1.00 \\
Lift       & Restricted feet & 2 (diagonal) & 3 (tripod) & 4 (center only) \\
Velocity   & Speed limit ($\times$ baseline) & 1.0 & 0.75 & 0.5 \\
\bottomrule
\end{tabular}
\end{table}

\begin{table}[h]
\caption{Speed limits in the \texttt{Velocity} task, in m/s.}
\label{tab:velocity_thresholds}
\centering
\small
\begin{tabular}{lccccc}
\toprule
& \textbf{Ant} & \textbf{HalfCheetah} & \textbf{Hopper} & \textbf{Humanoid} & \textbf{Walker2d} \\
\midrule
Level 1 ($1.0\times$)  & 2.62 & 3.21 & 0.74 & 1.41 & 2.34 \\
Level 2 ($0.75\times$) & 1.97 & 2.41 & 0.56 & 1.06 & 1.76 \\
Level 3 ($0.5\times$)  & 1.31 & 1.60 & 0.37 & 0.71 & 1.17 \\
\bottomrule
\end{tabular}
\end{table}

\subsection{Agents}
\label{app:agents}

CRAX provides eight agent morphologies (\cref{tab:agents}), split into three groups. The 3D agents (Point, Ant, Humanoid, Spider) move freely in the plane and are the ones used in the navigation suite (\texttt{Goal}, \texttt{Push}, \texttt{Circle}), where they perceive goals and hazards through lidar and compass sensors rather than through privileged positions. The 2D agents (HalfCheetah, Walker2d, Hopper) are the standard planar locomotion models, restricted to the sagittal plane and used in the \texttt{Velocity} and \texttt{Pathway} tasks. Reacher is a fixed-base 2-link arm, used only in the \texttt{Reach} task. \cref{tab:suites} lists which tasks are compatible with which agent.

Agents that can lose balance (Ant, Humanoid, Spider, Walker2d, Hopper) terminate the episode when the torso leaves a healthy height or angle range, following the standard MuJoCo convention. Point terminates on the same mechanism, though for it the range acts as a sanity bound against the sphere being launched off the ground rather than as a fall condition. HalfCheetah and Reacher never terminate early. Observation vectors consist of joint positions and velocities with the global root coordinates removed, extended per agent with contact forces (Ant), body inertias and actuator forces (Humanoid), or lidar and compass readings (Point). The exact composition is defined in the released environment code.

\begin{table}[!t]
\caption{Agent morphologies. \textbf{Act} and \textbf{Obs} give the action and observation
dimensionality, \textbf{DoF} the degrees of freedom including the root, and $\Delta t$ the
control timestep. \textbf{Healthy range} gives the torso height (and where applicable the
torso angle) outside which the episode terminates.}
\label{tab:agents}
\centering
\small
\begin{tabular}{llccccll}
\toprule
\textbf{Agent} & \textbf{Type} & \textbf{Act} & \textbf{Obs} & \textbf{DoF} & \textbf{Bodies} & \textbf{$\Delta t$ (s)} & \textbf{Healthy range} \\
\midrule
Point       & Holonomic sphere & 2  & 62  & 3  & 1  & 0.008 & 0.05--0.3 \\
Ant         & 3D quadruped     & 8  & 27  & 15 & 13 & 0.05  & 0.2--1.0 \\
Humanoid    & 3D bipedal       & 17 & 376 & 23 & 13 & 0.015 & 1.0--2.0 \\
Spider      & 3D hexapod       & 12 & 35  & 19 & 13 & 0.05  & 0.2--1.0 \\
\midrule
HalfCheetah & 2D planar runner & 6  & 18  & 9  & 8  & 0.05  & no termination \\
Walker2d    & 2D bipedal       & 6  & 17  & 9  & 7  & 0.008 & 0.8--2.0, $|\theta| < 1.0$ \\
Hopper      & 2D one-legged    & 3  & 11  & 6  & 4  & 0.008 & $> 0.7$, $|\theta| < 0.2$ \\
\midrule
Reacher     & 2-link arm       & 2  & 11  & 4  & 3  & 0.02  & no termination \\
\bottomrule
\end{tabular}
\end{table}

\subsection{Difficulty Levels}
\label{app:difficulty}

\begin{figure}[tbp]
    \centering
    \begin{tabular}{c c c c}
        & \textbf{Level 1} & \textbf{Level 2} & \textbf{Level 3} \\[3pt]

        \rotatebox{90}{\texttt{Goal}} &
        \includegraphics[width=0.22\textwidth]{figures/safe_point_goal/safe_point_goal_level_1.png} &
        \includegraphics[width=0.22\textwidth]{figures/safe_point_goal/safe_point_goal_level_2.png} &
        \includegraphics[width=0.22\textwidth]{figures/safe_point_goal/safe_point_goal_level_3.png} \\[6pt]

        \rotatebox{90}{\texttt{Reach}} &
        \includegraphics[width=0.22\textwidth]{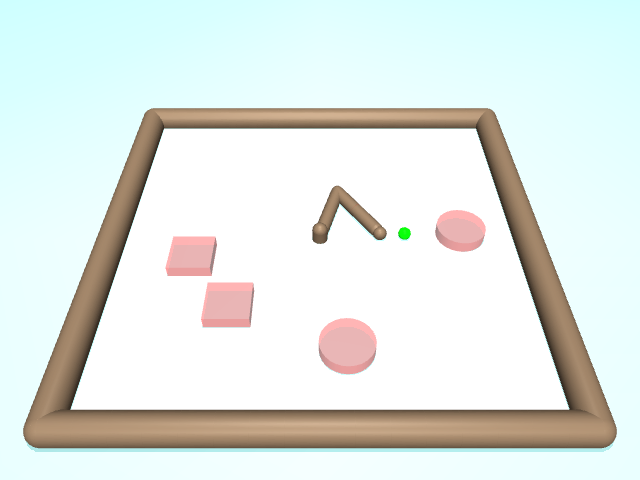} &
        \includegraphics[width=0.22\textwidth]{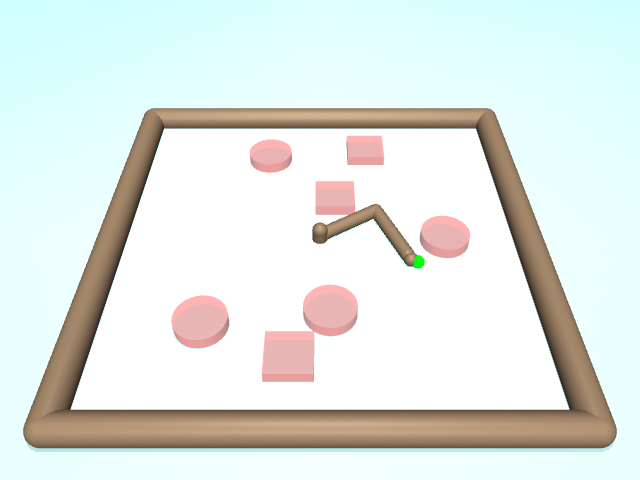} &
        \includegraphics[width=0.22\textwidth]{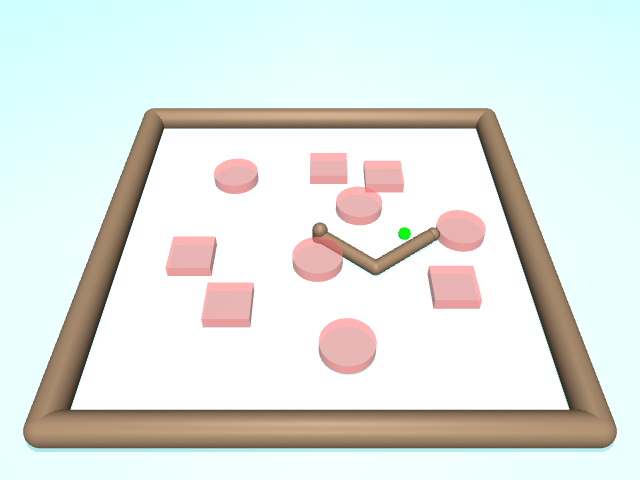} \\[6pt]

        \rotatebox{90}{\texttt{Pathway}} &
        \includegraphics[width=0.22\textwidth]{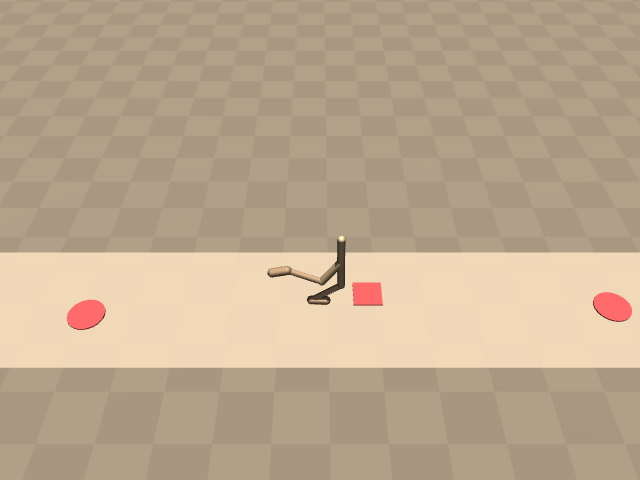} &
        \includegraphics[width=0.22\textwidth]{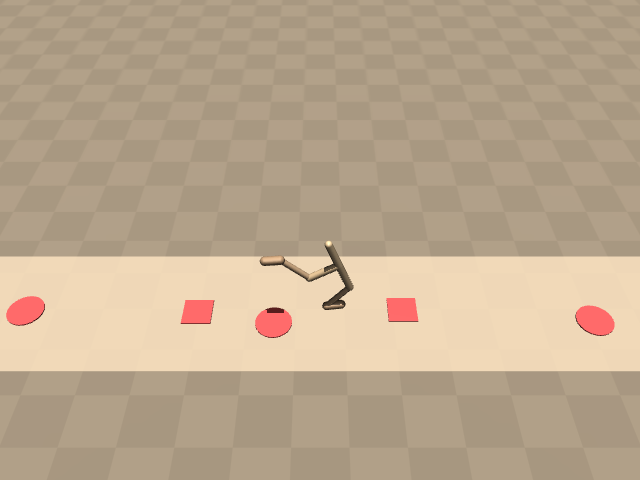} &
        \includegraphics[width=0.22\textwidth]{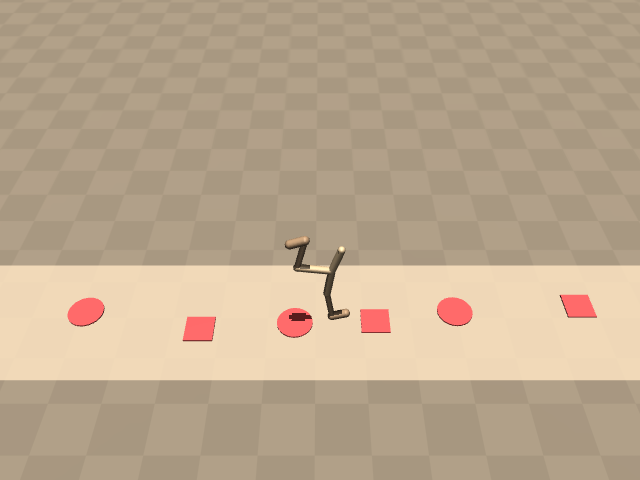} \\[6pt]

        \rotatebox{90}{\texttt{Lift}} &
        \includegraphics[width=0.22\textwidth]{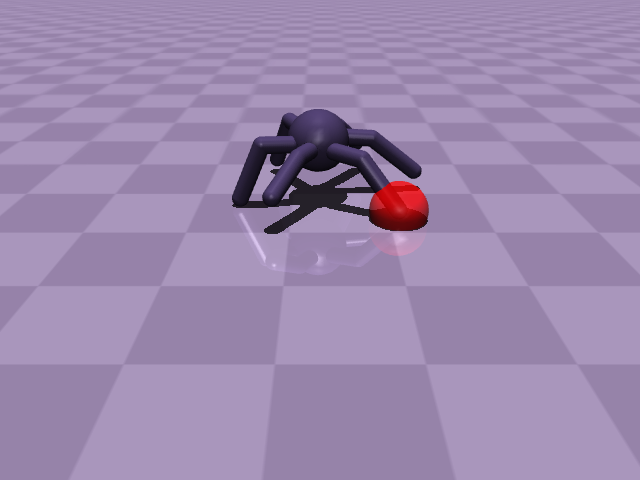} &
        \includegraphics[width=0.22\textwidth]{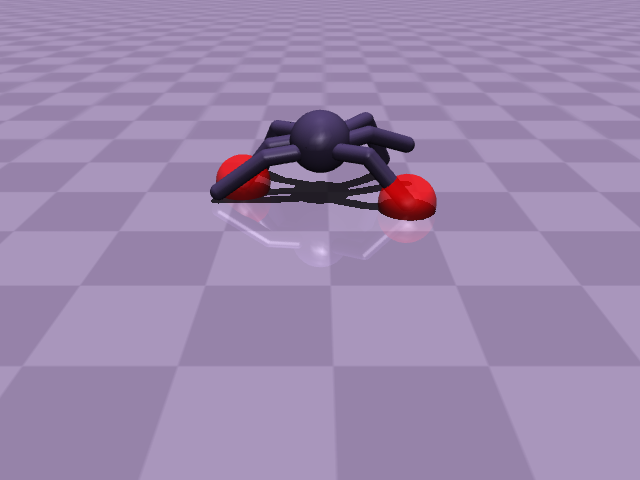} &
        \includegraphics[width=0.22\textwidth]{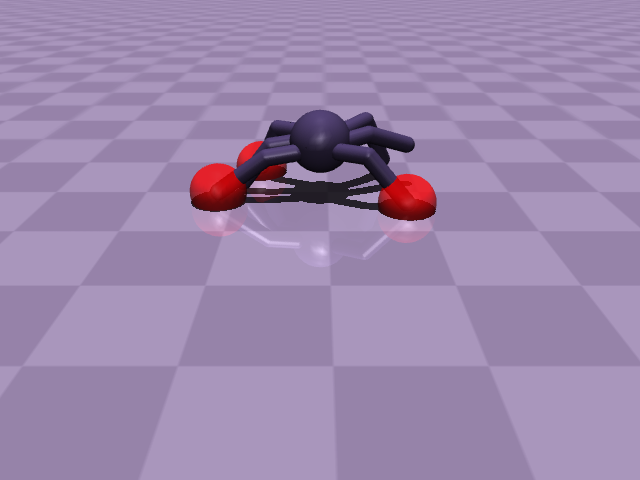} \\[6pt]

        \rotatebox{90}{\texttt{Circle}} &
        \includegraphics[width=0.22\textwidth]{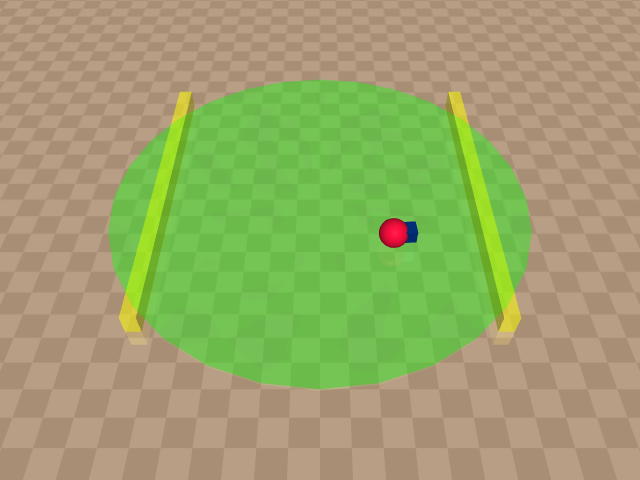} &
        \includegraphics[width=0.22\textwidth]{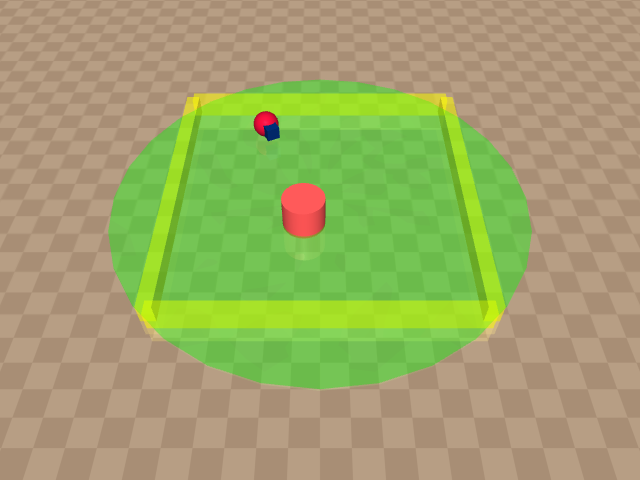} &
        \includegraphics[width=0.22\textwidth]{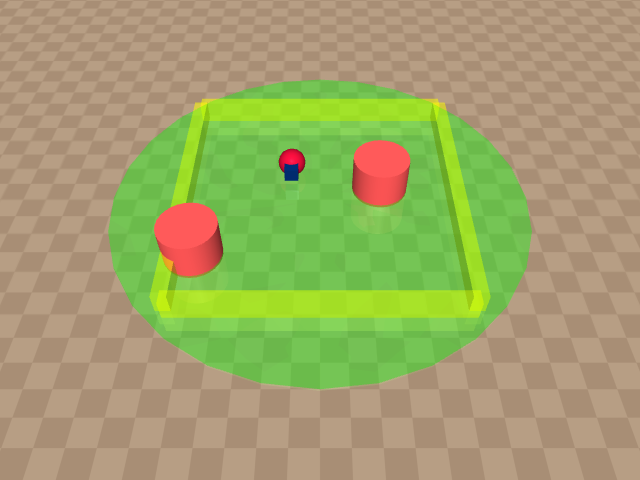} \\[6pt]

        \rotatebox{90}{\texttt{Height}} &
        \includegraphics[width=0.22\textwidth]{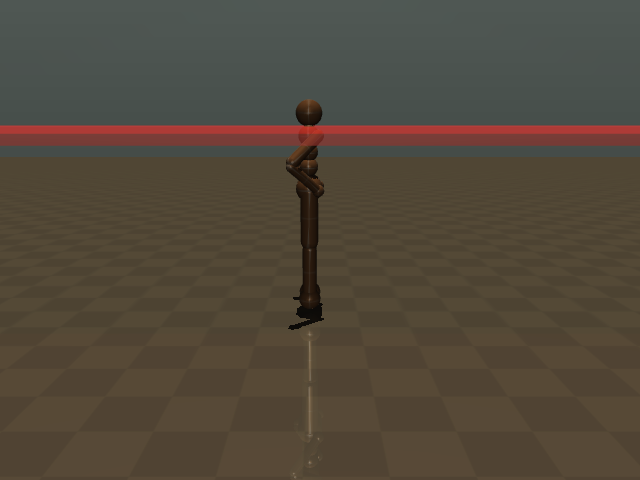} &
        \includegraphics[width=0.22\textwidth]{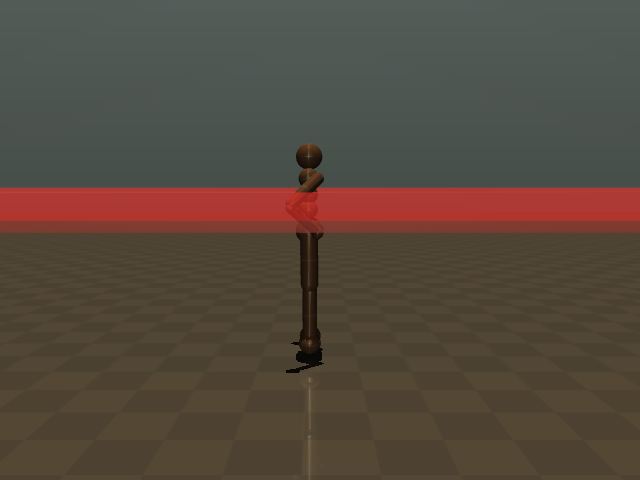} &
        \includegraphics[width=0.22\textwidth]{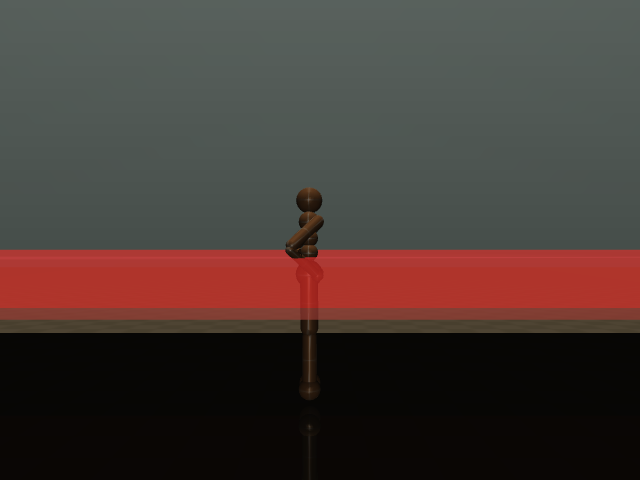} \\
    \end{tabular}

    \caption{Visual comparison of difficulty levels across environments. Increasing difficulty generally adds more hazards, tightens constraints, or reduces margins for error.}
    \label{fig:difficulty_grid}
\end{figure}

Each task varies a single dominant parameter across its three levels (\cref{tab:difficulty}). \cref{fig:difficulty_grid} shows the visual effect. The \texttt{Velocity} limit is set relative to the speed an unconstrained agent reaches on the same morphology, so the absolute threshold differs per agent (\cref{tab:velocity_thresholds}).

\section{Seed Variance}
\label{app:seed_variance}

Reinforcement learning results are notoriously sensitive to the number of seeds used to produce them~\citep{agarwal2021deep}. This matters especially for reliably benchmarking results. In this section, we therefore measure how sensitive our baseline results are on CRAX tasks, extending the summary in \cref{sec:reliable_assessment}. We run 20 seeds on all 8 environments for 500M steps each. As described in \cref{sec:computational_efficiency}, the JAX-enabled throughput makes this affordable.

\begin{figure}[!t]
    \centering
    \begin{subfigure}[b]{0.4\textwidth}
        \includegraphics[width=\textwidth]{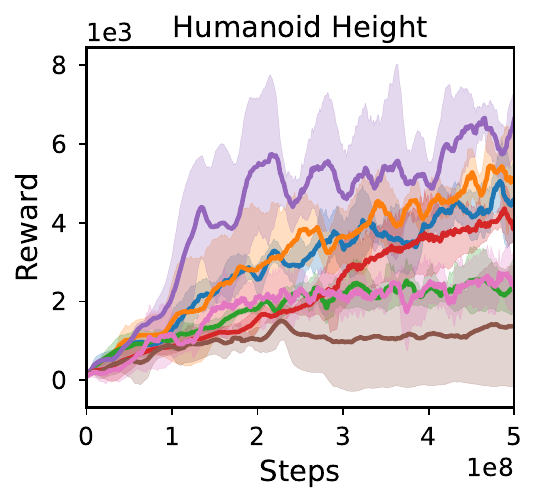}
        \caption{Training curves averaged over 3 seeds.}
        \label{fig:3_seeds}
    \end{subfigure}
    \hspace{3em}
    \begin{subfigure}[b]{0.4\textwidth}
        \includegraphics[width=\textwidth]{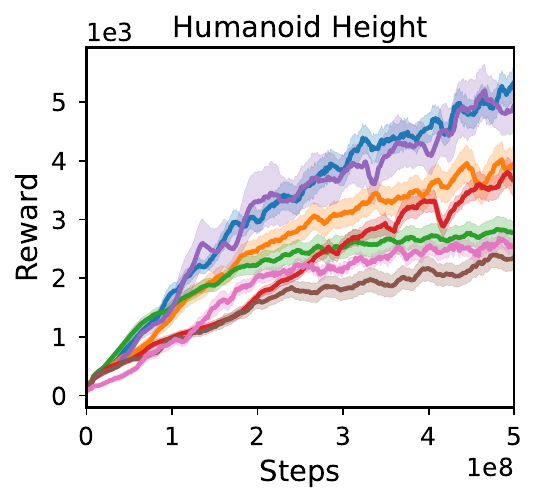}
        \caption{Training curves averaged over 20 seeds.}
        \label{fig:20_seeds}
    \end{subfigure}
    \caption{Training curves on Level~1 with 95\% confidence intervals over 3 seeds (\textbf{left}) and 20 seeds (\textbf{right}). The legend is omitted, since the point is the width of the bands rather than which method is which.}
    \label{fig:3_vs_20_seeds}
\end{figure}

\cref{fig:3_vs_20_seeds} shows what visual effect this has on the training curves. At 3 seeds, the confidence bands overlap across most of training, and the ordering between methods is not resolved. To quantify this, we take the half-width of the 95\% confidence interval around the across-seed mean of the final score and express it as a percentage of that mean, so tasks with different reward scales can be compared. Each value averages over the eight baselines and over both reward and cost. \cref{fig:seed_variance_main} shows how this metric changes as results with additional seeds are added.

Most of the gain comes early. Averaged over tasks, the interval narrows from about 90\% at 2 seeds to 40\% at 3 and then only to 22\% at 5. Since every additional seed costs the same as the first, the return on seeds past five is small on most tasks.

Low seed counts can also understate variance. On \texttt{Reacher Reach} and \texttt{Point Push} the interval is wider at 20 seeds than at 3, because three runs happened to agree and produced a tight estimate that the next 17 contradicted. A task can therefore look settled at small $n$ with nothing in the confidence interval to indicate otherwise.

How many seeds an environment needs, of course, depends on the learning algorithm. For instance, in our case, \texttt{Point Push}, \texttt{Spider Lift}, and \texttt{Walker Pathway} stay at around 30\% at 10 seeds, while \texttt{Point Goal} and \texttt{Point Circle} are only at around 5\%. We report ten seeds in the main paper, which narrows the interval to about 5\% on the easiest tasks and 30\% on the hardest.


\section{Efficiency and Safety During Training}
\label{app:sample-efficiency}

The results in \cref{sec:emperical_eval} report performance at a fixed budget of 500M environment steps, which says nothing about how quickly a method gets there or how much constraint violation it accumulates on the way. Two algorithms that end at the same reward and cost can differ substantially in how they spend the budget. We therefore report two additional metrics computed from the training curves of the same runs.

\textbf{Reward efficiency} (RewEff) is the area under the reward curve, normalized by the span of training and by the best final reward achieved on that task. For a policy $\pi$ trained over the interval $[T_0, T_1]$, let $\hat{R}^\pi(t)$ denote its evaluated episodic return after $t$ environment steps. Then
\begin{equation}
\text{RewEff}(\pi) = \frac{1}{\hat{R}^\star (T_1 - T_0)} \int_{T_0}^{T_1} \hat{R}^\pi(t)\, dt ,
\end{equation}
where $\hat{R}^\star$ is the largest final return observed on that task across the compared methods, with final return taken as the mean of $\hat{R}^\pi$ over the last 10\% of the curve. The unconstrained baseline is excluded from $\hat{R}^\star$, since normalizing by a method that ignores the cost budget would compress all constrained methods toward zero on the tasks where safety is expensive. The best method on each task is therefore pinned near 1 by construction, and RewEff should be read as a relative score within the compared set. Values fall in $[0, 1]$ in practice and are comparable across tasks. A method that reaches a good policy early scores close to the best final performer, while one that only catches up at the end of the budget
scores lower.

\textbf{Cumulative violation} (CumViol) measures how much a method exceeds the safety threshold $d$ over the course of training, rather than only at convergence. Writing $\hat{C}^\pi(t)$ for the evaluated episodic cost of $\pi$ after $t$ environment steps, we integrate the excess cost above the bound, normalized by the span of training and by the budget itself,
\begin{equation}
\text{CumViol}(\pi) = \frac{1}{d\,(T_1 - T_0)} \int_{T_0}^{T_1} \max\big(\hat{C}^\pi(t) - d,\, 0\big)\, dt .
\end{equation}
CumViol is thus the average overshoot during training, expressed as a fraction of the cost budget: a value of zero means the method stayed within the budget throughout training, and a value of $0.5$ means that, averaged over training, its cost exceeded the bound by half the budget. Because the excess is measured against the same budget on every task, values are directly comparable across environments. 

\begin{figure}[!t]
    \centering
    \includegraphics[width=\linewidth]{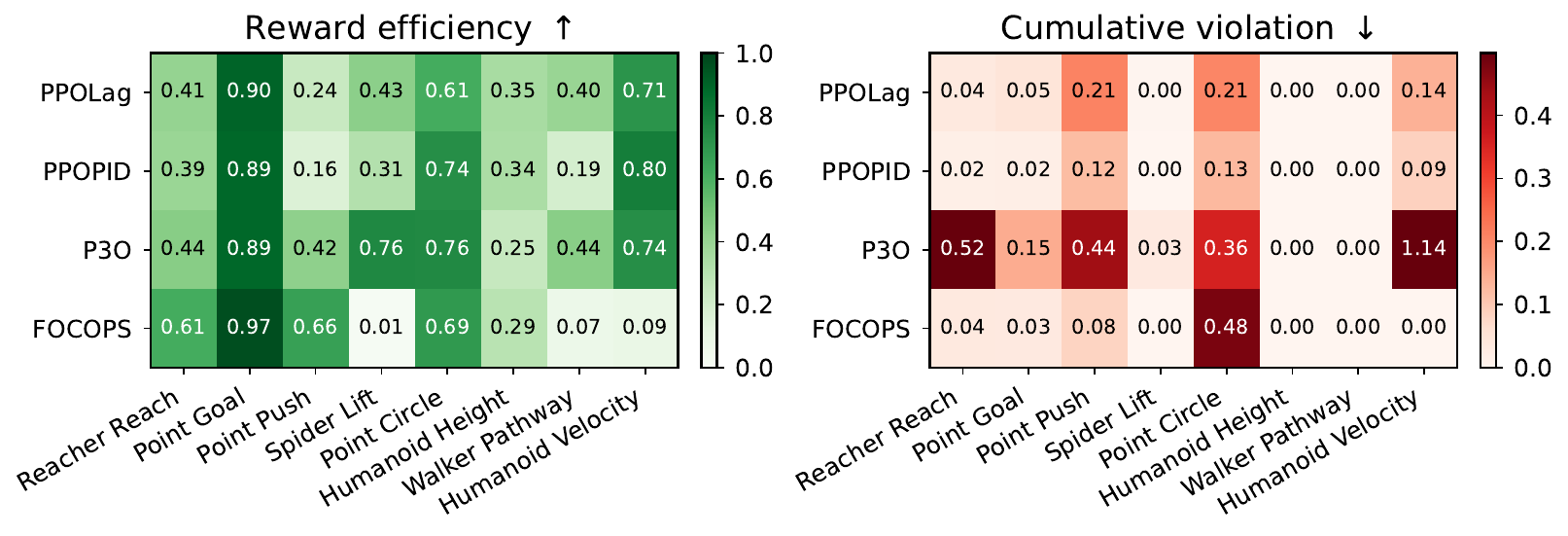}
    \caption{Efficiency and safety during training of the baselines on Level 1 tasks. \textbf{Left}: reward efficiency, the normalized area under the reward curve, where higher
    values indicate that a method reached strong performance earlier in training.
    \textbf{Right}: cumulative constraint violation during training, where zero means the method never exceeded the cost budget. Averages over 10 seeds.}
    \label{fig:sample-efficiency}
\end{figure}

\cref{fig:sample-efficiency} reports both metrics for PPOLag, PPOPID, P3O, and FOCOPS on Level~1.
The ranking by reward efficiency differs from the ranking by final reward.
P3O learns faster than both Lagrangian methods on five of the eight tasks and has the highest mean efficiency (0.59), but it also violates the constraint most during training, with the highest cumulative violation on every task except \texttt{Point Circle} and an average overshoot of more than the full budget on \texttt{Humanoid Velocity}.
PPOLag and PPOPID reach similar mean efficiency (0.51 and 0.48) but differ per task, with PPOLag ahead on \texttt{Point Push}, \texttt{Spider Lift}, and \texttt{Walker Pathway} and PPOPID ahead on \texttt{Point Circle} and \texttt{Humanoid Velocity}.
PPOPID also violates less than PPOLag on every task where either violates, consistent with its design goal of damping the overshoot of the Lagrange multiplier~\citep{DBLP:conf/icml/StookeAA20}.
The small gap between the two at convergence (\cref{sec:emperical_eval}) therefore hides different training trajectories.
FOCOPS shows its affinity for navigation from the start of training, as it is the most efficient method on \texttt{Reacher Reach}, \texttt{Point Goal}, and \texttt{Point Push}, but barely learns on \texttt{Spider Lift}, \texttt{Walker Pathway}, and \texttt{Humanoid Velocity}.

Where violations occur also depends on the task.
No method ever exceeds the bound on \texttt{Humanoid Height} and \texttt{Walker Pathway}, where even unconstrained PPO ends within it (\cref{tab:alg_comparison_detailed}), so the Level~1 threshold is not binding there, and differences between methods lie on the reward side alone.
On \texttt{Humanoid Velocity}, in contrast, all methods except FOCOPS exceed the bound during training even though all of them satisfy it at the end, which final-performance tables cannot show.
The navigation tasks \texttt{Point Push} and \texttt{Point Circle} are the only ones where every method violates the constraint noticeably, since the agent must first discover which behaviors are unsafe before it can avoid them.

\begin{figure}[!t]
    \centering
    \includegraphics[width=\linewidth]{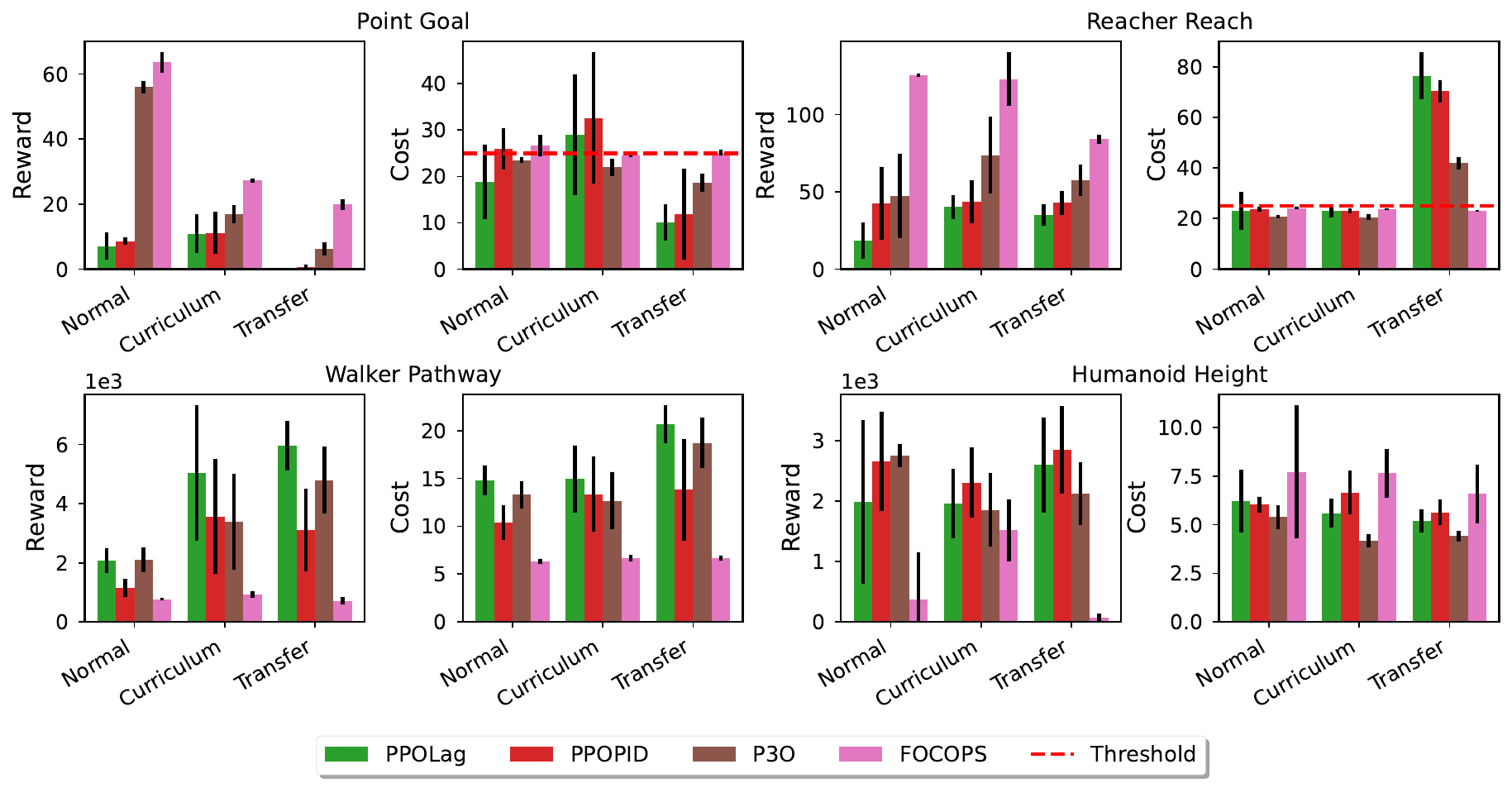}
    \caption{\textbf{Curriculum learning and safety transfer in CRAX environments.} 
    We compare direct training (\textbf{Normal}), curriculum learning across difficulty levels (\textbf{Curriculum}), and transfer from an unconstrained PPO policy (\textbf{Transfer}) on Level 3 tasks.}
    \label{fig:curriculum_transfer}
\end{figure}

\section{Curriculum Learning and Safety Transfer}
\label{app:curriculum_transfer}

As seen in \cref{fig:curves_level_3}, when trained directly on the hardest difficulty level, agents often struggle to discover a good policy, as exploration becomes dominated by constraint violations and sparse progress. 
Training RL agents on progressively more complex settings has been shown to substantially improve learning efficiency, a paradigm commonly referred to as curriculum learning \citep{bengio2009curriculum, narvekar2020curriculum}. 
In parallel, transfer learning aims to reuse knowledge acquired in one environment to accelerate learning in a related setting ~\citep{taylor2009transfer, zubia2025robust}.

We investigate whether curriculum and transfer learning can improve performance on the most difficult tasks in CRAX. 
In the curriculum setting, agents are trained sequentially on increasing difficulty levels, carrying over parameters between stages, with the data budget split equally across levels.
This exposes the agent to simpler dynamics before confronting denser hazards. 
In the transfer setting, we first train an unconstrained PPO policy directly on Level 3 and subsequently use its parameters to initialize a safe RL algorithm, which is then trained with the remaining half of the allowed timesteps. 

As shown in \cref{fig:curriculum_transfer}, the impact of curriculum learning and transfer is strongly environment- and algorithm-dependent. 
In \texttt{Point Goal}, neither curriculum nor transfer improves over direct training. 
In \texttt{Reacher Reach}, curriculum learning boosts performance for all methods except PPOPID. 
Transfer proves ineffective, as all methods except FOCOPS violate the threshold after transfer, suggesting that unconstrained policies struggle to adapt for safety in this environment.
Curriculum learning improves performance in \texttt{Walker Pathway} and \texttt{Humanoid Height}, whereas safe transfer yields benefits only in \texttt{Walker Pathway}.
Here, direct training yields overly conservative policies that underutilize the available safety budget, while safety-transferred agents make better use of it.
Overall, these results indicate that curriculum learning and transfer can be beneficial in some circumstances for learning challenging safety-constrained tasks.

\begin{figure}[!t]
    \centering
    \includegraphics[width=\linewidth]{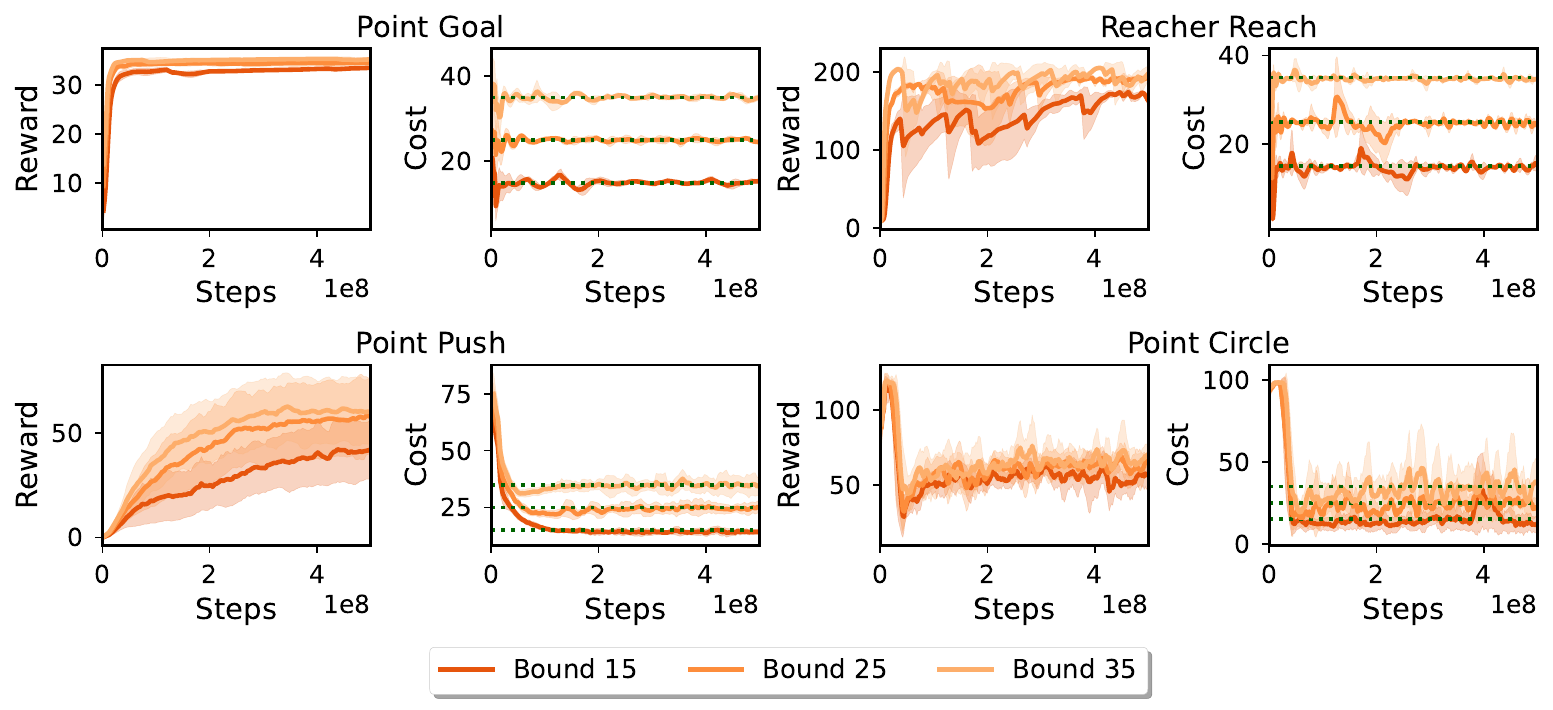}
    \caption{PPOLag increasingly sacrifices rewards to adhere to tighter safety bounds in all tasks.}
    \label{fig:safety_bounds}
\end{figure}

\section{Varying Safety Bounds}
\label{app:safety_bounds}

Safety requirements in real-world applications can vary substantially across use cases and are often accompanied by trade-offs with performance. For instance, autonomous vehicle navigation must contend with unpredictable pedestrian behavior and sensor limitations, making perfect safety unattainable in certain circumstances. The objective instead becomes to minimize unsafe behavior. In contrast, domains such as nuclear reactor control require absolute precision and tolerate no margin for error. To support such requirements, CRAX environments expose an adjustable safety bound. We examine how varying this constraint induces safety-performance trade-offs by evaluating PPOLag under stricter and more lenient cost budgets $d \in \{15, 25~(\text{default}), 35\}$. As shown in \cref{fig:safety_bounds}, PPOLag reliably adapts to the varied bounds across all environments, with a non-linear trade-off in performance. Tightening the bound leads to a substantially larger drop in score than the gains obtained by increasing it by the same amount. Moreover, some safe RL methods~\citep{wachi2024survey} are explicitly designed to enforce strict safety guarantees rather than negotiate trade-offs, and can therefore also be evaluated under the most stringent safety bounds.

\section{Off-Policy Baselines}
\label{app:off_policy}

The baselines in \cref{sec:emperical_eval} are all on-policy and most build on PPO. On-policy methods suit a fully compiled JAX training loop particularly well, since environment stepping, rollout collection, and the policy update compile into a single on-device program, and the regular alternation of large synchronous rollouts and updates keeps the accelerator busy. Off-policy methods fit this design less cleanly. A replay buffer requires dynamically indexed reads and writes, and the ratio of environment steps to gradient updates is less regular. They still depend on environment interaction, however, and a replay buffer does not substitute for simulation throughput, so they also benefit from accelerated simulation.

To investigate how well \acronym~supports off-policy learning, we implement SAC and SACLag and compare them with PPO and PPOLag on four Level~1 environments.
SAC~\citep{haarnoja2018soft} is an off-policy actor-critic method that maximizes an entropy-regularized return, training twin Q-critics on transitions sampled from a replay buffer.
SACLag~\citep{ha2020learning} adds a cost critic and a Lagrange multiplier $\lambda$, so that the actor maximizes $Q_r - \lambda Q_c$ while $\lambda$ grows as long as the episodic cost exceeds the bound, mirroring the dual update of PPOLag.

Two choices keep the comparison fair. First, all methods receive the same budget of 500M environment steps. Because off-policy methods reuse each transition, matching the budget does not by itself fix the amount of learning, so we also control how often each transition is replayed. Each actor step of \acronym's SAC collects $N$ transitions across $N$ parallel environments and is followed by \texttt{grad\_updates\_per\_step} gradient steps on minibatches of \texttt{batch\_size}, so each transition is replayed $(\texttt{grad\_updates\_per\_step} \times \texttt{batch\_size}) / N$ times on average. We set $N = 2048$, \texttt{batch\_size} $= 1024$, and \texttt{grad\_updates\_per\_step} $= 32$, which gives a replay ratio of $16$ while still saturating the GPU with large synchronous rollouts. Second, the replay buffer holds the most recent $10^6$ transitions rather than the full budget, since storing all 500M would be infeasible and most of them would be far off-policy by the end of training. For SACLag, the multiplier learning rate is expressed on a per-episode scale and normalized internally by the episode length, which makes it directly comparable to PPOLag's \texttt{lagrangian\_coef\_rate}.

\begin{figure}[!t]
    \centering
    \includegraphics[width=\linewidth]{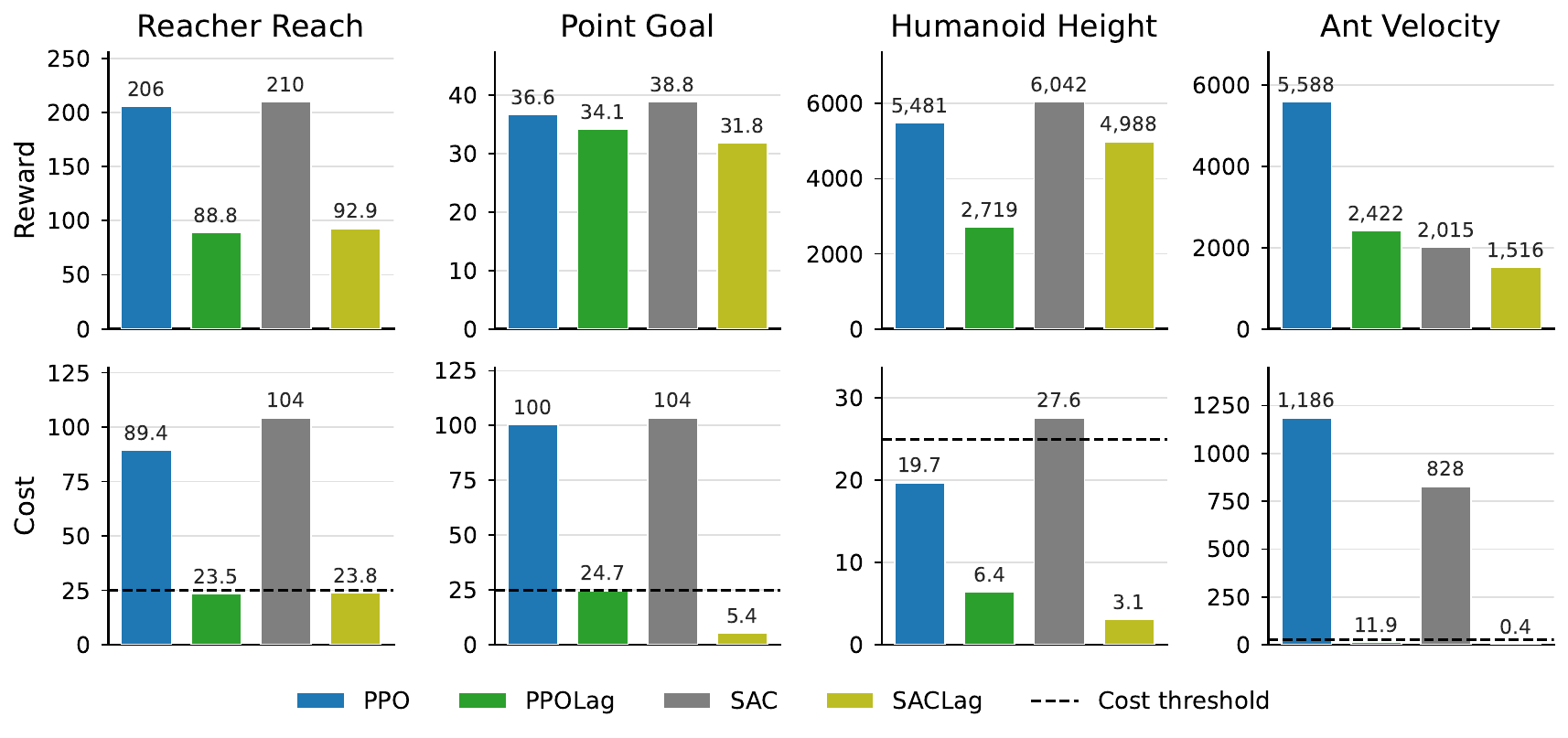}
    \caption{Final reward (top) and cost (bottom) of on-policy (PPO, PPOLag) and off-policy (SAC, SACLag) methods on four Level~1 environments. The dashed line marks the cost threshold of 25.}
    \label{fig:ppo_vs_sac}
\end{figure}

\begin{figure}[!t]
    \centering
    \includegraphics[width=.8\linewidth]{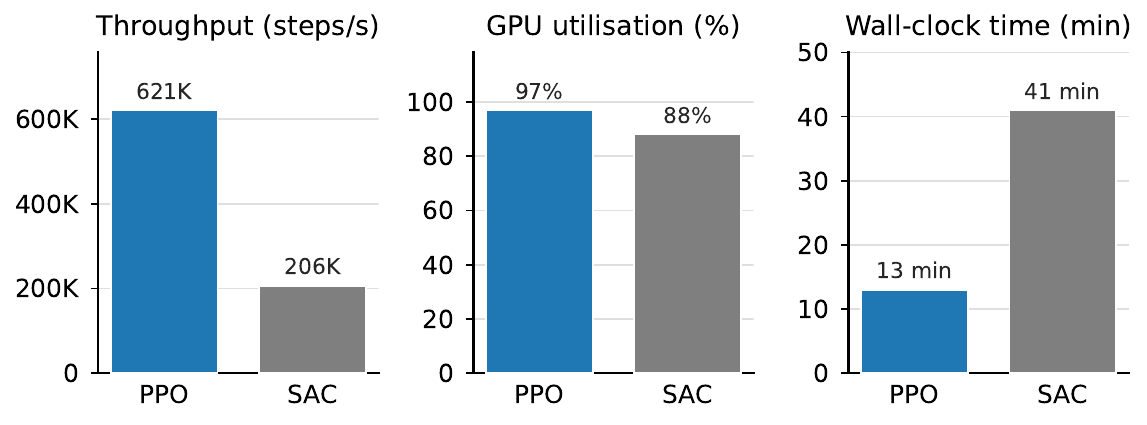}
    \caption{Training efficiency of PPO and SAC on \texttt{Reacher Reach} Level~1 for a 500M-step run on a single NVIDIA H100 GPU.}
    \label{fig:ppo_vs_sac_efficiency}
\end{figure}

\cref{fig:ppo_vs_sac} shows that the off-policy methods perform comparably to their on-policy counterparts, with neither family dominating. SACLag satisfies the cost bound on all four environments and exceeds PPOLag in reward on \texttt{Humanoid Height} (4,988 against 2,719) and marginally on \texttt{Reacher Reach}, while PPOLag leads on \texttt{Point Goal} and \texttt{Ant Velocity}. On these two, SACLag settles far below the bound (costs of 5.4 and 0.4), whereas PPOLag uses most of the available budget. Unconstrained SAC exceeds the bound on all four environments, including \texttt{Humanoid Height}, where PPO stays within it.

\cref{fig:ppo_vs_sac_efficiency} compares training efficiency on \texttt{Reacher Reach}. SAC reaches 206K steps per second against 621K for PPO, so a full run takes 41 minutes rather than 13, and GPU utilization drops from 97\% to 88\%. Off-policy training is therefore about three times slower in \acronym, but its throughput remains far above the 5K to 20K steps per second that Safety-Gymnasium reached in \cref{sec:computational_efficiency}, although that measurement used a different environment.

\section{Fidelity to Safety-Gymnasium}
\label{app:sg_fidelity}

The Navigation and Velocity task in CRAX descend from Safety-Gymnasium (SG). We did not aim for exact replicas of the SG versions. Instead, we made the tasks harder, since CRAX's throughput lets us train for far more environment steps, and we wish the tasks to remain a meaningful challenge for future methods. Point Goal and Point Push, for instance, place more goals and denser hazards of mixed geometry and size within a walled arena, and Point Push additionally makes the goal move at higher difficulty levels. Absolute scores on these tasks are therefore not comparable to published Safety-Gymnasium numbers.

The \texttt{Velocity} task is an exception, however. The task is fully specified by a standard MuJoCo locomotion model together with a scalar speed threshold, so it can, in principle, be aligned closely across the two backends. We use this alignment to check how faithfully CRAX reproduces SG's learning behavior. We pick the Ant agent, whose morphology both benchmarks share and whose Level 1 cost threshold of $2.62$ matches the one used in \texttt{SafetyAntVelocity-v1}.

\begin{figure}[!t]
  \centering
  \includegraphics[width=\linewidth]{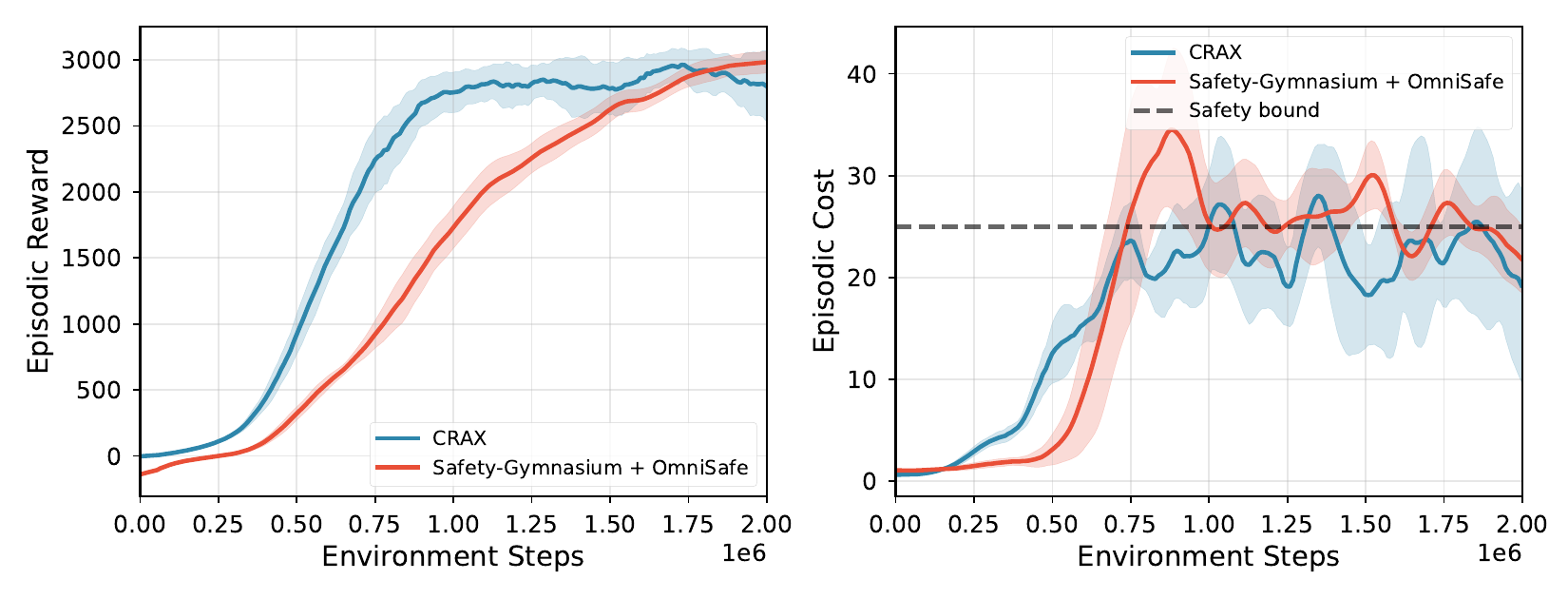}
  \caption{Safe Ant Velocity in CRAX and in Safety-Gymnasium with OmniSafe, both trained
  with PPO-Lagrangian for 2M environment steps over 10 seeds under a cost threshold of
  $25$. Shaded regions denote 95\% confidence intervals.}
  \label{fig:sg_fidelity}
\end{figure}

\paragraph{Setup.}
On Safety-Gymnasium we train \texttt{SafetyAntVelocity-v1} with OmniSafe's
PPO-Lagrangian under its default configuration. On CRAX we train \texttt{safe\_velocity\_ant} with our own PPO-Lagrangian, disabling the reward scaling (\cref{tab:hyperparams}) and fixing the episode length to 1000 steps so that returns are on the same scale. Both use a cost threshold of $25$, a budget of 2M environment steps, and 10 seeds. \cref{fig:sg_fidelity} depicts the training curves with 95\% confidence intervals.

Several differences remain that we did not attempt to remove. MJX and CPU MuJoCo resolve contacts differently. The two PPO implementations do not share network sizes or optimizer defaults, and CRAX squashes actions through a $\tanh$ while OmniSafe samples raw Gaussian actions that MuJoCo clips at the actuator. The comparison is between two implementations of the same task rather than two runs of the same program, so small numerical gaps are
expected.

\paragraph{Results.}
Both setups converge to an episodic reward close to $3000$ with episodic cost oscillating around the bound of $25$, and both reproduce the same Lagrangian behavior. A few differences stand out. (1) CRAX reaches its reward plateau after $\sim1.0M$ steps
against $\sim1.9M$ for SG. We do not claim per-step equivalence, as the contact dynamics and PPO implementations both differ. (2) CRAX overshoots the safety bound less. (3) At initialization, SG starts near $-175$ reward and CRAX near $-10$. Squashing bounds the control penalty from the first step, whereas unclipped Gaussian samples incur a large
\texttt{ctrl\_cost} before MuJoCo clips them. The gap closes once both policies move away from their initial action scale. This comparison is meant to serve as a sanity check on the inherited part of the benchmark.

\section{Hyperparameters}
\label{app:hyperparameters}

The network architecture, optimizer, and base PPO hyperparameters follow the Brax PPO implementation~\citep{DBLP:conf/nips/FreemanFRGMB21}. \cref{tab:hyperparams} lists the shared configuration, \cref{tab:hyperparams_alg} the algorithm-specific settings, and \cref{tab:hyperparams_sac_vision} the settings of the off-policy baselines and of pixel-based training. All experiments use these values unless stated otherwise. Reported reward and cost are undiscounted episode sums from the training rollouts, logged every $10^6$ environment steps.

\begin{table}[htbp]
\caption{Shared configuration of the PPO-based methods.}
\label{tab:hyperparams}
\centering\small
\begin{tabular}{@{}ll@{}}
\toprule
\textbf{Parameter} & \textbf{Value} \\
\midrule
\multicolumn{2}{@{}l}{\emph{Optimization}} \\
Optimizer & Adam ($\beta_1 = 0.9$, $\beta_2 = 0.999$, $\epsilon = 10^{-8}$) \\
Learning rate & $5 \cdot 10^{-4}$, constant \\
Gradient clipping & None \\
Entropy coefficient & $5 \cdot 10^{-3}$ \\
Discount $\gamma$ & 0.99 \\
GAE $\lambda$ & 0.95 \\
Clip $\epsilon$ & 0.3 (not used by FOCOPS) \\
Reward scaling & 0.1 \\
Advantage normalization & Yes \\
Value loss coefficient & 0.25 \\
\addlinespace
\multicolumn{2}{@{}l}{\emph{Networks}} \\
Policy & MLP, $4 \times 32$, Swish \\
Action distribution & Tanh-squashed Gaussian, state-dependent std (softplus $+\,10^{-3}$) \\
Reward critic & MLP, $5 \times 256$, Swish \\
Cost critic & Separate MLP, $5 \times 256$, Swish (methods with a cost critic) \\
Observation normalization & Running mean and standard deviation \\
\addlinespace
\multicolumn{2}{@{}l}{\emph{Training}} \\
Environment steps & $5 \cdot 10^{8}$ \\
Parallel environments & 2048 \\
Unroll length & 8 \\
Batch size & 1024 \\
Minibatches per update & 32 \\
SGD epochs per batch & 6 \\
Episode length & 200 (Reach), 500 (Lift), 1000 (Goal, Button, Circle, Height), \\
 & 2000 (Push, Pathway, Velocity) \\
Cost bound $d$ & 25 per episode \\
\bottomrule
\end{tabular}
\end{table}

\begin{table}[htbp]
\caption{Algorithm-specific hyperparameters. Multiplier and penalty updates use the per-step bound $d/T$, where $T$ is the episode length.}
\label{tab:hyperparams_alg}
\centering\small
\begin{tabular}{@{}lll@{}}
\toprule
\textbf{Method} & \textbf{Parameter} & \textbf{Value} \\
\midrule
PPOCost & Cost weight & 1.0 \\
\addlinespace
PPOLag & Multiplier learning rate & 3 \\
 & Initial $\lambda$ & 0 \\
\addlinespace
PPOPID & Gains $(K_p, K_i, K_d)$ & $(10,\ 0.01,\ 0.01)$ \\
 & Integral clip / $\lambda$ clip & 1.0 / $10^{6}$ \\
 & Derivative EMA $\beta$ & 0.95 \\
\addlinespace
PPOSaut\'e & Budget discount & 0.99 (equal to $\gamma$) \\
 & Violation penalty & $-1$ \\
 & Normalized budget observation & Yes \\
\addlinespace
P3O & Initial $\kappa$ & 0.01 \\
 & $\kappa$ increase / decrease factor & 1.1 / 0.9 \\
 & Maximum $\kappa$ & 50 \\
\addlinespace
FOCOPS & Initial $\nu$ / $\nu$ learning rate / maximum $\nu$ & 0.1 / 1.0 / 100 \\
 & KL coefficient $\lambda$ & 1.5 \\
 & Temperature $\eta$ & 0.02 \\
\addlinespace
CRPO & Switching tolerance (per step) & 0 \\
\bottomrule
\end{tabular}
\end{table}

\begin{table}[htbp]
\caption{Configuration of the off-policy baselines (\cref{app:off_policy}) and of pixel-based training (\cref{sec:learning_from_pixels}).}
\label{tab:hyperparams_sac_vision}
\centering\small
\begin{tabular}{@{}ll@{}}
\toprule
\textbf{Parameter} & \textbf{Value} \\
\midrule
\multicolumn{2}{@{}l}{\emph{SAC and SACLag}} \\
Learning rate (policy and critics) & $5 \cdot 10^{-4}$ \\
Entropy temperature & Tuned automatically (learning rate $3 \cdot 10^{-4}$, initial $\alpha = 1$) \\
Target entropy & $-|\mathcal{A}|/2$ \\
Policy & MLP, $2 \times 256$, ReLU \\
Critics & Twin Q-networks, $2 \times 256$, ReLU (separate cost pair for SACLag) \\
Discount $\gamma$ / target smoothing $\tau$ & 0.99 / 0.005 \\
Parallel environments & 2048 \\
Batch size & 1024 \\
Gradient updates per environment step & 32 \\
Replay buffer & $10^{6}$ transitions, FIFO, uniform sampling \\
Reward scaling & 0.1 \\
SACLag multiplier learning rate & 0.01 per episode (divided by $T$) \\
SACLag initial / maximum $\lambda$ & 0 / 2 \\
\addlinespace
\multicolumn{2}{@{}l}{\emph{Pixel-based training}} \\
Observation & One $64 \times 64$ RGB frame \\
Encoder & Nature-DQN CNN (32@$8{\times}8$/4, 64@$4{\times}4$/2, 64@$3{\times}3$/1), ReLU, \\
 & 1024 features per camera, shared by policy and critics \\
Heads & Policy $2 \times 256$, reward critic $2 \times 256$, cost critic $5 \times 256$, Swish \\
Augmentation & Random shift, 4-pixel edge padding \\
\bottomrule
\end{tabular}
\end{table}

\subsection{Hyperparameter Sweeps}
\label{app:sweeps}

\begin{figure}[!t]
    \centering
    \includegraphics[width=\linewidth]{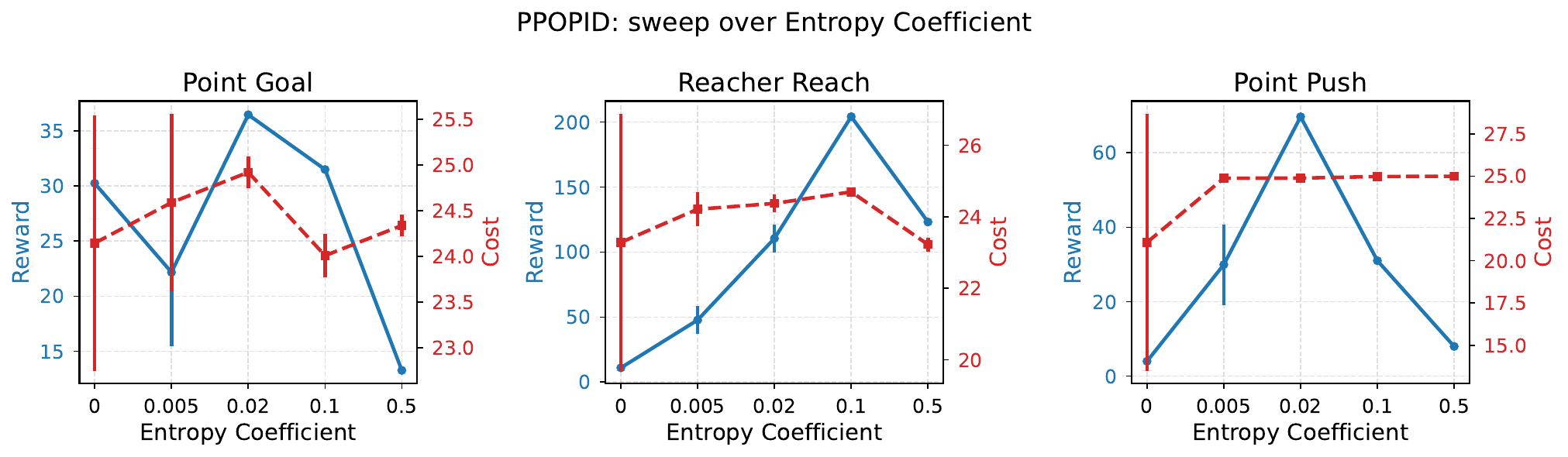}
    \caption{\textbf{Example sweep} of the entropy coefficient for PPOPID. Reward (blue, left axis) peaks at intermediate values, while cost (red, right axis) stays within about one unit of the bound of 25. Note the narrow range of the cost axis. Error bars denote 95\% confidence intervals across three seeds.}
    \label{fig:sweep_example}
\end{figure}

The Brax defaults were tuned for unconstrained locomotion, so we ran two sweeps on \texttt{Point Goal}, \texttt{Point Push}, and \texttt{Reacher Reach} at Level~1, varying one hyperparameter at a time over three seeds. The first varies the shared RL hyperparameters of PPOLag, PPOPID, P3O, and FOCOPS around the configuration in \cref{tab:hyperparams} for 300M steps (\cref{tab:sweep_shared_goal,tab:sweep_shared_push,tab:sweep_shared_reacher}). The second varies the safety-specific hyperparameters of each method for 500M steps (\cref{tab:sweep_safety}). We aim to determine the best hyperparameters and measure their sensitivity. \cref{fig:sweep_example} shows one sweep in detail.

Cost is largely insensitive to these choices. PPOLag, PPOPID, and FOCOPS stay within about 1.5 of the bound on \texttt{Point Goal} and \texttt{Point Push} in all but one configuration, and exceed it more on \texttt{Reacher Reach} only in a few cases, such as PPOLag without entropy regularization (32.0) and PPOPID with a GAE $\lambda$ of 0.9 (31.4). P3O is less robust and exceeds the bound on \texttt{Reacher Reach} under its base configuration (32.2) and several others, up to 49.8. PPOSaut\'e is the most sensitive method, since its budget discount decides whether it ignores the constraint, with costs of 84 to 102 at 0.95, or satisfies it at near-zero or negative reward at 0.999.

Reward is far more sensitive, mostly to the shared hyperparameters. For the three PPO-based methods, lowering the clip $\epsilon$ from 0.3 to 0.1 raises the reward on \texttt{Point Push} from between 25 and 34 to between 81 and 82, and raising the entropy coefficient from 0.005 to 0.02 raises it to between 70 and 75. A larger learning rate also helps, whereas the discount and GAE $\lambda$ matter less. The safety-specific hyperparameters of these methods change reward by at most a factor of about 2.5, as for P3O's initial $\kappa$ on \texttt{Reacher Reach} (31.4 to 76.5). FOCOPS is the least sensitive method, with similar reward across all settings except a zero entropy coefficient on \texttt{Reacher Reach} (22.8 against 112.9). \texttt{Point Goal} is the least sensitive task, with rewards between 26 and 38 for all methods other than PPOSaut\'e, apart from an entropy coefficient of 0.5 for PPOPID.

\begin{table}[htbp]
\caption{Shared RL hyperparameter sweep on \texttt{Point Goal} Level~1 (300M steps, three seeds). R: reward, C: cost. The base value of each hyperparameter is marked with \textsuperscript{$\ast$}.}
\label{tab:sweep_shared_goal}
\centering\small
\begin{tabular}{llrrrrrrrr}
\toprule
 & & \multicolumn{2}{c}{PPOLag} & \multicolumn{2}{c}{PPOPID} & \multicolumn{2}{c}{P3O} & \multicolumn{2}{c}{FOCOPS} \\
\cmidrule(lr){3-4}\cmidrule(lr){5-6}\cmidrule(lr){7-8}\cmidrule(lr){9-10}
Hyperparameter & Value & R & C & R & C & R & C & R & C \\
\midrule
Learning rate & $10^{-4}$ & 31.6 & 24.8 & 32.2 & 24.8 & 26.2 & 24.9 & 36.7 & 24.9 \\
 & $5\cdot10^{-4}$\textsuperscript{$\ast$} & 33.9 & 23.8 & 34.0 & 23.9 & 34.3 & 24.7 & 36.8 & 24.8 \\
 & $10^{-3}$ & 34.9 & 24.8 & 35.4 & 25.0 & 34.8 & 25.5 & 35.6 & 25.1 \\
\addlinespace
Entropy coef. & 0 & 31.8 & 23.3 & 30.2 & 24.1 & 31.3 & 24.0 & 37.4 & 24.9 \\
 & 0.005\textsuperscript{$\ast$} & 33.9 & 23.8 & 34.0 & 23.9 & 34.3 & 24.7 & 36.8 & 24.8 \\
 & 0.02 & 36.4 & 24.9 & 36.5 & 24.9 & 36.4 & 25.0 & 32.7 & 25.2 \\
\addlinespace
Discount $\gamma$ & 0.95 & 34.7 & 25.0 & 34.9 & 25.0 & 34.2 & 25.2 & 36.7 & 24.8 \\
 & 0.99\textsuperscript{$\ast$} & 33.9 & 23.8 & 34.0 & 23.9 & 34.3 & 24.7 & 36.8 & 24.8 \\
 & 0.995 & 34.2 & 24.9 & 33.9 & 25.0 & 31.5 & 24.4 & 36.6 & 25.2 \\
\addlinespace
GAE $\lambda$ & 0.9 & 34.7 & 24.7 & 34.1 & 24.3 & 34.1 & 25.1 & 36.9 & 25.0 \\
 & 0.95\textsuperscript{$\ast$} & 33.9 & 23.8 & 34.0 & 23.9 & 34.3 & 24.7 & 36.8 & 24.8 \\
 & 0.99 & 34.2 & 25.0 & 33.9 & 25.1 & 33.6 & 24.8 & 36.7 & 24.8 \\
\addlinespace
Clip $\epsilon$ & 0.1 & 37.2 & 24.9 & 37.3 & 25.0 & 37.4 & 25.3 & -- & -- \\
 & 0.2 & 36.2 & 24.9 & 36.0 & 24.9 & 36.1 & 25.2 & -- & -- \\
 & 0.3\textsuperscript{$\ast$} & 33.9 & 23.8 & 34.0 & 23.9 & 34.3 & 24.7 & -- & -- \\
\bottomrule\end{tabular}\end{table}

\begin{table}[htbp]
\caption{Shared RL hyperparameter sweep on \texttt{Point Push} Level~1 (300M steps, three seeds). R: reward, C: cost. The base value of each hyperparameter is marked with \textsuperscript{$\ast$}.}
\label{tab:sweep_shared_push}
\centering\small
\begin{tabular}{llrrrrrrrr}
\toprule
 & & \multicolumn{2}{c}{PPOLag} & \multicolumn{2}{c}{PPOPID} & \multicolumn{2}{c}{P3O} & \multicolumn{2}{c}{FOCOPS} \\
\cmidrule(lr){3-4}\cmidrule(lr){5-6}\cmidrule(lr){7-8}\cmidrule(lr){9-10}
Hyperparameter & Value & R & C & R & C & R & C & R & C \\
\midrule
Learning rate & $10^{-4}$ & 10.1 & 24.7 & 8.5 & 25.0 & 20.9 & 25.8 & 49.8 & 24.6 \\
 & $5\cdot10^{-4}$\textsuperscript{$\ast$} & 25.3 & 24.6 & 31.4 & 25.2 & 34.0 & 25.3 & 50.3 & 24.2 \\
 & $10^{-3}$ & 51.4 & 25.1 & 58.4 & 24.7 & 60.2 & 25.3 & 47.7 & 24.9 \\
\addlinespace
Entropy coef. & 0 & 4.7 & 24.6 & 4.0 & 21.1 & 9.8 & 24.6 & 53.5 & 26.4 \\
 & 0.005\textsuperscript{$\ast$} & 25.3 & 24.6 & 31.4 & 25.2 & 34.0 & 25.3 & 50.3 & 24.2 \\
 & 0.02 & 70.1 & 25.0 & 69.7 & 24.9 & 74.5 & 25.7 & 29.3 & 24.8 \\
\addlinespace
Discount $\gamma$ & 0.95 & 33.8 & 25.1 & 38.3 & 25.0 & 39.7 & 25.8 & 43.2 & 24.3 \\
 & 0.99\textsuperscript{$\ast$} & 25.3 & 24.6 & 31.4 & 25.2 & 34.0 & 25.3 & 50.3 & 24.2 \\
 & 0.995 & 14.7 & 25.1 & 23.0 & 24.9 & 23.4 & 25.5 & 48.0 & 25.3 \\
\addlinespace
GAE $\lambda$ & 0.9 & 19.9 & 24.8 & 27.6 & 25.0 & 38.0 & 25.0 & 47.3 & 24.6 \\
 & 0.95\textsuperscript{$\ast$} & 25.3 & 24.6 & 31.4 & 25.2 & 34.0 & 25.3 & 50.3 & 24.2 \\
 & 0.99 & 21.3 & 25.0 & 30.0 & 25.0 & 31.6 & 25.2 & 48.4 & 25.0 \\
\addlinespace
Clip $\epsilon$ & 0.1 & 81.2 & 24.9 & 81.4 & 25.1 & 82.1 & 25.9 & -- & -- \\
 & 0.2 & 74.7 & 25.1 & 75.3 & 24.9 & 66.7 & 25.4 & -- & -- \\
 & 0.3\textsuperscript{$\ast$} & 25.3 & 24.6 & 31.4 & 25.2 & 34.0 & 25.3 & -- & -- \\
\bottomrule\end{tabular}\end{table}

\begin{table}[htbp]
\caption{Shared RL hyperparameter sweep on \texttt{Reacher Reach} Level~1 (300M steps, three seeds). R: reward, C: cost. The base value of each hyperparameter is marked with \textsuperscript{$\ast$}.}
\label{tab:sweep_shared_reacher}
\centering\small
\begin{tabular}{llrrrrrrrr}
\toprule
 & & \multicolumn{2}{c}{PPOLag} & \multicolumn{2}{c}{PPOPID} & \multicolumn{2}{c}{P3O} & \multicolumn{2}{c}{FOCOPS} \\
\cmidrule(lr){3-4}\cmidrule(lr){5-6}\cmidrule(lr){7-8}\cmidrule(lr){9-10}
Hyperparameter & Value & R & C & R & C & R & C & R & C \\
\midrule
Learning rate & $10^{-4}$ & 20.7 & 23.8 & 26.0 & 24.4 & 61.0 & 49.8 & 94.8 & 24.1 \\
 & $5\cdot10^{-4}$\textsuperscript{$\ast$} & 38.0 & 25.6 & 48.8 & 25.4 & 28.9 & 32.2 & 112.9 & 24.3 \\
 & $10^{-3}$ & 82.3 & 24.4 & 59.2 & 24.1 & 83.9 & 22.3 & 109.0 & 24.2 \\
\addlinespace
Entropy coef. & 0 & 11.8 & 32.0 & 11.0 & 23.3 & 29.4 & 23.5 & 22.8 & 22.1 \\
 & 0.005\textsuperscript{$\ast$} & 38.0 & 25.6 & 48.8 & 25.4 & 28.9 & 32.2 & 112.9 & 24.3 \\
 & 0.02 & 117.3 & 24.3 & 110.5 & 24.4 & 65.8 & 22.3 & 111.5 & 24.3 \\
\addlinespace
Discount $\gamma$ & 0.95 & 49.8 & 23.5 & 34.7 & 21.8 & 43.0 & 30.4 & 116.8 & 24.4 \\
 & 0.99\textsuperscript{$\ast$} & 38.0 & 25.6 & 48.8 & 25.4 & 28.9 & 32.2 & 112.9 & 24.3 \\
 & 0.995 & 28.5 & 25.4 & 31.2 & 26.3 & 21.4 & 29.6 & 106.2 & 24.2 \\
\addlinespace
GAE $\lambda$ & 0.9 & 42.0 & 24.8 & 39.4 & 31.4 & 25.3 & 27.2 & 116.3 & 24.3 \\
 & 0.95\textsuperscript{$\ast$} & 38.0 & 25.6 & 48.8 & 25.4 & 28.9 & 32.2 & 112.9 & 24.3 \\
 & 0.99 & 46.9 & 22.8 & 50.6 & 24.3 & 49.0 & 26.9 & 112.1 & 24.2 \\
\addlinespace
Clip $\epsilon$ & 0.1 & 108.7 & 24.3 & 94.6 & 24.1 & 116.4 & 22.1 & -- & -- \\
 & 0.2 & 70.4 & 23.3 & 83.8 & 24.1 & 90.4 & 22.2 & -- & -- \\
 & 0.3\textsuperscript{$\ast$} & 38.0 & 25.6 & 48.8 & 25.4 & 28.9 & 32.2 & -- & -- \\
\bottomrule\end{tabular}\end{table}

\begin{table}[htbp]
\caption{Safety-specific hyperparameter sweep on Level~1 (500M steps, three seeds). R: reward, C: cost. Base values are marked with \textsuperscript{$\ast$}.}
\label{tab:sweep_safety}
\centering\small
\begin{tabular}{lllrrrrrr}
\toprule
 & & & \multicolumn{2}{c}{Point Goal} & \multicolumn{2}{c}{Point Push} & \multicolumn{2}{c}{Reacher Reach} \\
\cmidrule(lr){4-5}\cmidrule(lr){6-7}\cmidrule(lr){8-9}
Method & Hyperparameter & Value & R & C & R & C & R & C \\
\midrule
PPOLag & Multiplier LR & 3\textsuperscript{$\ast$} & 34.5 & 24.5 & 32.1 & 25.2 & 56.8 & 24.0 \\
 &  & 10 & 34.4 & 24.9 & 39.4 & 25.1 & 50.7 & 22.3 \\
 &  & 30 & 34.1 & 25.0 & 23.7 & 24.9 & 40.6 & 28.3 \\
\midrule
PPOPID & $K_p$ & 1 & 34.8 & 28.3 & 37.4 & 23.5 & 81.2 & 25.4 \\
 &  & 10\textsuperscript{$\ast$} & 34.6 & 25.1 & 20.8 & 25.2 & 67.9 & 24.2 \\
 &  & 30 & 34.2 & 25.1 & 30.1 & 25.9 & 40.5 & 23.2 \\
\addlinespace
 & $K_i$ & 0.001 & 34.3 & 25.0 & 38.2 & 24.6 & 62.4 & 24.2 \\
 &  & 0.01\textsuperscript{$\ast$} & 34.6 & 25.1 & 20.8 & 25.2 & 67.9 & 24.2 \\
 &  & 0.1 & 34.2 & 24.4 & 36.3 & 25.0 & 62.5 & 24.4 \\
\midrule
PPOSaut\'e & Budget discount & 0.95 & 36.9 & 99.8 & 54.7 & 84.5 & 194.4 & 102.2 \\
 &  & 0.99\textsuperscript{$\ast$} & 34.6 & 93.6 & 16.6 & 48.4 & 24.8 & 12.1 \\
 &  & 0.999 & -0.6 & 0.2 & -5.1 & 4.0 & 34.4 & 11.4 \\
\addlinespace
 & Violation penalty & -20 & -11.2 & 12.2 & -26.4 & 4.5 & -135.3 & 7.9 \\
 &  & -5 & 20.1 & 92.8 & -12.4 & 7.6 & -27.7 & 8.8 \\
 &  & -1\textsuperscript{$\ast$} & 34.6 & 93.6 & 16.6 & 48.4 & 24.8 & 12.1 \\
\midrule
P3O & Initial $\kappa$ & 0.001 & 33.8 & 24.4 & 57.1 & 25.6 & 66.5 & 23.4 \\
 &  & 0.01\textsuperscript{$\ast$} & 34.5 & 25.2 & 47.9 & 25.3 & 31.4 & 25.4 \\
 &  & 0.1 & 34.6 & 25.0 & 34.0 & 24.6 & 76.5 & 22.6 \\
\addlinespace
 & $\kappa$ increase factor & 1.05 & 33.2 & 25.2 & 63.3 & 26.2 & 39.9 & 30.4 \\
 &  & 1.1\textsuperscript{$\ast$} & 34.5 & 25.2 & 47.9 & 25.3 & 31.4 & 25.4 \\
 &  & 1.3 & 33.8 & 24.5 & 28.8 & 22.5 & 67.5 & 21.1 \\
\midrule
FOCOPS & $\lambda$ (KL coef.) & 0.5 & 37.9 & 25.0 & 71.2 & 24.0 & 123.7 & 24.3 \\
 &  & 1.5\textsuperscript{$\ast$} & 37.2 & 25.0 & 52.8 & 25.4 & 120.7 & 24.1 \\
 &  & 5 & 34.0 & 24.9 & 36.0 & 24.8 & 120.8 & 24.2 \\
\addlinespace
 & $\nu$ learning rate & 0.3 & 37.3 & 24.9 & 55.5 & 24.2 & 123.2 & 24.1 \\
 &  & 1\textsuperscript{$\ast$} & 37.2 & 25.0 & 52.8 & 25.4 & 120.7 & 24.1 \\
 &  & 3 & 37.3 & 24.9 & 53.2 & 24.8 & 131.0 & 24.1 \\
\bottomrule\end{tabular}\end{table}

\section{Detailed Baseline Results}
\label{app:extended_results}

In this section, we provide detailed experimental results that complement the main findings.

\subsection{Final Performance}
\cref{tab:alg_comparison_detailed} provides a comprehensive overview of the baseline results across all difficulty levels and tasks.

\begin{table}[htbp]
\small
\caption{Detailed algorithm comparison across environments and difficulty levels. R: Reward ($\uparrow$ higher is better), C: Cost ($\downarrow$ lower is better). \textcolor{safegreen}{Green} indicates safe (cost $< 25$). \textbf{Bold} indicates best safe result.}
\label{tab:alg_comparison_detailed}
\centering
\begin{tabular}{p{1.0008cm}p{0.70cm}p{0.70cm}p{0.70cm}p{0.70cm}p{0.70cm}p{0.70cm}p{0.70cm}p{0.70cm}p{0.70cm}p{0.70cm}p{0.70cm}p{0.70cm}}
\toprule
 & \multicolumn{6}{c}{Point Goal} & \multicolumn{6}{c}{Reacher Reach} \\
\cmidrule(lr){2-7} \cmidrule(lr){8-13}
 & \multicolumn{2}{c}{Level 1} & \multicolumn{2}{c}{Level 2} & \multicolumn{2}{c}{Level 3} & \multicolumn{2}{c}{Level 1} & \multicolumn{2}{c}{Level 2} & \multicolumn{2}{c}{Level 3} \\
\cmidrule(lr){2-3} \cmidrule(lr){4-5} \cmidrule(lr){6-7} \cmidrule(lr){8-9} \cmidrule(lr){10-11} \cmidrule(lr){12-13}
Algorithm & R $\uparrow$ & C $\downarrow$ & R $\uparrow$ & C $\downarrow$ & R $\uparrow$ & C $\downarrow$ & R $\uparrow$ & C $\downarrow$ & R $\uparrow$ & C $\downarrow$ & R $\uparrow$ & C $\downarrow$ \\
\midrule
PPO & 36.6 & 100.4 & 30.0 & 97.3 & 25.7 & 192.7 & 205.9 & 84.8 & 203.6 & 74.1 & 206.2 & 105.6 \\
PPOCost & 27.1 & \textcolor{safegreen}{10.2} & 23.5 & 50.2 & 14.8 & 87.5 & 82.6 & \textcolor{safegreen}{19.8} & 144.9 & 51.4 & 78.6 & 36.4 \\
PPOLag & \textbf{34.3} & \textbf{\textcolor{safegreen}{24.9}} & 16.7 & \textcolor{safegreen}{7.3} & 7.0 & \textcolor{safegreen}{18.8} & 100.2 & \textcolor{safegreen}{23.9} & 66.6 & \textcolor{safegreen}{24.9} & 18.5 & \textcolor{safegreen}{23.1} \\
PPOPID & 34.0 & \textcolor{safegreen}{24.9} & 25.6 & 25.3 & 8.6 & 26.0 & 89.4 & \textcolor{safegreen}{23.4} & 129.9 & 26.1 & 42.7 & \textcolor{safegreen}{23.6} \\
PPOSaute & 33.9 & 94.0 & 61.3 & 126.9 & 52.7 & 191.0 & \textbf{176.6} & \textbf{\textcolor{safegreen}{6.2}} & 104.2 & \textcolor{safegreen}{13.1} & 21.8 & \textcolor{safegreen}{11.6} \\
P3O & 33.6 & 25.2 & 64.1 & \textcolor{safegreen}{23.4} & \textbf{56.0} & \textbf{\textcolor{safegreen}{23.5}} & 98.6 & \textcolor{safegreen}{24.6} & 139.3 & \textcolor{safegreen}{24.7} & 47.5 & \textcolor{safegreen}{20.8} \\
FOCOPS & 37.2 & 25.1 & \textbf{73.3} & \textbf{\textcolor{safegreen}{24.6}} & 63.6 & 26.6 & 149.4 & \textcolor{safegreen}{24.6} & 177.5 & \textcolor{safegreen}{24.6} & \textbf{125.4} & \textbf{\textcolor{safegreen}{24.2}} \\
CRPO & 33.2 & \textcolor{safegreen}{24.8} & 18.9 & 26.1 & 12.1 & 25.9 & 109.9 & \textcolor{safegreen}{21.3} & \textbf{186.9} & \textbf{\textcolor{safegreen}{23.0}} & 140.0 & 25.0 \\
\bottomrule
\end{tabular}
\\[1em]
\begin{tabular}{p{1.0008cm}p{0.70cm}p{0.70cm}p{0.70cm}p{0.70cm}p{0.70cm}p{0.70cm}p{0.70cm}p{0.70cm}p{0.70cm}p{0.70cm}p{0.70cm}p{0.70cm}}
\toprule
 & \multicolumn{6}{c}{Walker Pathway} & \multicolumn{6}{c}{Humanoid Velocity} \\
\cmidrule(lr){2-7} \cmidrule(lr){8-13}
 & \multicolumn{2}{c}{Level 1} & \multicolumn{2}{c}{Level 2} & \multicolumn{2}{c}{Level 3} & \multicolumn{2}{c}{Level 1} & \multicolumn{2}{c}{Level 2} & \multicolumn{2}{c}{Level 3} \\
\cmidrule(lr){2-3} \cmidrule(lr){4-5} \cmidrule(lr){6-7} \cmidrule(lr){8-9} \cmidrule(lr){10-11} \cmidrule(lr){12-13}
Algorithm & R $\uparrow$ & C $\downarrow$ & R $\uparrow$ & C $\downarrow$ & R $\uparrow$ & C $\downarrow$ & R $\uparrow$ & C $\downarrow$ & R $\uparrow$ & C $\downarrow$ & R $\uparrow$ & C $\downarrow$ \\
\midrule
PPO & 8262.4 & \textcolor{safegreen}{15.0} & \textbf{10433.8} & \textbf{\textcolor{safegreen}{21.8}} & 11024.5 & 36.7 & 5376.4 & 1153.0 & 5966.2 & 1244.7 & 2583.0 & 606.5 \\
PPOCost & 7163.5 & \textcolor{safegreen}{7.8} & 5121.1 & \textcolor{safegreen}{7.8} & \textbf{4302.8} & \textbf{\textcolor{safegreen}{8.9}} & 2220.5 & \textcolor{safegreen}{1.0} & 1542.4 & \textcolor{safegreen}{0.4} & 887.3 & \textcolor{safegreen}{1.5} \\
PPOLag & 8974.3 & \textcolor{safegreen}{16.6} & 6823.0 & \textcolor{safegreen}{20.0} & 2076.2 & \textcolor{safegreen}{14.8} & \textbf{2442.8} & \textbf{\textcolor{safegreen}{12.0}} & 1594.2 & \textcolor{safegreen}{13.9} & 844.1 & \textcolor{safegreen}{14.5} \\
PPOPID & 4633.2 & \textcolor{safegreen}{12.6} & 4186.2 & \textcolor{safegreen}{13.3} & 1147.7 & \textcolor{safegreen}{10.4} & 2416.0 & \textcolor{safegreen}{14.3} & 1258.6 & \textcolor{safegreen}{9.0} & \textbf{902.5} & \textbf{\textcolor{safegreen}{4.2}} \\
PPOSaute & 9095.2 & \textcolor{safegreen}{17.7} & 6978.6 & \textcolor{safegreen}{19.5} & 9304.0 & 35.1 & 2866.5 & 480.5 & 1372.0 & 267.3 & 972.3 & 220.5 \\
P3O & \textbf{9137.1} & \textbf{\textcolor{safegreen}{16.2}} & 7460.7 & \textcolor{safegreen}{19.2} & 2098.4 & \textcolor{safegreen}{13.3} & 1625.6 & \textcolor{safegreen}{19.3} & \textbf{1785.6} & \textbf{\textcolor{safegreen}{18.1}} & 856.4 & \textcolor{safegreen}{19.7} \\
FOCOPS & 876.5 & \textcolor{safegreen}{6.2} & 856.3 & \textcolor{safegreen}{6.8} & 769.1 & \textcolor{safegreen}{6.3} & 228.3 & \textcolor{safegreen}{3.1} & 119.8 & \textcolor{safegreen}{3.3} & 79.1 & \textcolor{safegreen}{4.5} \\
CRPO & 1147.4 & \textcolor{safegreen}{7.3} & 785.3 & \textcolor{safegreen}{7.0} & 628.0 & \textcolor{safegreen}{7.3} & 2824.4 & 29.0 & 2953.1 & 30.0 & 2845.7 & 36.3 \\
\bottomrule
\end{tabular}
\\[1em]
\begin{tabular}{p{1.0008cm}p{0.70cm}p{0.70cm}p{0.70cm}p{0.70cm}p{0.70cm}p{0.70cm}p{0.70cm}p{0.70cm}p{0.70cm}p{0.70cm}p{0.70cm}p{0.70cm}}
\toprule
 & \multicolumn{6}{c}{Spider Lift} & \multicolumn{6}{c}{Point Push} \\
\cmidrule(lr){2-7} \cmidrule(lr){8-13}
 & \multicolumn{2}{c}{Level 1} & \multicolumn{2}{c}{Level 2} & \multicolumn{2}{c}{Level 3} & \multicolumn{2}{c}{Level 1} & \multicolumn{2}{c}{Level 2} & \multicolumn{2}{c}{Level 3} \\
\cmidrule(lr){2-3} \cmidrule(lr){4-5} \cmidrule(lr){6-7} \cmidrule(lr){8-9} \cmidrule(lr){10-11} \cmidrule(lr){12-13}
Algorithm & R $\uparrow$ & C $\downarrow$ & R $\uparrow$ & C $\downarrow$ & R $\uparrow$ & C $\downarrow$ & R $\uparrow$ & C $\downarrow$ & R $\uparrow$ & C $\downarrow$ & R $\uparrow$ & C $\downarrow$ \\
\midrule
PPO & 16.8 & 59.5 & 17.0 & 59.1 & 15.6 & 135.9 & 64.5 & 227.7 & 54.6 & 368.1 & 49.5 & 461.3 \\
PPOCost & 8.6 & \textcolor{safegreen}{23.0} & 16.3 & 56.7 & 17.5 & 115.8 & 26.9 & 150.8 & 40.1 & 333.0 & 39.5 & 418.0 \\
PPOLag & 15.5 & \textcolor{safegreen}{9.6} & 4.9 & \textcolor{safegreen}{8.8} & 0.0 & \textcolor{safegreen}{1.4} & 26.6 & \textcolor{safegreen}{17.6} & 10.9 & \textcolor{safegreen}{24.4} & 5.8 & \textcolor{safegreen}{24.9} \\
PPOPID & 15.4 & \textcolor{safegreen}{12.4} & 10.2 & \textcolor{safegreen}{14.7} & 0.0 & \textcolor{safegreen}{1.6} & 17.2 & \textcolor{safegreen}{19.9} & 7.4 & 25.2 & 5.9 & \textcolor{safegreen}{24.5} \\
PPOSaute & 11.7 & 49.8 & 13.0 & 76.1 & 17.0 & 107.2 & 41.7 & 244.3 & 34.1 & 279.7 & 25.0 & 293.6 \\
P3O & \textbf{16.6} & \textbf{\textcolor{safegreen}{12.1}} & \textbf{13.6} & \textbf{\textcolor{safegreen}{14.5}} & \textbf{0.1} & \textbf{\textcolor{safegreen}{1.4}} & 47.7 & 26.7 & 14.1 & 26.3 & 13.9 & 25.6 \\
FOCOPS & 0.4 & \textcolor{safegreen}{1.6} & 0.1 & \textcolor{safegreen}{1.0} & 0.0 & \textcolor{safegreen}{1.3} & \textbf{64.3} & \textbf{\textcolor{safegreen}{24.4}} & \textbf{40.5} & \textbf{\textcolor{safegreen}{24.6}} & \textbf{40.8} & \textbf{\textcolor{safegreen}{24.3}} \\
CRPO & 9.4 & \textcolor{safegreen}{6.7} & 0.2 & \textcolor{safegreen}{0.9} & 0.0 & \textcolor{safegreen}{1.1} & 17.3 & 25.0 & 7.7 & 26.3 & 10.1 & 25.2 \\
\bottomrule
\end{tabular}
\\[1em]
\begin{tabular}{p{1.0008cm}p{0.70cm}p{0.70cm}p{0.70cm}p{0.70cm}p{0.70cm}p{0.70cm}p{0.70cm}p{0.70cm}p{0.70cm}p{0.70cm}p{0.70cm}p{0.70cm}}
\toprule
 & \multicolumn{6}{c}{Point Circle} & \multicolumn{6}{c}{Humanoid Height} \\
\cmidrule(lr){2-7} \cmidrule(lr){8-13}
 & \multicolumn{2}{c}{Level 1} & \multicolumn{2}{c}{Level 2} & \multicolumn{2}{c}{Level 3} & \multicolumn{2}{c}{Level 1} & \multicolumn{2}{c}{Level 2} & \multicolumn{2}{c}{Level 3} \\
\cmidrule(lr){2-3} \cmidrule(lr){4-5} \cmidrule(lr){6-7} \cmidrule(lr){8-9} \cmidrule(lr){10-11} \cmidrule(lr){12-13}
Algorithm & R $\uparrow$ & C $\downarrow$ & R $\uparrow$ & C $\downarrow$ & R $\uparrow$ & C $\downarrow$ & R $\uparrow$ & C $\downarrow$ & R $\uparrow$ & C $\downarrow$ & R $\uparrow$ & C $\downarrow$ \\
\midrule
PPO & 127.0 & 45.0 & 127.2 & 98.8 & 127.5 & 99.4 & \textbf{5230.3} & \textbf{\textcolor{safegreen}{20.2}} & 3532.3 & 52.1 & 3868.4 & 53.8 \\
PPOCost & 103.8 & \textcolor{safegreen}{11.9} & 82.8 & \textcolor{safegreen}{9.3} & 80.8 & 38.1 & 3873.8 & \textcolor{safegreen}{15.0} & 3657.3 & 33.8 & 3753.8 & 57.5 \\
PPOLag & 89.9 & 25.3 & 64.5 & \textcolor{safegreen}{24.2} & 53.5 & \textcolor{safegreen}{18.7} & 2792.9 & \textcolor{safegreen}{6.4} & 1703.3 & \textcolor{safegreen}{5.5} & 1983.7 & \textcolor{safegreen}{6.2} \\
PPOPID & 90.5 & \textcolor{safegreen}{21.8} & 68.9 & 26.5 & 39.9 & \textcolor{safegreen}{21.6} & 3716.6 & \textcolor{safegreen}{7.6} & 1087.3 & \textcolor{safegreen}{5.4} & 2657.9 & \textcolor{safegreen}{6.0} \\
PPOSaute & 120.2 & 46.0 & 128.6 & 97.3 & 123.1 & 96.8 & 4847.5 & \textcolor{safegreen}{17.3} & 5652.4 & 60.9 & 5910.1 & 35.7 \\
P3O & 96.7 & 29.0 & 60.1 & 28.3 & 61.3 & 27.5 & 2328.8 & \textcolor{safegreen}{6.9} & \textbf{2310.9} & \textbf{\textcolor{safegreen}{4.5}} & \textbf{2753.3} & \textbf{\textcolor{safegreen}{5.4}} \\
FOCOPS & 88.7 & 32.5 & 62.2 & 25.1 & \textbf{57.7} & \textbf{\textcolor{safegreen}{24.0}} & 2563.5 & \textcolor{safegreen}{6.4} & 1244.1 & \textcolor{safegreen}{6.9} & 376.4 & \textcolor{safegreen}{7.7} \\
CRPO & \textbf{114.5} & \textbf{\textcolor{safegreen}{24.7}} & \textbf{88.4} & \textbf{\textcolor{safegreen}{24.4}} & 67.4 & 25.3 & 1847.4 & \textcolor{safegreen}{4.9} & 2013.3 & \textcolor{safegreen}{5.5} & 1419.9 & \textcolor{safegreen}{5.3} \\
\bottomrule
\end{tabular}
\end{table}

\subsection{Training curves}

\begin{figure}[ht]
    \centering
    \includegraphics[width=\linewidth]{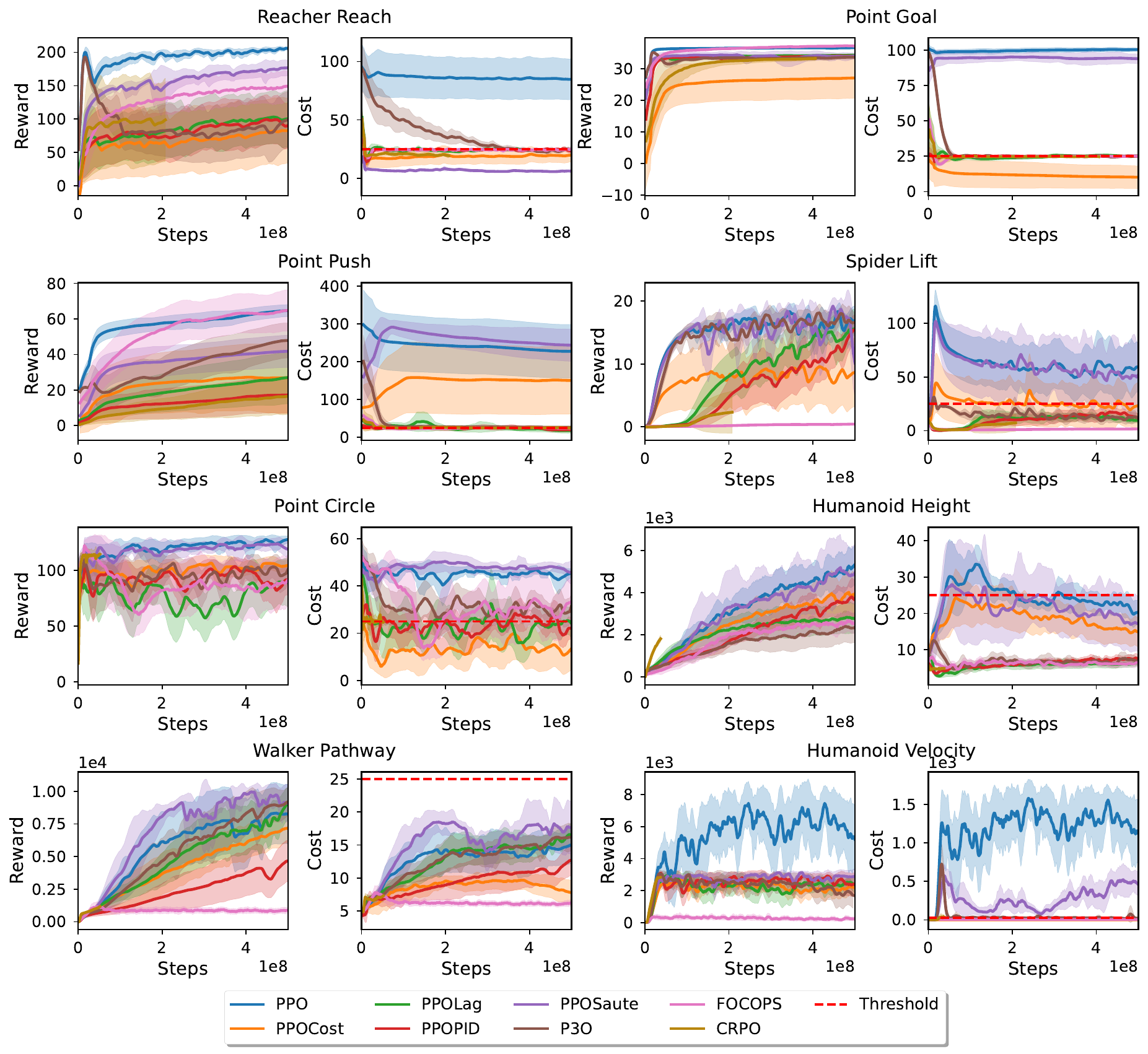}
    \caption{Level 1 training curves.}
    \label{fig:curves_level_1}
\end{figure}

\begin{figure}[ht]
    \centering
    \includegraphics[width=\linewidth]{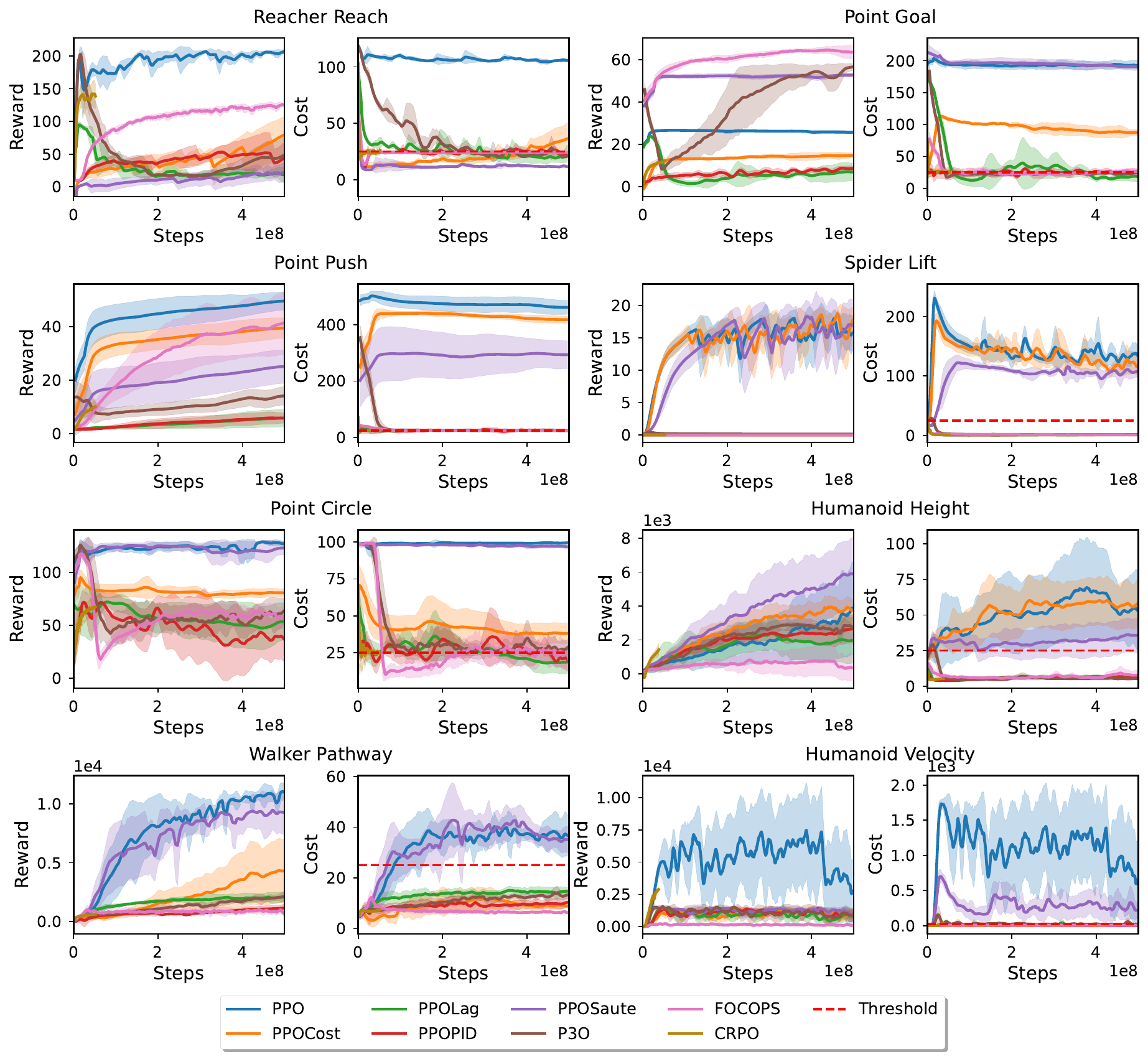}
    \caption{Level 3 training curves.}
    \label{fig:curves_level_3}
\end{figure}

\cref{fig:curves_level_1,fig:curves_level_3} show training curves for all baseline methods across difficulty levels, illustrating differences in learning dynamics and convergence behavior.

\begin{figure}[ht]
    \centering
    \includegraphics[width=\linewidth]{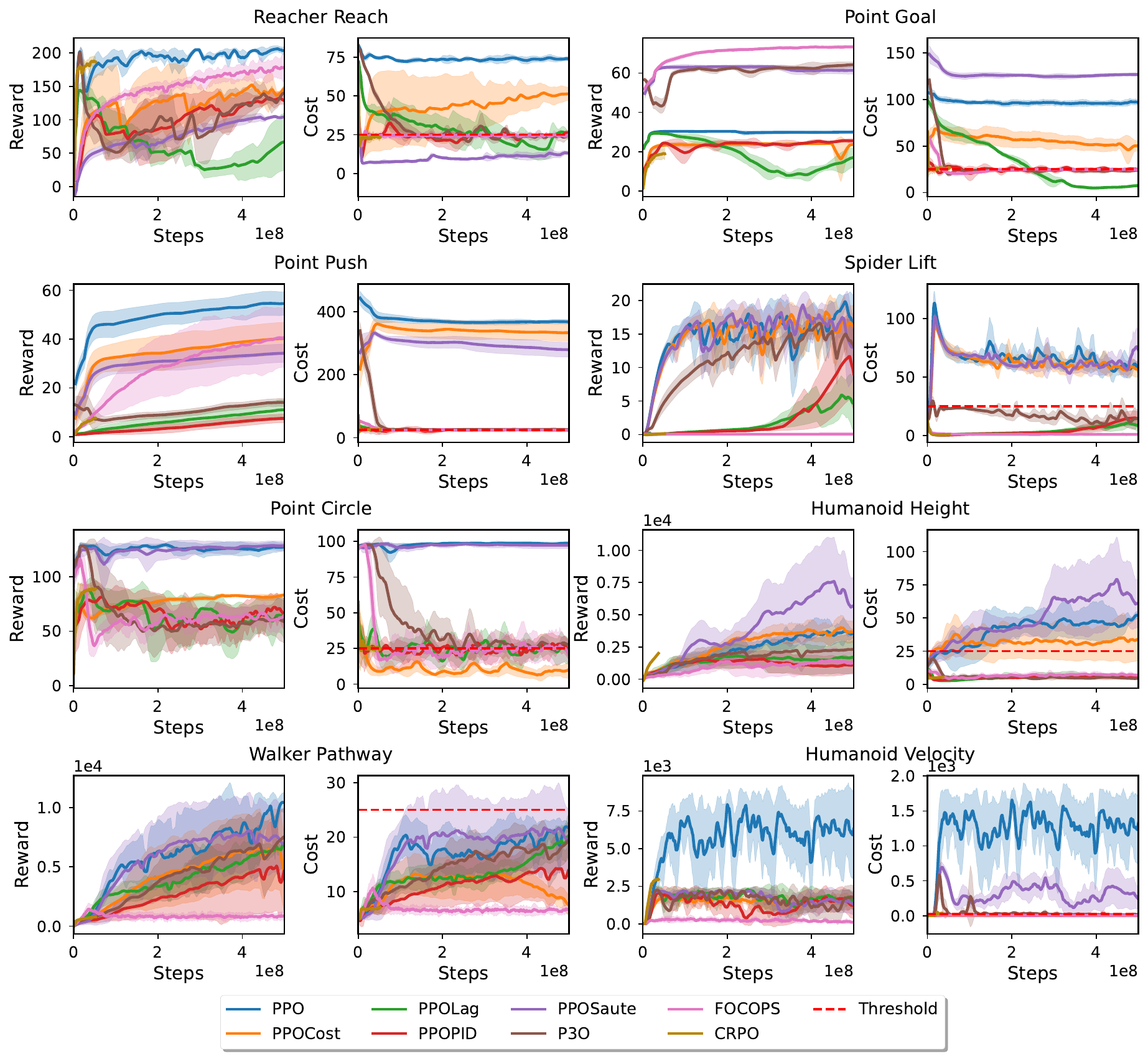}
    \caption{Level 2 training curves.}
    \label{fig:curves_level_2}
\end{figure}

\clearpage

\clearpage
\end{document}